\documentclass[11pt]{article}
\pdfoutput=1
\usepackage[preprint]{acl}
\usepackage{siunitx}
\usepackage{subfig}
\usepackage{subcaption}
\usepackage{color,soul}

\usepackage{times}
\usepackage{latexsym}
\usepackage[T1]{fontenc}
\usepackage[utf8]{inputenc}
\usepackage{microtype}
\usepackage{inconsolata} 
\usepackage{csquotes}
\usepackage{enumitem}
\usepackage{pifont}

\usepackage{amssymb}
\usepackage{amsthm}
\usepackage{mathtools}
\usepackage{amsmath, amssymb}   

\usepackage[ruled,vlined,linesnumbered]{algorithm2e}
\SetKwComment{tcp}{\small\textcolor{gray}{$\triangleright$\,}}{}
\DontPrintSemicolon

\usepackage{booktabs}
\usepackage{multirow}
\usepackage{multicol}
\usepackage{tabularx}
\usepackage{tabulary}
\usepackage{adjustbox}
\usepackage{siunitx}
\usepackage{makecell}

\newtheorem{definition}{Definition}
\newtheorem{theorem}{Theorem}
\newtheorem{lemma}{Lemma}
\newtheorem{proposition}{Proposition}
\newtheorem{corollary}{Corollary}

\usepackage{graphicx}
\usepackage{bm}
\usepackage{xcolor}
\usepackage{xcolor,colortbl}

\definecolor{darkgreen}{rgb}{0.0, 0.5, 0.0}
\definecolor{darkred}{rgb}{0.6, 0.0, 0.0}
\definecolor{navy}{rgb}{0.0, 0.0, 0.5}

\usepackage[most]{tcolorbox} 
\usepackage{cleveref}
\usepackage{url}
\usepackage{makecell}
\usepackage{fontawesome5}

\newtcbtheorem[auto counter, number within=section, 
crefname={example}{Example},
Crefname={Example}{Example}]
{exmp}{Exam\smash{p}le} 
{colback=pale_red!5, colframe=DarkPurple, left=.02in, right=.02in,bottom=.02in, top=.02in}{exmp}  

\newtcbtheorem[auto counter, 
crefname={prompt}{Prompt},
Crefname={Prompt}{Prompt}]
{prompt}{Prom\smash{p}t} 
{colback=teal!5!white, colframe=teal!60!black, left=.02in, right=.02in,bottom=.02in, top=.02in}{prompt}  

\usepackage{listings}
\definecolor{Violet}{RGB}{148,0,211}
\definecolor{DarkPurple}{RGB}{75,0,130}
\definecolor{lightergray}{RGB}{230,230,230}
\definecolor{DarkRed}{RGB}{130,25,0}
\definecolor{PurpleRed}{RGB}{204,0,102}
\definecolor{DarkGreen}{RGB}{30,130,30}
\definecolor{DarkBlue}{RGB}{0,0,250}
\definecolor{DarkYellow}{RGB}{255,128,0}
\definecolor{darkyellow}{RGB}{200,120,0}
\definecolor{light-gray}{gray}{0.95}
\definecolor{lightgreen}{RGB}{231,255,219}
\definecolor{lightred}{RGB}{252,231,234}
\definecolor{lightyellow}{RGB}{250,253,191}
\definecolor{lightpurple}{RGB}{229,204,255}
\definecolor{lightblue}{RGB}{229,246,254}
\definecolor{value-modification}{RGB}{250, 217, 86}
\definecolor{digit-expansion}{RGB}{216, 194, 104}
\definecolor{integer-decimal-fraction}{RGB}{240, 133, 51}
\definecolor{semantic-paraphrasing}{RGB}{85, 157, 63}
\definecolor{complexity-increasing}{RGB}{58, 120, 175}
\definecolor{question-transformation}{RGB}{174, 205, 225}
\definecolor{interference-injection}{RGB}{255,204,229}
\definecolor{remove-constrain}{RGB}{204,204,255}

\newcommand{\xmark}{\textcolor{red}{\ding{55}}} 
\newcommand{\cmark}{\textcolor{DarkGreen}{\ding{51}}}

\definecolor{pale_green}{rgb}{0.55,0.75,0.60}
\definecolor{pale_red}{rgb}{0.90,0.61,0.58}
\definecolor{pale_yellow}{rgb}{0.95,0.92,0.72}

\tcbset{
  aibox/.style={
    width=\textwidth,
    top=0pt, bottom=0pt, left=5pt, right=5pt,
    colback=white,
    colframe=black,
    colbacktitle=black,
    enhanced,
    center,
    attach boxed title to top left={yshift=-0.1in,xshift=0.15in},
    boxed title style={boxrule=0pt,colframe=white,},
  }
}
\newtcolorbox{AIbox}[2][]{aibox,title=#2,#1}
\addtocontents{toc}{\protect\setcounter{tocdepth}{-1}} 

\usepackage{amsmath,amssymb,bbm}   

\usepackage{graphicx}    
\usepackage{booktabs}    

\usepackage{tcolorbox}
\usepackage{xcolor}

\definecolor{defaultcolor}{HTML}{DAE8FC} 
\definecolor{deterministiccolor}{HTML}{D5E8D4} 
\definecolor{exploratorycolor}{HTML}{FFF2CC} 
\definecolor{detexploratorycolor}{HTML}{FADBD8} 

\newcommand{\llmthinkbench}{\textsc{LLMThinkBench}}

\title{\textit{Do LLMs Overthink Basic Math Reasoning?}\\Benchmarking the Accuracy-Efficiency Tradeoff in Language Models}

\author{
 \textbf{Gaurav Srivastava\textsuperscript{$\heartsuit$}},
 \textbf{Aafiya Hussain\textsuperscript{$\heartsuit$}},
  \textbf{Sriram Srinivasan\textsuperscript{$\heartsuit$}},
 \textbf{Xuan Wang\textsuperscript{$\heartsuit$}}
\\
 \textsuperscript{$\heartsuit$}Department of Computer Science, Virginia Tech, Blacksburg, VA, USA,
\\
\small{
  \textsuperscript{$\heartsuit$}\texttt{(\href{mailto:gks@vt.edu}{gks}, \href{mailto:aafiyahussain@vt.edu}{aafiyahussain}, \href{mailto:sriramsrinivasan@vt.edu}{sriramsrinivasan}, \href{mailto:xuanw@vt.edu}{xuanw})@vt.edu}
  }
  \\[0.5em]
{\small\strut \faGlobe~\textbf{Leaderboard:} \href{https://ctrl-gaurav.github.io/LLMThinkBench/}{\textmd{\texttt{ctrl-gaurav.github.io/LLMThinkBench}}}} \\
{\small\strut \faGithub~\textbf{GitHub:} \href{https://github.com/ctrl-gaurav/LLMThinkBench}{\textmd{\texttt{ctrl-gaurav/LLMThinkBench}}} \quad \faPython~\textbf{PyPI:} \href{https://pypi.org/project/llmthinkbench/}{\textmd{\texttt{pypi.org/project/llmthinkbench}}}}
}

\begin{document}

\maketitle

\begin{abstract}
Large language models (LLMs) achieve impressive performance on complex mathematical benchmarks yet sometimes fail on basic math reasoning while generating unnecessarily verbose responses. In this paper, we present \textbf{\llmthinkbench}, a systematic benchmark and comprehensive empirical study to evaluate the efficiency of reasoning in LLMs, focusing on the fundamental tradeoff between accuracy and overthinking. \textbf{First,} we formalize the \textit{accuracy-verbosity tradeoff}. \textbf{Second,} we introduce the \emph{Overthinking Score}, a harmonic-mean metric combining accuracy and token-efficiency for holistic model evaluation. \textbf{Third,} we establish an evaluation protocol with dynamically-generated data across \textbf{14} basic math tasks. \textbf{Fourth,} we conduct a large-scale empirical study evaluating \textbf{53} LLMs, including reasoning and quantized variants across different reasoning budgets. \textbf{Fifth,} we release \llmthinkbench\ as an open-source Python package and public leaderboard for reproducibility. Our findings reveal: \textbf{\textit{1)}} model performance on complex benchmarks does not translate directly to basic math reasoning; \textbf{\textit{2)}} reasoning models generate \textbf{$\sim$18$\times$ more tokens} while sometimes achieving \textbf{lower accuracy} and exhibit catastrophic collapse when tokens are constrained, dropping by up to \textbf{$\sim$36\%}; \textbf{\textit{3)}} the accuracy-verbosity relationship is non-monotonic with extended reasoning budgets yielding diminishing returns (GPT-5/o-series models show zero accuracy gain from \textbf{low $\rightarrow$ medium $\rightarrow$ high} reasoning effort). \textit{Our findings challenge the assumption that longer reasoning in LLMs necessarily improves mathematical reasoning.}
\end{abstract}

\section{Introduction}
Modern large language models (LLMs) \cite{abdin2025phi4reasoningtechnicalreport, yang2025qwen3technicalreport, xu2025largereasoningmodelssurvey} often produce long chain-of-thought when answering arithmetic and other deterministic tasks. Intuitively, more reasoning tokens should help models avoid mistakes by giving them room to compute. However, models often generate verbose traces that do not improve final correctness and, under constrained budgets, can degrade accuracy \cite{lin2025plan, qu2025survey}. \textit{This reveals a fundamental issue:} while models can produce sophisticated-looking reasoning chains, they often generate excessive verbosity that neither improves accuracy nor demonstrates genuine understanding. This phenomenon is known as \textit{overthinking in LLMs}.

Consider a model computing $234 + 567$. While humans solve this with minimal steps, language models often generate hundreds of tokens explaining place values, carrying operations, and mathematical principles; yet sometimes arrive at incorrect answers. Recent studies \cite{srivastava2025reasoningabilitysmalllanguage, yan2025phdlevelllmstrulygrasp} show this paradox: models achieving 90\% accuracy on GSM8K may score below 40\% on basic addition. More concerning, reasoning models specifically trained for deeper thinking~\cite{abdin2025phi4reasoningtechnicalreport,illusion-of-thinking} perform \textit{worse} while generating more tokens. This inverse relationship between verbosity and accuracy suggests that current models conflate explanation with understanding \cite{zhang2024verbosity}, producing text that superficially resembles reasoning without actual problem-solving capability.

Previous work has approached this problem from different angles. Chain-of-thought prompting \cite{wei2022chain} and its variants \cite{yao2023tree, Besta_2024, pandey2025adaptive, yao2023beyond, jin2024graph, ranaldi2023empowering, li2025cot, srivastava2025debate} encourage step-by-step reasoning but increase computational costs without guaranteeing accuracy improvements \cite{huang2025reasoning, chen2025towards}. Recent studies on overthinking \cite{chen2025think23overthinkingo1like, sui2025stopoverthinkingsurveyefficient} focus on mitigation strategies like early stopping \cite{pu2025thoughtterminatorbenchmarkingcalibratingmitigating} or self-breaking \cite{zhao2025letllmsbreakfree}, but lack principled metrics to quantify the phenomenon. Efficiency-focused evaluations \cite{li2025thinkbenchevaluatingthinkingefficiency} measure thinking time or API calls but treat accuracy and efficiency as independent dimensions, missing their fundamental tradeoff. Most critically, existing benchmarks evaluate only final accuracy, remaining blind to computational waste that makes models impractical for deployment.

Three key gaps remain in proper understanding of overthinking: \textbf{\textit{(1)}} absence of principled metrics that jointly measure correctness and efficiency, as existing benchmarks treat these as separate dimensions rather than a tradeoff; \textbf{\textit{(2)}} reliance on static benchmarks vulnerable to contamination; and \textbf{\textit{(3)}} lack of robust parsing \textit{(fair evaluation)} to extract answers from diverse model outputs reliably. These gaps prevent distinguishing genuine reasoning from performative verbosity. To address these gaps: \textbf{\textit{1)}} we introduce the \textbf{Overthinking Score}, a harmonic-mean metric that penalizes imbalance between accuracy and token efficiency, forcing models to be both correct and concise. \textbf{\textit{2)}} we design a dynamic test generation protocol across $14$ basic math reasoning tasks ensuring fresh, reproducible evaluation. \textbf{\textit{3)}} we conduct large-scale empirical studies studying efficiency-accuracy tradeoffs missed by accuracy-only metrics. \textbf{\textit{4)}} we release \llmthinkbench, an open-source framework that packages the benchmark, the Overthinking Score, and our parsing pipeline into a single command line tool (available on PyPI as \texttt{llmthinkbench}), together with a public leaderboard so that others can reproduce our results and evaluate new models on the same protocol (see Section~\ref{sec:framework} and Appendix~\ref{app:framework}).

Our systematic benchmark evaluation of $53$ LLMs across multiple experimental conditions yields striking insights: \textbf{\textit{1)}} The Overthinking Score reveals efficiency-accuracy tradeoffs missed by accuracy-only metrics, e.g., smaller models outperforming larger ones through better token efficiency.
\textbf{\textit{2)}} Reasoning models produce \textbf{$\sim$6{,}780} tokens on average versus \textbf{$378$} for standard models while performing worse; accuracy collapses under token limits (from $72\%$ to $44\%$ at $1{,}024$ tokens); and quantized models retain strong reasoning ability, implying overthinking arises from training rather than hardware constraints. \textbf{\textit{3)}} Extended reasoning budgets demonstrate sharply diminishing returns, with models showing minimal accuracy gains beyond moderate token allocations while introducing contradictions and error accumulation in long chains, revealing fundamental limitations in current reasoning paradigms. \textbf{\textit{4)}} Through comprehensive ablation studies including quantization effects, constrained generation, and reasoning budget analysis, our benchmark provides a standardized framework for evaluating reasoning efficiency that complements existing accuracy-focused benchmarks.

\section{Related Work}

\paragraph{\textsc{Mathematical Reasoning in LLMs}} 
LLMs have achieved impressive performance on benchmark datasets such as GSM8K~\cite{cobbe2021trainingverifierssolvemath}, GSM-Plus~\cite{li2024gsmpluscomprehensivebenchmarkevaluating}, MATH~\cite{hendrycksmath2021}, and HARDMath~\cite{fan2024hardmathbenchmarkdatasetchallenging}. Yet recent work has highlighted a striking gap: models that excel at word problems often fail on simple arithmetic \cite{srivastava2025reasoningabilitysmalllanguage, srivastava2025thinkslm, srivastava2026effgenenablingsmalllanguage, bi2026judgeboard, yan2025phdlevelllmstrulygrasp, xu2024llm}. To expose these weaknesses, new benchmarks of simple numerical tasks have been proposed \cite{li2025exposingnumeracygapsbenchmark, rahman2025largelanguagemodelsnumberland, srivastava2025beyondbench}, showing that models struggle with core numeracy skills despite advanced reasoning capabilities elsewhere. Several approaches aim to improve mathematical reasoning. For instance, Graph of Thoughts \cite{Besta_2024} structures intermediate reasoning as a graph, improving performance on algorithmic tasks such as sorting. LogicPuzzleRL \cite{wong2025logicpuzzlerlcultivatingrobustmathematical} uses reinforcement learning on custom logic puzzles to strengthen algebraic, geometric, and combinatorial reasoning. While these approaches enhance problem-solving on complex tasks, they provide limited insight into why models underperform on simple arithmetic or how to measure reasoning efficiency.

\paragraph{\textsc{Overthinking in LLMs}} 
A complementary line of research shows that LLMs often produce unnecessarily long reasoning chains for trivial problems \cite{chen2025think23overthinkingo1like}. This phenomenon, termed \textit{overthinking}, wastes computation and sometimes reduces accuracy. Mitigation strategies include efficiency-oriented prompting, reasoning-output pruning, and model-based interventions \cite{sui2025stopoverthinkingsurveyefficient}. Specific methods such as ThoughtTerminator \cite{pu2025thoughtterminatorbenchmarkingcalibratingmitigating}, Self-Braking Tuning \cite{zhao2025letllmsbreakfree}, and process supervision approaches~\cite{lightman2023letsverifystepstep,luo2024improvemathematicalreasoninglanguage,latimer2025hindsight} actively shorten reasoning or improve step-level verification without hurting accuracy. However, prior work largely emphasizes mitigation heuristics rather than providing a principled way to \emph{quantify} overthinking. Without robust metrics, it remains unclear how to diagnose or compare models' efficiency in reasoning. \llmthinkbench\ fills this gap by pairing the \textbf{Overthinking Score} with a dynamic task generator and a reusable evaluation pipeline.

\section{\llmthinkbench{} Framework}
\label{sec:framework}

We package our benchmark, metric, and evaluation pipeline into a single open-source framework called \llmthinkbench. The framework has four parts that map one-to-one to the rest of this section: a task space of 14 deterministic basic-math tasks (\S\ref{sec:task_space}), a formal accuracy-verbosity space (\S\ref{sec:av_space}), the Overthinking Score that collapses that space into one number (\S\ref{sec:oscore}), and a runnable tool that ties these together. \llmthinkbench\ is installable from PyPI with \texttt{pip install llmthinkbench} and exposes a single command line entry point that generates fresh test instances, queries a model through Hugging Face, OpenAI, Anthropic, Google, or any user-supplied backend, parses answers with the hierarchical extractor described in \S\ref{sec:methodology}, and writes per-task and aggregate reports including the Overthinking Score. The same generator, parser, and scoring code produced every number in this paper, so third-party runs stay comparable to ours. A public leaderboard at \href{https://ctrl-gaurav.github.io/LLMThinkBench/}{\texttt{ctrl-gaurav.github.io/LLMThinkBench}} lists current results. Full usage, internals, and reproducibility notes are in Appendix~\ref{app:framework}.

\subsection{Task Space}
\label{sec:task_space}

We formalize the evaluation of basic mathematical reasoning through a task space $\mathcal{T} = \{\tau_1, \tau_2, ..., \tau_K\}$ comprising $K = 14$ deterministic arithmetic operations. Each task $\tau_i$ represents a mapping from an input domain $\mathcal{X}_i$ to an output domain $\mathcal{Y}_i$, where the ground truth function $f_i: \mathcal{X}_i \rightarrow \mathcal{Y}_i$ is computationally deterministic and mathematically well-defined.

\begin{definition}[Deterministic Arithmetic Task]
A deterministic arithmetic task $\tau = (\mathcal{X}, \mathcal{Y}, f, \mathcal{C})$ consists of:
\begin{itemize}
    \item An input space $\mathcal{X} \subseteq \mathbb{Z}^n$ or $\mathcal{X} \subseteq \mathbb{Z} \times \mathbb{Z}$
    \item An output space $\mathcal{Y} \subseteq \mathbb{Z}$ or $\mathcal{Y} \subseteq \mathbb{Q}$ or $\mathcal{Y} \subseteq \mathcal{P}(\mathbb{Z})$
    \item A deterministic function $f: \mathcal{X} \rightarrow \mathcal{Y}$ with complexity $\mathcal{C} \in \{O(n), O(n \log n), O(n^2)\}$
\end{itemize}
\end{definition}

Table~\ref{tab:task_suite} presents our complete task suite spanning four categories. For list-based operations, inputs are defined as $\mathcal{L} = \{l_1, l_2, ..., l_n\}$ where $l_i \in \mathbb{Z}$ and $n \in \mathbb{N}$. Complete mathematical foundations for each task, including uniqueness proofs and complexity analysis, are provided in Appendix~\ref{app:tasks}. Experimental configuration details including sampling strategies, hyperparameters, and computational resources are described in Appendix~\ref{app:experimental_config}.

\begin{table}[ht]
\centering
\scriptsize
\begin{adjustbox}{max width=\linewidth}
\begin{tabular}{p{0.8cm}p{1.8cm}p{1.6cm}p{1.8cm}}
\toprule
\textbf{Cat.} & \textbf{Task} & \textbf{Input} & \textbf{Output} \\
\midrule
\multirow{7}{*}{\rotatebox{90}{\scriptsize Basic}}
& Sorting & $\mathcal{L} \subset \mathbb{Z}^n$ & Ordered list \\
& Comparison & $(a, b)$ & $\{>, <, =\}$ \\
& Sum & $\mathcal{L} \subset \mathbb{Z}^n$ & $\sum_i \mathcal{L}_i$ \\
& Subtraction & $(a, b)$ & $b - a$ \\
& Multiplication & $\mathcal{L} \subset \mathbb{Z}^n$ & $\prod_i \mathcal{L}_i$ \\
& Division & $(a, b), b \neq 0$ & $a/b$ \\
& Abs. Diff & $(a, b)$ & $|a - b|$ \\
\midrule
\multirow{2}{*}{\rotatebox{90}{\scriptsize Ext.}} 
& Find Max & $\mathcal{L} \subset \mathbb{Z}^n$ & $\max(\mathcal{L})$ \\
& Find Min & $\mathcal{L} \subset \mathbb{Z}^n$ & $\min(\mathcal{L})$ \\
\midrule
\multirow{3}{*}{\rotatebox{90}{\scriptsize Stats}} 
& Mean & $\mathcal{L} \subset \mathbb{Z}^n$ & $\frac{1}{n}\sum_i \mathcal{L}_i$ \\
& Median & $\mathcal{L} \subset \mathbb{Z}^n$ & $\text{med}(\mathcal{L})$ \\
& Mode & $\mathcal{L} \subset \mathbb{Z}^n$ & $\arg\max_v f(v)$ \\
\midrule
\multirow{2}{*}{\rotatebox{90}{\scriptsize Count}} 
& Odd Count & $\mathcal{L} \subset \mathbb{Z}^n$ & $|\{x: x \bmod 2 = 1\}|$ \\
& Even Count & $\mathcal{L} \subset \mathbb{Z}^n$ & $|\{x: x \bmod 2 = 0\}|$ \\
\bottomrule
\end{tabular}
\end{adjustbox}
\caption{Task suite comprising 14 deterministic tasks with formally specified input-output mappings.}
\label{tab:task_suite}
\end{table}

\subsection{The Accuracy-Verbosity Space}
\label{sec:av_space}

We conceptualize model behavior in a two-dimensional space $\mathcal{S} = [0, 1] \times \mathbb{R}^+$ representing accuracy and average token count. Given a language model $M$ with parameters $\theta$, the response generation process is:

\begin{equation}
r = M(p(x, \tau_i); \theta, \phi)
\end{equation}

where $p: \mathcal{X}_i \times \mathcal{T} \rightarrow \mathcal{S}$ is a prompt construction function and $\phi = \{T, p_{top}, \text{max\_tokens}\}$ represents inference hyperparameters.

\begin{definition}[Overthinking]
A model $M$ exhibits \textbf{overthinking} on task $\tau_i$ if there exists a more concise model $M'$ such that:
\begin{align}
\mathbb{E}_{x \sim \mathcal{D}_i}[\mathbb{I}[\mathcal{P}(M'(x)) = f_i(x)]] &\geq \\\mathbb{E}_{x \sim \mathcal{D}_i}[\mathbb{I}[\mathcal{P}(M(x)) = f_i(x)]]\\
\mathbb{E}_{x \sim \mathcal{D}_i}[|M'(x)|] &< \mathbb{E}_{x \sim \mathcal{D}_i}[|M(x)|]
\end{align}
\end{definition}

\subsection{The Overthinking Score}
\label{sec:oscore}

We introduce a principled metric for quantifying the accuracy-verbosity tradeoff. For a model $M$ evaluated on task $\tau_i$, let $A_i$ denote its accuracy (proportion of correct answers) and $\bar{T}_i$ denote the average number of output tokens generated per instance. To normalize token counts across models, we define:

\begin{definition}[Token Efficiency]
Let $T_{min}$ and $T_{max}$ denote the minimum and maximum average token counts observed across all models on all tasks in our evaluation. The token efficiency for model $M$ on task $\tau_i$ is:
\begin{equation}
E_{t,i} = 1 - \frac{\bar{T}_i - T_{min}}{T_{max} - T_{min}}
\end{equation}
where $E_{t,i} \in [0, 1]$, with higher values indicating more efficient (concise) generation.
\end{definition}

\begin{definition}[Overthinking Score]
The Overthinking Score for model $M$ on task $\tau_i$ combines accuracy and token efficiency via harmonic mean:
\begin{equation}
\mathcal{O}_i = \frac{2 \cdot A_i \cdot E_{t,i}}{A_i + E_{t,i}}
\end{equation}
\end{definition}

The harmonic mean formulation ensures critical properties including boundedness ($\mathcal{O}_i \in [0, 1]$), symmetry, and severe penalization of imbalance (see Appendix~\ref{app:theoretical} for complete proofs and sensitivity analysis; Appendix~\ref{app:pareto} gives the Pareto frontier view of the accuracy-efficiency plane, which complements the single-number score). A model achieving 90\% accuracy with 10\% efficiency scores only $\mathcal{O} = 0.18$, while a balanced model with 60\% in both dimensions achieves $\mathcal{O} = 0.60$. We complement this metric with supplementary measurements including average token count, word count, and character count for comprehensive evaluation of the accuracy-efficiency balance.

\subsection{Alternative Formulations}

We evaluated several aggregation functions before adopting the harmonic mean. The arithmetic mean ($\mathcal{O}_{arith} = \frac{A + E_t}{2}$) fails to penalize imbalance adequately: a model with perfect accuracy but 10\% efficiency still scores 0.55. The geometric mean provides better balance sensitivity but lacks the strong penalty that characterizes overthinking behavior. Among all symmetric, homogeneous means, the harmonic mean maximally penalizes imbalance while maintaining smooth differentiability (detailed comparison and proofs in Appendix~\ref{app:alternative}).

\section{Experimental Methodology}
\label{sec:methodology}

\subsection{Dynamic Test Generation}

To ensure evaluation free from data contamination, we implement a dynamic test generation protocol. Let $\mathcal{G}: \Theta \times \mathcal{S} \rightarrow \mathcal{D}$ represent our generation function where $\Theta = \{N, F, [r_{min}, r_{max}], \mathcal{L}, s\}$ specifies generation parameters. For each task $\tau$ and fold $f \in \{1, ..., F\}$, we generate test instances by: \textbf{\textit{1)}} Setting seed $s_f = \text{hash}(s, f, \tau)$ for reproducibility; \textbf{\textit{2)}} Sampling list lengths from $\mathcal{L} = \{8, 16, 32, 64\}$ for list-based tasks (except multiplication which uses $\{2, 4, 8\}$ numbers to avoid overflow); \textbf{\textit{3)}} Drawing values from $\text{Uniform}[-1000, 1000]$ ensuring numerical diversity; \textbf{\textit{4)}} Computing ground truth $y = f_\tau(x)$ for validation

We use $N = 1000$ samples per fold with $F = 3$ folds for open-source models (100 samples for closed-source due to cost), generating 42,000 unique problems per model. The complete generation algorithm, validation, and reproducibility protocols are provided in Appendix~\ref{app:implementation} and Appendix~\ref{app:data_generation}.

\subsection{Model Selection and Inference}

We evaluate 53 models spanning multiple families (GPT \cite{achiam2023gpt}, Gemini \cite{comanici2025gemini}, Llama \cite{grattafiori2024llama}, Mistral \cite{parmar2024nemotron}, Qwen \cite{yang2025qwen3}, Phi \cite{abdin2025phi4reasoningtechnicalreport}) with parameters from 0.5B-72B. This includes base, instruction-tuned, and reasoning variants to isolate the impact of training objectives on overthinking behavior. 

\begin{figure*}[ht]
    \centering
    \includegraphics[width=1\textwidth]{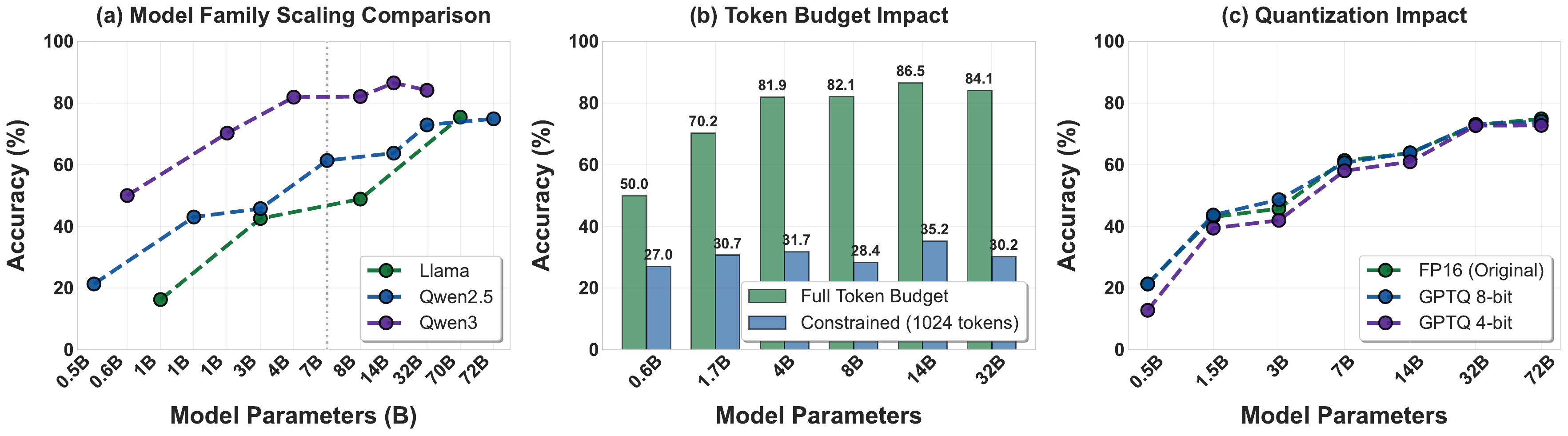}
    \caption{Three dimensions of our evaluation. \textbf{(a) Scaling performance across different families:} \textit{Llama} shows large jumps between small and large models. \textit{Qwen2.5} shows generally monotonic but sublinear scaling. \textit{Qwen3} exhibits non-monotonic behavior with 14B outperforming 32B. \textbf{(b) Token budget constraints:} \textit{Qwen3 reasoning models} accuracy under full budget vs. \textit{1024-token limit} reveals catastrophic degradation. \textbf{(c) Quantization robustness:} \textit{Qwen2.5 family} across FP16, 8-bit, and 4-bit precision shows size-dependent tolerance to compression.}
    \label{fig:combined}
\end{figure*}

\subsection{Prompt Construction and Response Parsing}

For each test instance, we construct standardized prompts: $p(x_i, \tau) = \text{desc}(\tau) \oplus \text{format}(x_i) \oplus \text{instr}$ where $\text{instr} = $ ``Your final answer must be in the format \textbackslash boxed\{answer\}''. Complete prompt templates for all 14 tasks are provided in Appendix~\ref{app:prompts}.

Extracting answers from diverse model outputs requires robust parsing. Through analysis of 5000+ responses, we developed a hierarchical extraction system achieving 98.7\% success rate (detailed design principles and validation procedures in Appendix~\ref{app:parsing_framework}): \textbf{\textit{1) Primary:}} Extract content within \textbackslash boxed\{\} patterns; \textbf{\textit{2) Secondary:}} Parse explicit answer markers (``The answer is...''); \textbf{\textit{3) Tertiary:}} Extract from code blocks or markdown formatting \textbf{\textit{4) Fallback:}} Apply task-specific heuristics.

Task-specific validation ensures extracted answers represent solutions rather than input echoes. We employ a hierarchical extraction strategy that prioritizes LaTeX boxed notation, then explicit answer markers, code block outputs, and finally last-line heuristics. The complete end-to-end pipeline (Algorithm~\ref{alg:pipeline}) and hierarchical answer extraction algorithm (Algorithm~\ref{alg:extraction}) are detailed in Appendix~\ref{app:implementation}.

\subsection{Statistical Analysis}

We use k-fold cross-validation with $k = 3$, reporting mean and standard deviation across folds. Token counting uses model-specific tokenizers when available, with API response tokens for closed-source models. Statistical significance testing, error taxonomy, and detailed empirical analysis including correlation studies are presented in Appendix~\ref{app:empirical}.

\section{Results and Insights}
\label{sec:results_insights}

We ran \llmthinkbench\ on 53 models across basic mathematical tasks, quantization settings, and constrained generation budgets (refer to Table \ref{tab:comprehensive_model_performance} for raw numbers; Appendix~\ref{app:extended_results} contains the extended tables). Our high level results show that \emph{larger models or models that produces more thinking tokens are not reliably better}. Models that perform well on complex benchmarks often fail at elementary operations; models trained to reason more tend to produce long, confident-looking text that sometimes \textit{hurts} correctness; and quantization often preserves basic arithmetic ability for large models.

\subsection{\textsc{The Basic-Math Paradox}}
\label{sec:basic_math_paradox}

\begin{figure}[ht]
    \centering
    \includegraphics[width=0.48\textwidth]{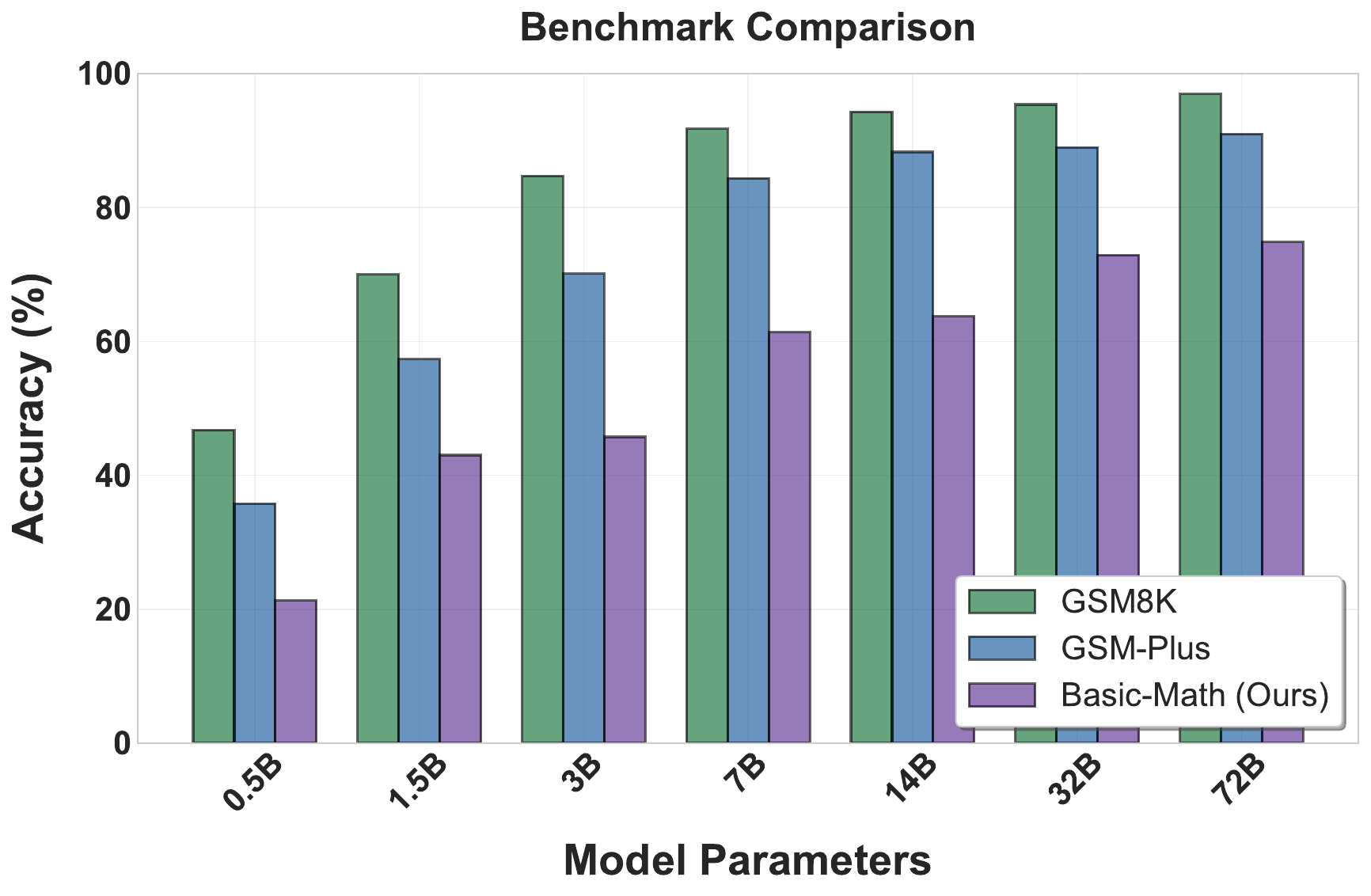}
    \caption{\textbf{Benchmark paradox:} \textit{Qwen2.5 family} performance on GSM8K, GSM-Plus, and basic math reasoning shows significant discrepancies across models.}
    \label{fig:benchmark_comparison}
    \vspace{-1.5em}
\end{figure}

Figure~\ref{fig:benchmark_comparison} shows the discrepancy between benchmark performance (GSM8K, GSM-Plus) and basic arithmetic capability for the Qwen2.5 family. Models achieving over 95\% on GSM8K score below 75\% on our basic math tasks, revealing that complex benchmark performance does not transfer to fundamental operations. This pattern suggests models learn \emph{heuristics} or pattern-matching that work for the training distribution of complex benchmarks rather than compositional arithmetic ones.

\begin{tcolorbox}[
    colback=teal!5!white,
    colframe=teal!60!black,
    title=\textbf{\textit{Finding 1 --- Basic-Math Paradox}},
    fonttitle=\bfseries,
    boxrule=0.7pt,
    arc=1mm
]
\textbf{Models with high complex benchmark performance degrade on basic math reasoning.} Performance is task-specific and suggests reliance on memorized heuristics rather than compositional computation.
\end{tcolorbox}

\begin{table*}[ht]
\centering
\scriptsize
\begin{adjustbox}{width=\textwidth}
\begin{tabulary}{1.35\textwidth}{LCCCCCCC}
\toprule
    \textbf{Models} & \textbf{Parameters (B)} & \textbf{Accuracy (Avg\%)} & \textbf{Instruction Following (Avg\%)} & \textbf{Overthinking Score (Avg)} & \textbf{Output Tokens (Avg)} & \textbf{Output Words (Avg)} & \textbf{Output Chars (Avg)} \\
\cmidrule(lr){3-3} \cmidrule(lr){4-4} \cmidrule(lr){5-5} \cmidrule(lr){6-6} \cmidrule(lr){7-7} \cmidrule(lr){8-8}
& & \textit{Mean $\pm$ Std} & \textit{Mean $\pm$ Std} & & & & \\
\midrule
\rowcolor{gray!10}
\multicolumn{8}{c}{\textbf{\textit{Qwen3 Family}}} \\
\midrule
Qwen3 & 0.6 & 49.99 $\pm$ 3.53 & 83.85 $\pm$ 2.23 & 0.545 & 3162.8 & 1620.9 & 8301.8 \\
Qwen3 & 1.7 & 70.24 $\pm$ 3.62 & 86.54 $\pm$ 2.35 & 0.647 & 3157.2 & 1620.7 & 8445.9 \\
Qwen3 & 4 & 81.90 $\pm$ 3.00 & 91.57 $\pm$ 2.00 & 0.698 & 3091.2 & 1623.1 & 8489.9 \\
Qwen3 & 8 & 82.10 $\pm$ 3.17 & 91.58 $\pm$ 2.13 & 0.704 & 3027.8 & 1584.6 & 8260.3 \\
Qwen3 & 14 & \textbf{\underline{86.52}} $\pm$ 2.94 & 99.27 $\pm$ 2.29 & 0.727 & 3607.6 & 1941.2 & 10556.1 \\
Qwen3 & 32 & 84.13 $\pm$ 2.80 & 93.05 $\pm$ 3.13 & 0.668 & 2845.9 & 1497.5 & 7790.1 \\
\midrule
\rowcolor{gray!10}
\multicolumn{8}{c}{\textbf{\textit{Phi Family}}} \\
\midrule
Phi-4 & 14 & \textbf{\underline{78.92}} $\pm$ 3.27 & 97.46 $\pm$ 1.06 & 0.863 & 378.6 & 194.6 & 989.9 \\
Phi-4-mini (I) & 3.8 & 54.55 $\pm$ 3.80 & 95.02 $\pm$ 1.23 & 0.697 & 292.1 & 146.6 & 684.9 \\
Phi-4-reasoning-plus & 14 & 69.54 $\pm$ 3.50 & 88.89 $\pm$ 1.80 & 0.234 & 6780.7 & 3972 & 23893 \\
Phi-4-reasoning & 14 & \textbf{72.23} $\pm$ 3.30 & 96.21 $\pm$ 1.50 & 0.352 & 6066.2 & 3710.8 & 23866.8 \\
Phi-4-mini-reasoning & 3.8 & 70.16 $\pm$ 3.40 & 89.56 $\pm$ 1.78 & 0.646 & 3171.9 & 1571.7 & 8450.5 \\
Phi-3-mini-128k (I) & 3.8 & 35.82 $\pm$ 3.74 & 96.58 $\pm$ 0.79 & 0.566 & 89.4 & 40.6 & 208.9 \\
Phi-3-medium-4k (I) & 14 & 43.47 $\pm$ 4.05 & 89.87 $\pm$ 2.27 & 0.543 & 189.3 & 109.6 & 553.6 \\
Phi-3-medium-128k (I) & 14 & 40.76 $\pm$ 4.10 & 96.26 $\pm$ 1.27 & 0.568 & 140 & 74.8 & 367.3 \\
\midrule
\rowcolor{gray!10}
\multicolumn{8}{c}{\textbf{\textit{Llama Family}}} \\
\midrule
Llama-3.2 (I) & 1 & 16.25 $\pm$ 2.17 & 47.15 $\pm$ 3.97 & 0.278 & 336.3 & 159 & 756.9 \\
Llama-3.2 (I) & 3 & 42.54 $\pm$ 3.49 & 89.88 $\pm$ 1.73 & 0.591 & 279.7 & 144.6 & 694.7 \\
Llama-3.1 (I) & 8 & 48.84 $\pm$ 3.63 & 85.66 $\pm$ 2.53 & 0.646 & 366.4 & 203.4 & 977.7 \\
Llama-3.1 (I) & 70 & \textbf{\underline{75.43}} $\pm$ 3.49 & 98.12 $\pm$ 4.06 & 0.848 & 251.2 & 135.5 & 654.6 \\
Llama-3.3 (I) & 70 & \textbf{74.59} $\pm$ 3.44 & 97.40 $\pm$ 5.80 & 0.840 & 312.8 & 174.1 & 859.7 \\
\midrule
\rowcolor{gray!10}
\multicolumn{8}{c}{\textbf{\textit{Qwen2.5 Family}}} \\
\midrule
Qwen2.5 (I) & 0.5 & 21.31 $\pm$ 2.35 & 77.57 $\pm$ 2.75 & 0.348 & 432.3 & 223.2 & 1144.5 \\
Qwen2.5 (I) & 1.5 & 43.03 $\pm$ 3.48 & 85.45 $\pm$ 2.26 & 0.596 & 264.7 & 134.1 & 626.7 \\
Qwen2.5 (I) & 3 & 45.75 $\pm$ 3.50 & 92.35 $\pm$ 1.09 & 0.619 & 331.3 & 176.5 & 861.4 \\
Qwen2.5 (I) & 7 & 61.36 $\pm$ 3.99 & 96.47 $\pm$ 0.99 & 0.750 & 286.9 & 149.5 & 747.2 \\
Qwen2.5 (I) & 14 & 63.74 $\pm$ 3.82 & 97.83 $\pm$ 0.63 & 0.769 & 260.2 & 137.1 & 685.7 \\
Qwen2.5 (I) & 32 & \textbf{72.90} $\pm$ 3.77 & 99.26 $\pm$ 2.69 & 0.832 & 260.9 & 139.1 & 673.6 \\
Qwen2.5 (I) & 72 & \textbf{\underline{74.87}} $\pm$ 3.60 & 97.12 $\pm$ 5.00 & 0.840 & 339.2 & 179.8 & 887.4 \\
\midrule
\rowcolor{gray!10}
\multicolumn{8}{c}{\textbf{\textit{Other Open-Source Models}}} \\
\midrule
SmolLM2 (I) & 1.7 & 16.69 $\pm$ 2.40 & 68.98 $\pm$ 3.19 & 0.285 & 213 & 93.5 & 481.5 \\
Mistral-7B (I) & 7 & 27.66 $\pm$ 3.09 & 96.26 $\pm$ 0.72 & 0.431 & 207.1 & 113.7 & 585.9 \\
Mistral-Nemo (I) & 12 & 35.43 $\pm$ 3.40 & 82.95 $\pm$ 2.94 & 0.517 & 377 & 234.2 & 1123.7 \\
Qwen2.5-Math (I) & 1.5 & \textbf{51.43} $\pm$ 3.97 & 94.04 $\pm$ 1.61 & 0.656 & 397.1 & 210 & 1076.9 \\
Qwen2.5-Math (I) & 7 & \textbf{\underline{60.68}} $\pm$ 4.01 & 94.36 $\pm$ 1.75 & 0.740 & 411.7 & 221.5 & 1156 \\
\midrule
\rowcolor{gray!10}
\multicolumn{8}{c}{\textbf{\textit{OpenAI Models}}} \\
\midrule
GPT-4.1 & -- & 89.88 $\pm$ 2.50 & 97.79 $\pm$ 1.80 & 0.927 & 338.8 & 152.2 & 759.5 \\
GPT-4.1-mini & -- & \textbf{\underline{90.23}} $\pm$ 2.30 & 98.14 $\pm$ 1.50 & 0.930 & 328.8 & 145.1 & 740.7 \\
GPT-4.1-nano & -- & 75.35 $\pm$ 3.20 & 95.58 $\pm$ 2.10 & 0.843 & 338.8 & 148.7 & 760.3 \\
GPT-4o & -- & \textbf{87.56} $\pm$ 2.40 & 99.42 $\pm$ 1.20 & 0.918 & 290.5 & 154.8 & 749 \\
GPT-4o-mini & -- & 75.00 $\pm$ 2.80 & 97.67 $\pm$ 1.90 & 0.841 & 341.3 & 172 & 848.7 \\
\midrule
\rowcolor{gray!10}
\multicolumn{8}{c}{\textbf{\textit{Google Models}}} \\
\midrule
Gemini-2.0-flash-lite & -- & \textbf{\underline{73.33}} $\pm$ 2.90 & 99.58 $\pm$ 1.10 & 0.836 & 215.5 & 116.9 & 523 \\
Gemini-2.0-flash & -- & \textbf{69.60} $\pm$ 3.10 & 94.44 $\pm$ 2.00 & 0.811 & 234.5 & 118.9 & 517.9 \\
Gemini-2.5-flash & -- & 55.18 $\pm$ 3.80 & 63.49 $\pm$ 2.70 & 0.705 & 186.3 & 104.8 & 539.5 \\
\bottomrule
\end{tabulary}
\end{adjustbox}
\caption{Comprehensive performance evaluation of 53 language models on basic mathematical reasoning tasks, including base, instruction-tuned, reasoning, and quantized variants (full quantization results in Appendix~\ref{app:extended_results}; this table shows FP16/default precision). Within each model family, \textbf{\underline{bold+underline}} indicates best accuracy and \textbf{bold} indicates second-best accuracy. All metrics represent averages across 14 tasks with 1,000 samples each (open-source) or 100 samples (closed-source). Instruct-tuned models are marked with (I).}
\label{tab:comprehensive_model_performance}
\vspace{-1em}
\end{table*}

\subsection{\textsc{Scaling Effects: Non-Monotonic and Diminishing Returns}}
\label{sec:scaling_effects}

Figure~\ref{fig:combined}(a) shows scaling behavior across three model families, revealing that parameter count does not guarantee better basic math performance. Within the \textit{Qwen3} family we observe \textbf{non-monotonic scaling}: \textbf{Qwen3-14B = 86.52\%} vs \textbf{32B = 84.13\%}-a decline of \(\mathbf{2.39}\)\% despite 2.3$\times$ more parameters. By contrast, smaller scale steps yield large gains: \textbf{Qwen3-0.6B = 49.99\%} \(\rightarrow\) \textbf{1.7B = 70.24\%} (\(\mathbf{+20.25}\) points, \(\mathbf{+40.51\%}\) relative). The \textit{Qwen2.5} family shows monotonic but sublinear improvement: \textbf{Qwen2.5-14B = 63.74\%} \(\rightarrow\) \textbf{32B = 72.90\%} (\(+\mathbf{9.16}\) points), indicating that scaling helps in some regimes but with diminishing returns. 
\begin{tcolorbox}[
    colback=teal!5!white,
    colframe=teal!60!black,
    title=\textbf{\textit{Finding 2 --- Non-Monotonic Scaling}},
    fonttitle=\bfseries,
    boxrule=0.7pt,
    arc=1mm
]
\textbf{Parameter scaling shows diminishing and sometimes negative returns.} Larger models do not guarantee better performance, with mid-sized variants often achieving superior efficiency-accuracy balance.
\end{tcolorbox}
The \textit{Llama} family shows large jumps between small and large models with significant variance across parameter counts. The vertical dotted line at 7B marks where performance curves begin to flatten, showing diminishing returns from further scaling. As models scale they can accumulate competing heuristics: some improve benchmark-style problems while others interfere on low-level operations, producing the non-monotonic scaling we observe. Notably, \textbf{GPT-4.1-mini = 90.23\%} slightly edges \textbf{GPT-4.1 = 89.88\%}, showing smaller/optimized variants can be practically preferable.

\begin{figure*}[t]
    \centering
    \scriptsize
    \setlength{\tabcolsep}{1pt}
    \begin{minipage}[t]{0.24\textwidth}
    \begin{tcolorbox}[colback=lightred!30, colframe=darkred, title=\textbf{(a) Extreme Overthinking}, fonttitle=\bfseries\scriptsize, height=5.5cm, top=1mm, bottom=1mm, left=1mm, right=1mm]
    \textbf{Task:} $846 \div 18$

    \textbf{Phi-4 (47 tok):}
    \begin{verbatim}
846/18=47 \boxed{47}
    \end{verbatim}
    \vspace{-2mm}

    \textbf{Phi-4-reas (1,456 tok):}
    \begin{verbatim}
Long div...
V1: 47x18=846
V2: factorization
  846=2x3x3x47
  18=2x3x3
V3: estimation
  846~900, 18~20
  900/20=45
\boxed{47}
    \end{verbatim}
    \vspace{-1.8mm}
    \textbf{Result:} 31x, no gain
    \end{tcolorbox}
    \end{minipage}%
    \hfill
    \begin{minipage}[t]{0.24\textwidth}
    \begin{tcolorbox}[colback=lightyellow!50, colframe=darkyellow, title=\textbf{(b) Pathological Failure}, fonttitle=\bfseries\scriptsize, height=5.5cm, top=1mm, bottom=1mm, left=1mm, right=1mm]
    \textbf{Task:} Mode([5,5,5,7,7,9])

    \textbf{Expected:} 5

    \textbf{Qwen3-4B (1,024 tok):}
    \begin{verbatim}
Frequency:
5: 3 times
7: 2 times
9: 1 time

Mode=5nnnnnnnnnnn
nnnnnnnnnnnnnnnnn
nnnnnnnnnnnnnnnnn
nnnnnnnnnnnnnnnnn
nnnnnnnnnnnnnnnnn
[... 900+ 'n' ...]
    \end{verbatim}
    \vspace{-1mm}
    \textbf{Result:} Stopping failure
    \end{tcolorbox}
    \end{minipage}%
    \hfill
    \begin{minipage}[t]{0.24\textwidth}
    \begin{tcolorbox}[colback=lightgreen!30, colframe=darkgreen, title=\textbf{(c) Helpful Reasoning}, fonttitle=\bfseries\scriptsize, height=5.5cm, top=1mm, bottom=1mm, left=1mm, right=1mm]
    \textbf{Task:} Sort 8 numbers

    \textbf{Phi-4-reas (312 tok):}
    \begin{verbatim}
1: Separate
  Neg:-789,-456,-123
  Pos:847,456,234,
      901,123
2: Sort negs
  -789<-456<-123
3: Sort pos
  123<234<456<
  847<901
4: Combine
\boxed{[-789,-456,
-123,123,234,...]}
    \end{verbatim}
    \vspace{-1mm}
    \textbf{Result:} 3.5x justified
    \end{tcolorbox}
    \end{minipage}%
    \hfill
    \begin{minipage}[t]{0.24\textwidth}
    \begin{tcolorbox}[colback=lightred!30, colframe=darkred, title=\textbf{(d) Self-Contradiction}, fonttitle=\bfseries\scriptsize, height=5.5cm, top=1mm, bottom=1mm, left=1mm, right=1mm]
    \textbf{Task:} Max([12,45,23,89,34])

    \textbf{O3-mini (2,400 tok):}
    \begin{verbatim}
Scan: max=89

Wait, reconsider...
is 89 largest?
Maybe 45... no 89>45

Verify:
89>12? Yes
89>45? Yes...
Duplicates? None.
Recalculate...
[1000+ tokens]
\boxed{89}
    \end{verbatim}
    \vspace{-3.4mm}
    \textbf{Result:} Self-doubt loop
    \end{tcolorbox}
    \end{minipage}
    \vspace{-1mm}
    \caption{\textbf{Reasoning Pattern Case Studies:} \textbf{(a)} Extreme overthinking with 31x token waste through 3x verification. \textbf{(b)} Pathological failure showing stopping mechanism breakdown ($\infty$ character repetition). \textbf{(c)} Helpful reasoning where 3.5x verbosity provides verifiable steps. \textbf{(d)} Self-contradiction loops. Complete case study in Appendix~\ref{app:case_studies}.}
    \label{fig:case_study_visual}
\end{figure*}

\begin{figure*}[ht]
    \centering
    \includegraphics[width=1\textwidth]{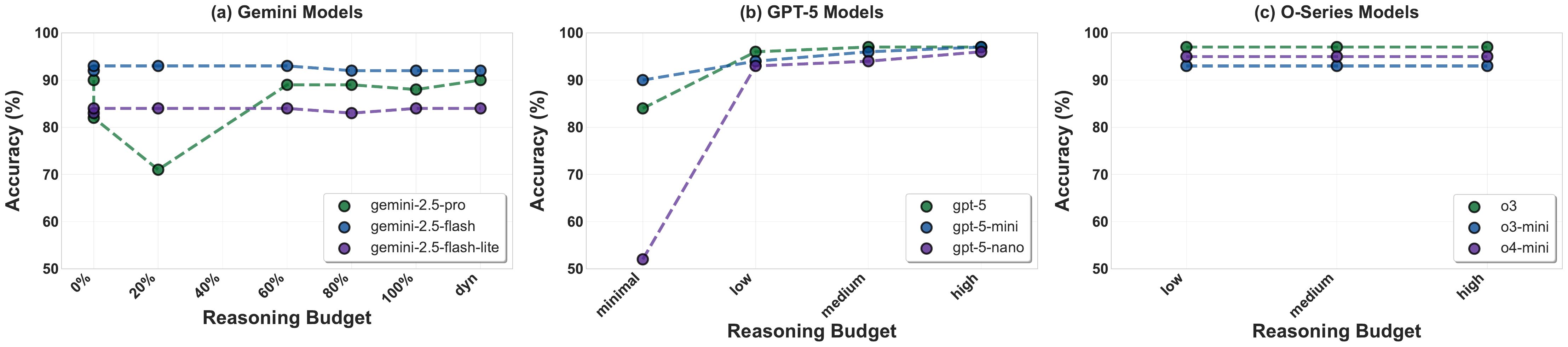}
    \caption{\textbf{Reasoning budget analysis across Gemini, GPT-5, and O-series models.} Increasing the reasoning budget yields minimal gains, showing diminishing returns and near-plateau performance at higher effort levels.}
    \label{fig:reasoning_budget}
\end{figure*} 

\subsection{\textsc{Overthinking: Verbosity Hurts Correctness}}

As shown in Table~\ref{tab:comprehensive_model_performance}, models trained or fine-tuned to produce long chain-of-thoughts (COT) often generate more tokens yet obtain \emph{lower} accuracy than concise variants: for example, \textbf{Phi-4} scores \textbf{78.92\%} with \(\sim378.6\) tokens on average, while \textbf{Phi-4-reasoning} scores \textbf{72.23\%} with \(\sim6066.2\) tokens: an accuracy gap of \textbf{6.69} points (\(\approx\)\textbf{-8.48\%} relative) despite using \(\sim16.02\times\) more tokens. When constrained to 1,024 tokens, Phi-4-reasoning falls to \textbf{53.48\%} (an absolute loss of \textbf{18.75} points, \(\approx\)\textbf{-25.96\%} relative from 72.23\%), and Phi-4-reasoning-plus drops from \textbf{69.54\%} to \textbf{44.33\%} (absolute \textbf{-25.21\%}, \(\approx\)\textbf{-36.25\%} relative).
\begin{tcolorbox}[
    colback=teal!5!white,
    colframe=teal!60!black,
    title=\textbf{\textit{Finding 3 --- Overthinking}},
    fonttitle=\bfseries,
    boxrule=0.7pt,
    arc=1mm
]
\textbf{Reasoning models produce long, repetitive explanations that sometimes reduce accuracy.} Models tend to produce reasoning-like text rather than performing correct computation.
\end{tcolorbox}
Even under constraint, Phi-4-reasoning uses \(\sim1013.5\) tokens on average, which is about \(\mathbf{2.68\times}\) the token usage of concise Phi-4. This pattern suggests COT supervision shifts the optimization target toward producing plausible intermediate text rather than verifiable intermediate computations, i.e., the model learns to generate \emph{plausible-looking exploration} that increases likelihood but is not reliably grounded in correct arithmetic, which explains why penalizing verbosity or enforcing step-level checks often recovers accuracy. To understand \textit{which reasoning patterns help versus hurt}, Table~\ref{tab:case_study_compact} compares Phi-4 and Phi-4-reasoning on identical tasks, while Figure~\ref{fig:case_study_visual} presents actual model outputs (detailed case studies in Appendix~\ref{app:case_studies}).

\subsection{\textsc{Performance Under Token Constraints}}

When we restrict output length, non-reasoning models remain relatively stable while reasoning models suffer sharp, sometimes catastrophic drops. Figure~\ref{fig:combined}(b) visualizes this phenomenon for the Qwen3 family, showing dramatic accuracy degradation when generation is constrained to 1,024 tokens. Their accuracy follows a steep curve: below certain token budgets (e.g., $\sim$512 tokens) performance collapses; between 512–1024 tokens results are unstable; above large budgets they improve but still lag concise models. For example, \textbf{Phi-4} achieves \(\mathbf{78.92\%}\) with \(\sim378.6\) tokens on average, whereas \textbf{Phi-4-reasoning} scores \(\mathbf{72.23\%}\) while using \(\sim6066.2\) tokens ($\approx$\(\mathbf{16\times}\) more); constraining Phi-4-reasoning to a 1,024-token budget reduces its accuracy to \(\mathbf{53.48\%}\) (an \(\mathbf{18.75}\) point absolute loss, $\approx$\(\mathbf{26\%}\) relative), and Phi-4-reasoning-plus shows a similar, even larger collapse when constrained (see Appendix~\ref{app:constrained_results} for complete results). Allowing the long-form reasoning models thousands of extra tokens recovers some accuracy but at a very small marginal return and often still leaves these models behind concise baselines; reasoning models appear to rely on a fixed, large exploration budget rather than adapting computation to problem difficulty.

\begin{tcolorbox}[
    colback=teal!5!white,
    colframe=teal!60!black,
    title=\textbf{\textit{Finding 4 --- Efficiency Cliff}},
    fonttitle=\bfseries,
    boxrule=0.7pt,
    arc=1mm
]
\textbf{Constraining tokens causes sudden accuracy collapses,} showing models lack adaptive stopping and rely on long exploration.
\end{tcolorbox}

\begin{table}[ht]
\centering
\scriptsize
\begin{adjustbox}{max width=\linewidth}
\setlength{\tabcolsep}{3pt}
\begin{tabular}{lccccl}
\toprule
\textbf{Task} & \textbf{Model} & \textbf{Correct?} & \textbf{Tokens} & \textbf{Token Ratio} & \textbf{Reasoning Pattern} \\
\midrule
\rowcolor{lightgreen}
Sum (8 nums) & Phi-4 & \cmark & 272 & 1× & Direct step-by-step \\
\rowcolor{lightred}
 & Phi-4-reasoning & \cmark & 4,837 & 17.8× & Redundant verification, wasteful \\
\midrule
\rowcolor{lightgreen}
Complex Sort & Phi-4 & \cmark & 89 & 1× & Direct output \\
\rowcolor{lightyellow}
 & Phi-4-reasoning & \cmark & 312 & 3.5× & Helpful decomposition (justified) \\
\midrule
\rowcolor{lightgreen}
Division & Phi-4 & \cmark & 47 & 1× & Minimal steps \\
\rowcolor{lightred}
 & Phi-4-reasoning & \cmark & 1,456 & 31× & Triple verification, no benefit \\
\midrule
\rowcolor{lightgreen}
Multiplication & Phi-4 & \cmark & 156 & 1× & Sequential multiply \\
\rowcolor{lightred}
 & Phi-4-reasoning & \xmark & 3,214 & 20.6× & Over-analyzed, arithmetic error \\
\midrule
\rowcolor{lightgreen}
Mean calc & Phi-4 & \cmark & 198 & 1× & Sum then divide \\
\rowcolor{lightred}
 & Phi-4-reasoning & \xmark & 2,891 & 14.6× & Lost in verification, wrong final \\
\bottomrule
\end{tabular}
\end{adjustbox}
\caption{\textbf{Correct and incorrect reasoning outcomes.} \colorbox{lightgreen}{Green}: Efficient and correct. \colorbox{lightyellow}{Yellow}: Helpful structured reasoning where verbosity aids correctness. \colorbox{lightred}{Red}: Wasteful overthinking: either no benefit despite high token cost, or incorrect answers due to over-verification.}
\label{tab:case_study_compact}
\vspace{-1.5em}
\end{table}

\subsection{\textsc{Thinking Budget: Diminishing Returns and Error Accumulation}}

\textbf{Allocating larger reasoning budgets produces minimal accuracy improvements while introducing systematic inefficiencies.} Figure~\ref{fig:reasoning_budget} illustrates this across three model families. For Gemini models, Gemini-2.5-Flash gains only \textbf{1\%} from the disabled baseline (92\%$\to$93\%) and returns to 92\% at the maximum budget, while Gemini-2.5-Pro peaks at its lowest tested budget (128 tokens, 90\%) with no gain at higher budgets. More strikingly, GPT-5 achieves identical \textbf{97\% accuracy} on both medium and high effort levels despite using progressively more tokens, showing \textbf{zero} marginal benefit from extended reasoning. The O-series models (O3, O3-mini, O4-mini) exhibit similar plateaus: O3 maintains \textbf{97\% accuracy} across all effort levels (low, medium, high) while generating $\sim$8\% more tokens at higher budgets. This pattern reveals models lack \textit{adaptive computation}: they cannot modulate reasoning depth based on problem difficulty. Instead, they apply fixed exploration patterns regardless of whether a task requires simple arithmetic or multi-step verification, resulting in systematic waste on easy problems and insufficient adaptation on hard ones. Empirically, the accuracy curve follows \(A(b)=A_{\max}(1-e^{-b/\tau})+\epsilon(b)\), where $\tau\!\approx\!1000$ tokens captures saturation scale and $\epsilon(b)$ represents error accumulation. Beyond moderate budgets (1,000–3,000 tokens), models often degrade by contradicting earlier reasoning, overwriting correct steps with plausible but incorrect ones; extended generation sometimes amplifies such errors (see Table~\ref{tab:aggregated_results} in Appendix~\ref{app:reasoning_budget}).

\begin{tcolorbox}[
    colback=teal!5!white,
    colframe=teal!60!black,
    title=\textbf{\textit{Finding 5 --- Thinking Budget}},
    fonttitle=\bfseries,
    boxrule=0.7pt,
    arc=1mm
]
\textbf{Extra thinking token budget gives small returns and increases the risk of contradictory or noisy reasoning.}
\end{tcolorbox}

\subsection{\textsc{Quantization Robustness}}

Quantization shows a clear size-dependent pattern illustrated in Figure~\ref{fig:combined}(c): larger models tolerate aggressive compression with minimal accuracy loss, while very small models suffer large relative drops; for instance, the Qwen2.5-32B model moves from \(\mathbf{73.08\%}\) (8-bit) to \(\mathbf{72.67\%}\) (4-bit), a negligible change, whereas Qwen2.5-0.5B falls from \(\mathbf{21.31\%}\) to \(\mathbf{12.77\%}\) under 4-bit quantization ($\approx$\(\mathbf{40\%}\) relative loss). Mid-sized models show modest penalties (e.g., \(\sim\!3\%\!-\!5\%\) relative in several 14B examples), which implies that redundancy in large models preserves low-precision arithmetic behavior and that the primary issues we observe (overthinking, token brittleness) are behavioral and not simply precision-limited. Complete quantization results are presented in Appendix~\ref{app:quantization_results}.

\begin{tcolorbox}[
    colback=teal!5!white,
    colframe=teal!60!black,
    title=\textbf{\textit{Finding 6 --- Quantization Robustness}},
    fonttitle=\bfseries,
    boxrule=0.7pt,
    arc=1mm
]
\textbf{Even aggressive quantization has minimal effect on large models, whereas small models suffer significant performance degradation.}
\end{tcolorbox}

\subsection{\textsc{Insights from the Overthinking Score Metric}}
As shown in Table \ref{tab:comprehensive_model_performance}, when we combine accuracy and efficiency into the Overthinking Score, model rankings change: some smaller or mid-sized instruction-tuned models beat larger, verbose ones. The score highlights a sweet-spot around mid-large sizes (roughly 14--20B) where models have enough capacity for correct computation without developing overthinking habits. For example, \textbf{GPT-4.1-mini} scores \(\mathbf{0.930}\) vs \textbf{GPT-4.1} at \(\mathbf{0.927}\) (an absolute \(\mathbf{0.003}\) improvement, $\approx$\(\mathbf{0.3\%}\) relative), while \textbf{Qwen3-14B} (score \(\mathbf{0.727}\)) outperforms Qwen3-32B (\(\mathbf{0.668}\)) by \(\mathbf{0.059}\) points ($\approx$\(\mathbf{8.8\%}\) relative); conversely, reasoning-tuned variants suffer large penalties on this metric (for example, Phi-4-reasoning at \(\mathbf{0.352}\) vs Phi-4 at \(\mathbf{0.863}\), a \(\sim\!59\%\) relative deficit), showing that efficiency-aware evaluation identifies different, deployment-relevant winners than accuracy-only comparisons. The inverted-U with size suggests two things: scale gives models the ability to store and execute correct computational patterns, but excessive scale (without appropriate inductive bias) enables learned behaviors that favor verbose, high-likelihood text. Instruction tuning toward concise, correct answers often shifts models back toward efficiency.

\begin{tcolorbox}[
    colback=teal!5!white,
    colframe=teal!60!black,
    title=\textbf{\textit{Finding 7 --- Overthinking Score Insights}},
    fonttitle=\bfseries,
    boxrule=0.7pt,
    arc=1mm
]
\textbf{When efficiency is included, mid-sized instruction-tuned models often become the most practical choices.}
\end{tcolorbox}

\subsection{\textsc{Fine-Grained Analysis of Reasoning Content}}
\label{sec:fine_grained}

\textbf{Long traces are dominated by four wasteful patterns, not random verbosity.} We hand-annotated $\sim$5{,}000 responses from 12 reasoning and 12 standard models and found four patterns that cover most long traces plus one that is genuinely helpful: redundant verification loops, self-contradiction loops, irrelevant exploration, pathological stopping failures ($<$2\% of traces; Appendix~\ref{app:pathological}), and helpful decomposition ($\sim$11\%, where extra tokens contribute verifiable sub-steps). Table~\ref{tab:case_study_compact} and Figure~\ref{fig:case_study_visual} show one example per pattern, and the full taxonomy with per-family counts is in Appendix~\ref{app:pattern_taxonomy}. The wasteful patterns are behavioral rather than capability gaps, and they concentrate in CoT-tuned variants, which is why the O-Score penalizes these variants even when raw accuracy is competitive.

\begin{tcolorbox}[
    colback=teal!5!white,
    colframe=teal!60!black,
    title=\textbf{\textit{Finding 8 --- Pattern Taxonomy}},
    fonttitle=\bfseries,
    boxrule=0.7pt,
    arc=1mm
]
\textbf{Not all tokens are equal.} Most wasted tokens fall into four recurring failure modes rather than random verbosity, and the patterns are concentrated in CoT-tuned models.
\end{tcolorbox}

\subsection{\textsc{Root Causes and Mitigation}}
\label{sec:root_causes}

\textbf{CoT supervision is the main driver, and prompting or tools do not close the gap.} Matched pairs isolate the effect of training: Qwen2.5-14B-instruct averages 260 tokens while Qwen3-14B uses 3{,}608 (13.9$\times$), and Phi-4 averages 379 tokens while Phi-4-reasoning uses 6{,}066 (16.0$\times$) with accuracy dropping from 78.92\% to 72.23\%; the pattern holds across four matched pairs (Appendix~\ref{app:root_cause}). Concise prompts trim Phi-4-reasoning's tokens by $-$63\% but lose accuracy and still use 5.9$\times$ more tokens than Phi-4 (Appendix~\ref{app:concise_prompting}), and adding a calculator, Python REPL, or code executor raises accuracy by 2--30 points but leaves token overhead at 1.3--2.6$\times$ and does not stop accuracy from collapsing as problem size grows (Appendix~\ref{app:tool_augmented}). GPT-5, O3, and O4-mini hit their top accuracy at the \textit{low} effort setting and stay flat through medium and high while token usage keeps climbing, a signature consistent with process-reward maximization, so architectural routing looks more promising than prompt-level tweaks (leave-one-out ranking stability: Kendall's $\tau = 0.87$, Spearman $\rho = 0.92$; Appendix~\ref{app:stability}).

\begin{tcolorbox}[
    colback=teal!5!white,
    colframe=teal!60!black,
    title=\textbf{\textit{Finding 9 --- Root Causes}},
    fonttitle=\bfseries,
    boxrule=0.7pt,
    arc=1mm
]
\textbf{CoT supervision drives overthinking; concise prompts and tool use reduce but do not close the gap.}
\end{tcolorbox}

\section{Conclusion}
\label{sec:conclusion_short}
Our evaluation of 53 LLMs through \llmthinkbench\ reveals: \textbf{\textit{1)}} Parameter scaling gives diminishing and non-monotonic returns: mid-sized models (14-20B) often match or outperform larger siblings on accuracy-efficiency tradeoffs. \textbf{\textit{2)}} ``Thinking'' supervision often fails: models trained for long chains typically generate 18$\times$ more tokens while losing accuracy and collapsing under token constraints. \textbf{\textit{3)}} Extended reasoning budgets show diminishing returns and error accumulation, revealing models lack adaptive stopping and step-level verification. \textbf{\textit{4)}} Wasted tokens are not random; four recurring patterns (redundant verification, self-contradiction, irrelevant exploration, stopping failures) account for most long traces, with CoT supervision as the main driver. \textbf{\textit{5)}} Concise prompting and tool use reduce but do not remove the gap, suggesting architectural routing is the right direction.

\paragraph{\textsc{Practical Recommendations}} \textbf{Models:} GPT-4.1-mini (O-Score: 0.930), Qwen3-14B (0.727), Qwen2.5-7B for budget deployments. \textbf{Prompting:} Using formats (\textbackslash boxed\{\}), applying token budgets strategically, avoiding open-ended ``think step-by-step'' on basic tasks. All code, data generators, and the leaderboard are public.


\clearpage

\section*{Acknowledgements}
This work was supported by the NSF \#2442253, NSF NAIRR Pilot with PSC Neocortex and NCSA Delta, Cisco Research, NVIDIA, Amazon, the Commonwealth Cyber Initiative, the Amazon–Virginia Tech Center for Efficient and Robust Machine Learning, the Sanghani Center for AI and Data Analytics at Virginia Tech, and the Virginia Tech Innovation Campus. The views, findings, conclusions, and recommendations expressed in this work are those of the authors and do not necessarily reflect the opinions of the funding agencies.

\section*{Limitations}
Our study has a few limitations. \textbf{First,} we focus on basic math operations, which do not cover the full range of reasoning needed in real-world problem solving; these tasks expose one dimension of mathematical behavior. \textbf{Second,} our automated evaluation enables large-scale assessment across 53 models but does not replace the kind of interpretability human evaluation would add, and qualitative analysis of reasoning patterns is a natural complement. \textbf{Third,} our dynamic test generation prevents direct memorization, but models can still use statistical patterns from training; how much they genuinely compute versus pattern-match is an open question. \textbf{Fourth,} our evaluation relies on task-specific answer parsers that must track changes in model output formats and can introduce extraction bias.

\section*{Ethics Statement}
This work presents a systematic benchmark for assessing mathematical reasoning and overthinking tradeoffs in language models and does not generate harmful content or introduce direct ethical risks. Our evaluation uses publicly available models on deterministic mathematical tasks, requiring no collection of personal data or human annotations that could raise privacy concerns. All models are evaluated in accordance with their respective licensing agreements and usage policies.

\bibliography{custom}

\appendix

\clearpage

\onecolumn

\addtocontents{toc}{\protect\setcounter{tocdepth}{2}}

\renewcommand{\contentsname}{Contents of the Appendix}

\tableofcontents

\appendix

\section{Overthinking Score: Complete Theoretical Framework}
\label{app:theoretical}

This section provides a comprehensive theoretical analysis of the Overthinking Score, including rigorous proofs of its mathematical properties, sensitivity analysis, and justification for our choice of the harmonic mean formulation.

\subsection{Fundamental Properties and Proofs}

We provide complete proofs for the properties of the Overthinking Score $\mathcal{O}_i = \frac{2 \cdot A_i \cdot E_{t,i}}{A_i + E_{t,i}}$.

\begin{theorem}[Complete Properties of Overthinking Score]
The Overthinking Score $\mathcal{O}_i$ satisfies:
\begin{enumerate}
    \item \textbf{Boundedness:} $\mathcal{O}_i \in [0, 1]$ for all valid inputs
    \item \textbf{Symmetry:} $\mathcal{O}(A, E) = \mathcal{O}(E, A)$
    \item \textbf{Vanishing Limits:} $\lim_{E_t \to 0} \mathcal{O} = 0$ and $\lim_{A \to 0} \mathcal{O} = 0$
    \item \textbf{Monotonicity:} $\mathcal{O}$ is strictly increasing in both arguments
    \item \textbf{Sublinearity:} $\mathcal{O}(A, E) \leq \min(A, E)$
    \item \textbf{Perfection Condition:} $\mathcal{O} = 1 \iff A = E = 1$
\end{enumerate}
\end{theorem}

\begin{proof}
\textbf{Property 1 (Boundedness):} Since $A, E \in [0, 1]$, we have:
$$\mathcal{O} = \frac{2AE}{A + E} \leq \frac{2AE}{2\sqrt{AE}} = \sqrt{AE} \leq 1$$
where we used the AM-GM inequality. The lower bound follows from non-negativity of $A$ and $E$.

\textbf{Property 2 (Symmetry):} Direct from the commutative property:
$$\mathcal{O}(A, E) = \frac{2AE}{A + E} = \frac{2EA}{E + A} = \mathcal{O}(E, A)$$

\textbf{Property 3 (Vanishing Limits):} As $E_t \to 0$:
$$\mathcal{O} = \frac{2AE_t}{A + E_t} = \frac{2A}{A/E_t + 1}$$
Since $A/E_t \to \infty$ as $E_t \to 0$, we have $\mathcal{O} \to 0$. The proof for $A \to 0$ is symmetric.

\textbf{Property 4 (Monotonicity):} Taking the partial derivative with respect to $A$:
$$\frac{\partial \mathcal{O}}{\partial A} = \frac{2E^2}{(A + E)^2} > 0 \text{ for } E > 0$$
Similarly, $\frac{\partial \mathcal{O}}{\partial E} = \frac{2A^2}{(A + E)^2} > 0$ for $A > 0$.

\textbf{Property 5 (Sublinearity):} By the harmonic-minimum inequality:
\begin{align}
\mathcal{O} &= \frac{2AE}{A + E} \\
&\leq \frac{2 \min(A, E) \cdot \min(A, E)}{\min(A, E) + \min(A, E)} \\
&= \min(A, E)
\end{align}

\textbf{Property 6 (Perfection Condition):} If $\mathcal{O} = 1$, then:
\begin{align}
\frac{2AE}{A + E} = 1 &\implies 2AE = A + E \\
&\implies (A - E)^2 = 0
\end{align}
Therefore $A = E$. Substituting back: $\mathcal{O} = \frac{2A^2}{2A} = A = 1$, so $A = E = 1$.

Conversely, if $A = E = 1$, then $\mathcal{O} = \frac{2 \cdot 1 \cdot 1}{1 + 1} = 1$.
\end{proof}

\subsection{Sensitivity Analysis and Derivatives}

\begin{theorem}[Sensitivity Properties]
The Overthinking Score exhibits the following sensitivity characteristics:
\begin{align}
\frac{\partial \mathcal{O}}{\partial A} &= \frac{2E^2}{(A + E)^2}\\
\frac{\partial \mathcal{O}}{\partial E} &= \frac{2A^2}{(A + E)^2}\\
\frac{\partial^2 \mathcal{O}}{\partial A \partial E} &= \frac{4AE}{(A + E)^3}
\end{align}
\end{theorem}

\begin{proof}
Starting with $\mathcal{O} = \frac{2AE}{A + E}$, we apply the quotient rule:

For $\frac{\partial \mathcal{O}}{\partial A}$:
\begin{align}
\frac{\partial \mathcal{O}}{\partial A} &= \frac{2E(A + E) - 2AE}{(A + E)^2} \\
&= \frac{2E^2}{(A + E)^2}
\end{align}

For $\frac{\partial \mathcal{O}}{\partial E}$:
\begin{align}
\frac{\partial \mathcal{O}}{\partial E} &= \frac{2A(A + E) - 2AE}{(A + E)^2} \\
&= \frac{2A^2}{(A + E)^2}
\end{align}

For the cross-partial:
\begin{align}
\frac{\partial^2 \mathcal{O}}{\partial A \partial E} &= \frac{\partial}{\partial A}\left(\frac{2A^2}{(A + E)^2}\right) \\
&= \frac{4A(A + E)^2 - 2A^2 \cdot 2(A + E)}{(A + E)^4} \\
&= \frac{4AE}{(A + E)^3}
\end{align}
\end{proof}

\begin{corollary}[Improvement Incentives]
The sensitivity to each component is proportional to the square of the other component, ensuring that improving the weaker dimension yields greater marginal benefit.
\end{corollary}

\subsection{Hessian Analysis and Concavity}

\begin{theorem}[Concavity Properties]
The Hessian matrix of the Overthinking Score:
\begin{equation}
H = \frac{4AE}{(A + E)^3} \begin{bmatrix}
-E & A+E \\
A+E & -A
\end{bmatrix}
\end{equation}
has eigenvalues $\lambda_1 = 0$ and $\lambda_2 = -\frac{4AE}{(A+E)^2} < 0$, confirming concavity along the improvement direction.
\end{theorem}

\begin{proof}
The Hessian matrix elements are:
$$H_{11} = \frac{\partial^2 \mathcal{O}}{\partial A^2} = -\frac{4E^2}{(A + E)^3}$$
$$H_{22} = \frac{\partial^2 \mathcal{O}}{\partial E^2} = -\frac{4A^2}{(A + E)^3}$$
$$H_{12} = H_{21} = \frac{4AE}{(A + E)^3}$$

Factoring out $\frac{4AE}{(A + E)^3}$:
$$H = \frac{4AE}{(A + E)^3} \begin{bmatrix}
-E & A+E \\
A+E & -A
\end{bmatrix}$$

The characteristic polynomial is:
$$\det(H - \lambda I) = \lambda^2 + \lambda \cdot \frac{4AE}{(A+E)^2} = 0$$

This yields eigenvalues $\lambda_1 = 0$ and $\lambda_2 = -\frac{4AE}{(A+E)^2}$.

The eigenvector for $\lambda_1 = 0$ is $(1, 1)^T$, representing the direction of balanced improvement. The negative eigenvalue $\lambda_2$ confirms concavity perpendicular to this direction.
\end{proof}

\subsection{Metric Design Rationale and Comparison}

The Overthinking Score must satisfy several mathematical properties to serve as a meaningful evaluation metric:

\begin{theorem}[Metric Properties]
The Overthinking Score $\mathcal{O}: [0,1]^2 \rightarrow [0,1]$ satisfies:
\begin{enumerate}
    \item \textbf{Symmetry:} $\mathcal{O}(a, e) = \mathcal{O}(e, a)$
    \item \textbf{Monotonicity:} $\mathcal{O}$ is increasing in both arguments
    \item \textbf{Boundary Conditions:}
        \begin{align}
        \mathcal{O}(0, e) &= 0 \quad \forall e \in [0,1] \\
        \mathcal{O}(a, 0) &= 0 \quad \forall a \in [0,1] \\
        \mathcal{O}(1, 1) &= 1
        \end{align}
    \item \textbf{Continuity:} $\mathcal{O}$ is continuous on $[0,1]^2$
    \item \textbf{Concavity:} $\mathcal{O}$ is jointly concave
\end{enumerate}
\end{theorem}

\begin{proof}
For the harmonic mean $\mathcal{O}(a, e) = \frac{2ae}{a+e}$:

\begin{enumerate}
    \item \textbf{Symmetry:} Immediate from commutativity of multiplication and addition.

    \item \textbf{Monotonicity:}
    \begin{equation}
    \frac{\partial \mathcal{O}}{\partial a} = \frac{2e^2}{(a+e)^2} > 0 \text{ for } e > 0
    \end{equation}
    Similarly for $\partial \mathcal{O}/\partial e$.

    \item \textbf{Boundary Conditions:} Direct substitution verifies all three conditions.

    \item \textbf{Continuity:} The function is a ratio of continuous functions with non-zero denominator on $(0,1]^2$, extended continuously to boundaries.

    \item \textbf{Concavity:} The Hessian matrix is:
    \begin{equation}
    H = \frac{4ae}{(a+e)^3} \begin{bmatrix}
    -e & a+e \\
    a+e & -a
    \end{bmatrix}
    \end{equation}
    The eigenvalues are $\lambda_1 = 0$ and $\lambda_2 = -\frac{4ae}{(a+e)^2} \leq 0$, confirming negative semi-definiteness.
\end{enumerate}
\end{proof}

\subsection{Comparison with Alternative Metrics}

We formally compare different aggregation methods:

\begin{proposition}[Penalty for Imbalance]
Define the imbalance penalty as $P(a, e) = M(a, e) - M(a, a)$ where $M$ is a mean function. For $a < e$:
\begin{align}
P_{\text{arith}}(a, e) &= \frac{e - a}{2} \\
P_{\text{geo}}(a, e) &= a(\sqrt{e/a} - 1) \\
P_{\text{harm}}(a, e) &= a\left(1 - \frac{2e}{a + e}\right) = \frac{a(a - e)}{a + e}
\end{align}
\end{proposition}

The harmonic mean penalty grows superlinearly with imbalance, providing stronger incentive for balanced optimization.

\subsection{Efficiency Normalization}

The token efficiency component requires careful normalization:

\begin{definition}[Token Efficiency]
Given token counts $\{t_1, \ldots, t_m\}$ from $m$ models, the efficiency of model $i$ is:
\begin{equation}
E_i = \begin{cases}
1 & \text{if } t_i = t_{\min} \\
0 & \text{if } t_i = t_{\max} \text{ and } t_{\max} > t_{\min} \\
1 - \frac{t_i - t_{\min}}{t_{\max} - t_{\min}} & \text{otherwise}
\end{cases}
\end{equation}
\end{definition}

\begin{lemma}[Efficiency Properties]
The efficiency function satisfies:
\begin{enumerate}
    \item $E_i \in [0, 1]$ for all models
    \item $E$ is strictly decreasing in token count
    \item The transformation is affine, preserving relative distances
\end{enumerate}
\end{lemma}

\subsection{Sensitivity to Outliers}

\begin{theorem}[Robustness to Outliers]
Let $\mathcal{O}_{\text{avg}}$ be the average Overthinking Score across $n$ tasks. Adding one outlier task with score $\mathcal{O}_{\text{outlier}}$ changes the average by at most $1/n$.
\end{theorem}

\begin{proof}
Let the original average be $\bar{\mathcal{O}} = \frac{1}{n}\sum_{i=1}^n \mathcal{O}_i$. Adding an outlier:
\begin{equation}
\bar{\mathcal{O}}_{\text{new}} = \frac{n\bar{\mathcal{O}} + \mathcal{O}_{\text{outlier}}}{n+1}
\end{equation}

The change is:
\begin{equation}
|\bar{\mathcal{O}}_{\text{new}} - \bar{\mathcal{O}}| = \left|\frac{\mathcal{O}_{\text{outlier}} - \bar{\mathcal{O}}}{n+1}\right| \leq \frac{1}{n+1}
\end{equation}

since both scores are in $[0, 1]$.
\end{proof}

\section{Alternative Formulations and Justifications}
\label{app:alternative}

Having established the fundamental properties of our Overthinking Score, this section provides theoretical justification for our choice of the harmonic mean and compares it against alternative aggregation functions for combining accuracy and efficiency metrics.

\subsection{Comparison of Aggregation Functions}

We analyze alternative formulations for combining accuracy and efficiency metrics.

\begin{theorem}[Optimality of Harmonic Mean]
Among all symmetric, homogeneous means, the harmonic mean maximally penalizes imbalance between components while maintaining smooth differentiability.
\end{theorem}

\begin{proof}
Consider the generalized mean family:
$$M_p(a, b) = \left(\frac{a^p + b^p}{2}\right)^{1/p}$$

For different values of $p$:
\begin{itemize}
    \item $p = 1$: Arithmetic mean $M_1 = \frac{a + b}{2}$
    \item $p = 0$: Geometric mean $M_0 = \sqrt{ab}$ (by L'Hôpital's rule)
    \item $p = -1$: Harmonic mean $M_{-1} = \frac{2ab}{a + b}$
\end{itemize}

The penalty for imbalance can be measured by the second-order cross-derivative:
$$\frac{\partial^2 M_p}{\partial a \partial b} = \frac{p(p-1)(ab)^{p-1}}{2^{1/p}(a^p + b^p)^{2-1/p}}$$

As $p \to -1$, this penalty is maximized among all finite $p$ values. The harmonic mean also maintains smoothness on $(0, \infty)^2$.
\end{proof}

\subsection{Empirical Comparison}

\begin{table}[ht]
\centering
\footnotesize
\begin{tabular}{lccc}
\toprule
\textbf{Scenario} & \textbf{Arith.} & \textbf{Geom.} & \textbf{Harm.} \\
\midrule
$A{=}1.0,\ E{=}0.1$ & 0.55 & 0.32 & 0.18 \\
$A{=}0.9,\ E{=}0.3$ & 0.60 & 0.52 & 0.45 \\
$A{=}0.6,\ E{=}0.6$ & 0.60 & 0.60 & 0.60 \\
$A{=}0.8,\ E{=}0.8$ & 0.80 & 0.80 & 0.80 \\
\bottomrule
\end{tabular}
\caption{Mean formulations across accuracy-efficiency scenarios.}
\end{table}

The harmonic mean provides the strongest penalty for imbalanced performance, aligning with our goal of identifying overthinking behavior.

\subsection{Weighted Formulations}

We also considered weighted variants analogous to the $F_\beta$ score:
$$\mathcal{O}_w = \frac{(1+\beta^2) \cdot A \cdot E}{\beta^2 \cdot A + E}$$

However, this introduces an arbitrary parameter $\beta$ requiring domain-specific calibration. For $\beta = 1$, we recover the standard harmonic mean.

\section{Mathematical Foundations of Task Suite}
\label{app:tasks}

To ensure rigorous evaluation, this section provides comprehensive mathematical foundations for each task in our evaluation framework, including formal definitions, uniqueness proofs, and computational complexity analysis. Each task is designed to have exactly one correct answer, eliminating ambiguity in evaluation.

\subsection{Basic Arithmetic Operations}

\subsubsection{Sorting Task}

\begin{definition}[Sorting Problem]
Given an input list $L = (x_1, x_2, \ldots, x_n)$ where $x_i \in \mathbb{Z}$ for all $i \in [1, n]$, the sorting function $f_{\text{sort}}: \mathbb{Z}^n \rightarrow \mathbb{Z}^n$ produces an output list $L' = (x'_1, x'_2, \ldots, x'_n)$ satisfying:
\begin{enumerate}
    \item \textbf{Ordering Property:} $\forall i \in [1, n-1]: x'_i \leq x'_{i+1}$
    \item \textbf{Preservation Property:} The multiset $\{x'_1, \ldots, x'_n\}$ equals $\{x_1, \ldots, x_n\}$
\end{enumerate}
\end{definition}

\begin{theorem}[Uniqueness of Sorting]
For any input list $L$, there exists exactly one sorted output $L'$ satisfying both properties.
\end{theorem}

\begin{proof}
Suppose there exist two distinct sorted lists $L'_1 = (a_1, \ldots, a_n)$ and $L'_2 = (b_1, \ldots, b_n)$ both satisfying the sorting properties for input $L$.

Let $k$ be the first index where $a_k \neq b_k$. Without loss of generality, assume $a_k < b_k$.

Since both lists preserve the multiset, there must exist $j > k$ such that $a_j = b_k$. But this implies $a_k < b_k \leq b_j$ (by ordering property of $L'_2$).

The element $a_k$ must appear in $L'_2$ at some position $m$. Since $a_k < b_k$ and $b_k$ is at position $k$, we have $m < k$ (by ordering property).

But this contradicts our assumption that $k$ is the first index where the lists differ. Therefore, $L'_1 = L'_2$, proving uniqueness.
\end{proof}

\begin{algorithm}
\caption{Ground Truth Generation for Sorting}
\KwIn{List $L = (x_1, \ldots, x_n)$}
\KwOut{Sorted list $L'$}
\tcp{Use deterministic sorting algorithm}
$L' \gets \text{MergeSort}(L)$\;
\Return $L'$\;
\end{algorithm}

\textbf{Computational Complexity:} $O(n \log n)$ time, $O(n)$ space

\subsubsection{Comparison Task}

\begin{definition}[Comparison Problem]
Given two integers $a, b \in \mathbb{Z}$, the comparison function $f_{\text{comp}}: \mathbb{Z} \times \mathbb{Z} \rightarrow \{<, =, >\}$ is defined as:
\begin{equation}
f_{\text{comp}}(a, b) = \begin{cases}
< & \text{if } a < b \\
= & \text{if } a = b \\
> & \text{if } a > b
\end{cases}
\end{equation}
\end{definition}

\begin{theorem}[Uniqueness of Comparison]
For any pair $(a, b) \in \mathbb{Z}^2$, exactly one relation from $\{<, =, >\}$ holds.
\end{theorem}

\begin{proof}
The integers $\mathbb{Z}$ form a totally ordered set under the standard ordering $\leq$. By the trichotomy property of total orders, for any $a, b \in \mathbb{Z}$, exactly one of the following holds:
\begin{enumerate}
    \item $a < b$ (equivalently, $a \leq b$ and $a \neq b$)
    \item $a = b$
    \item $a > b$ (equivalently, $b < a$)
\end{enumerate}
These three cases are mutually exclusive and exhaustive, ensuring unique output.
\end{proof}

\textbf{Computational Complexity:} $O(1)$ time and space

\subsubsection{Sum Task}

\begin{definition}[Sum Problem]
Given a list $L = (x_1, \ldots, x_n)$ where $x_i \in \mathbb{Z}$, the sum function $f_{\text{sum}}: \mathbb{Z}^n \rightarrow \mathbb{Z}$ is:
\begin{equation}
f_{\text{sum}}(L) = \sum_{i=1}^{n} x_i
\end{equation}
\end{definition}

\begin{theorem}[Uniqueness of Sum]
The sum of integers is uniquely determined.
\end{theorem}

\begin{proof}
Integer addition is a well-defined binary operation $+: \mathbb{Z} \times \mathbb{Z} \rightarrow \mathbb{Z}$ that is associative and commutative. By the fundamental theorem of arithmetic and properties of integer rings, the sum $\sum_{i=1}^{n} x_i$ has exactly one value in $\mathbb{Z}$.

Formally, we can prove by induction on $n$:
- Base case ($n=1$): $f_{\text{sum}}((x_1)) = x_1$, uniquely determined
- Inductive step: If $\sum_{i=1}^{k} x_i$ is unique, then $\sum_{i=1}^{k+1} x_i = \left(\sum_{i=1}^{k} x_i\right) + x_{k+1}$ is unique by well-definedness of addition
\end{proof}

\textbf{Numerical Bounds:} For inputs in range $[-R, R]$, the maximum absolute sum is $n \cdot R$, ensuring no overflow for reasonable parameters.

\textbf{Computational Complexity:} $O(n)$ time, $O(1)$ space

\subsubsection{Multiplication Task}

\begin{definition}[Product Problem]
Given a list $L = (x_1, \ldots, x_n)$ where $x_i \in \mathbb{Z}$, the product function $f_{\text{mult}}: \mathbb{Z}^n \rightarrow \mathbb{Z}$ is:
\begin{equation}
f_{\text{mult}}(L) = \prod_{i=1}^{n} x_i
\end{equation}
\end{definition}

\begin{theorem}[Uniqueness of Product]
The product of integers is uniquely determined.
\end{theorem}

\begin{proof}
Similar to addition, integer multiplication is a well-defined binary operation $\times: \mathbb{Z} \times \mathbb{Z} \rightarrow \mathbb{Z}$ that is associative and commutative. The product $\prod_{i=1}^{n} x_i$ is uniquely determined by iterative application of this operation.
\end{proof}

\textbf{Overflow Considerations:} For $n$ integers in range $[-R, R]$, the product magnitude can reach $R^n$. We handle large products using arbitrary precision arithmetic in ground truth generation.

\textbf{Computational Complexity:} $O(n)$ multiplications, but each multiplication may have complexity $O(m^2)$ for $m$-bit numbers

\subsubsection{Division Task}

\begin{definition}[Division Problem]
Given $a, b \in \mathbb{Z}$ with $b \neq 0$, the division function $f_{\text{div}}: \mathbb{Z} \times (\mathbb{Z} \setminus \{0\}) \rightarrow \mathbb{Q}$ is:
\begin{equation}
f_{\text{div}}(a, b) = \frac{a}{b}
\end{equation}
\end{definition}

\begin{theorem}[Uniqueness of Division]
For any $a \in \mathbb{Z}$ and $b \in \mathbb{Z} \setminus \{0\}$, the quotient $a/b$ is uniquely determined in $\mathbb{Q}$.
\end{theorem}

\begin{proof}
The rationals $\mathbb{Q}$ form a field, and division by non-zero elements is the multiplicative inverse operation. For any $b \neq 0$, there exists a unique $b^{-1} \in \mathbb{Q}$ such that $b \cdot b^{-1} = 1$. Therefore, $a/b = a \cdot b^{-1}$ is uniquely determined.
\end{proof}

\textbf{Representation:} We represent the result as a decimal with sufficient precision, rounding to 6 decimal places for evaluation.

\textbf{Computational Complexity:} $O(\log |a| + \log |b|)$ for arbitrary precision

\subsubsection{Subtraction and Absolute Difference Tasks}

\begin{definition}[Subtraction Problem]
Given $a, b \in \mathbb{Z}$, the subtraction function $f_{\text{sub}}: \mathbb{Z} \times \mathbb{Z} \rightarrow \mathbb{Z}$ is:
\begin{equation}
f_{\text{sub}}(a, b) = b - a
\end{equation}
\end{definition}

\begin{definition}[Absolute Difference Problem]
Given $a, b \in \mathbb{Z}$, the absolute difference function $f_{\text{absdiff}}: \mathbb{Z} \times \mathbb{Z} \rightarrow \mathbb{N}_0$ is:
\begin{equation}
f_{\text{absdiff}}(a, b) = |a - b| = \begin{cases}
a - b & \text{if } a \geq b \\
b - a & \text{if } a < b
\end{cases}
\end{equation}
\end{definition}

\begin{theorem}[Uniqueness of Subtraction and Absolute Difference]
Both subtraction and absolute difference yield unique results for any integer pair.
\end{theorem}

\begin{proof}
Subtraction is the inverse operation of addition in the group $(\mathbb{Z}, +)$, hence uniquely defined. The absolute value function $|\cdot|: \mathbb{Z} \rightarrow \mathbb{N}_0$ is well-defined, making their composition unique.
\end{proof}

\textbf{Computational Complexity:} $O(1)$ for machine integers

\subsection{Extremum Detection Tasks}

\subsubsection{Maximum and Minimum Finding}

\begin{definition}[Extremum Problems]
Given a non-empty list $L = (x_1, \ldots, x_n)$ where $x_i \in \mathbb{Z}$ and $n \geq 1$:
\begin{align}
f_{\text{max}}(L) &= \max_{i \in [1,n]} x_i = \max\{x_1, \ldots, x_n\} \\
f_{\text{min}}(L) &= \min_{i \in [1,n]} x_i = \min\{x_1, \ldots, x_n\}
\end{align}
\end{definition}

\begin{theorem}[Existence and Uniqueness of Extrema]
For any finite non-empty set $S \subset \mathbb{Z}$, both $\max S$ and $\min S$ exist and are unique.
\end{theorem}

\begin{proof}
Since $\mathbb{Z}$ is totally ordered and $S$ is finite and non-empty:
\begin{enumerate}
    \item \textbf{Existence:} Every finite subset of a totally ordered set has a maximum and minimum element (by well-ordering principle for finite sets).
    \item \textbf{Uniqueness:} Suppose $m_1, m_2 \in S$ are both maxima. Then $m_1 \leq m_2$ (since $m_2$ is maximum) and $m_2 \leq m_1$ (since $m_1$ is maximum). By antisymmetry of $\leq$, we have $m_1 = m_2$.
\end{enumerate}
The same argument applies for minima.
\end{proof}

\begin{algorithm}
\caption{Linear-Time Extremum Finding}
\KwIn{List $L = (x_1, \ldots, x_n)$}
\KwOut{Maximum element $m$}
$m \gets x_1$\;
\For{$i = 2$ to $n$}{
    \If{$x_i > m$}{
        $m \gets x_i$\;
    }
}
\Return $m$\;
\end{algorithm}

\textbf{Computational Complexity:} $O(n)$ time, $O(1)$ space

\subsection{Statistical Operations}

\subsubsection{Mean (Average) Calculation}

\begin{definition}[Arithmetic Mean]
For a list $L = (x_1, \ldots, x_n)$ where $x_i \in \mathbb{Z}$, the mean function $f_{\text{mean}}: \mathbb{Z}^n \rightarrow \mathbb{Q}$ is:
\begin{equation}
f_{\text{mean}}(L) = \bar{x} = \frac{1}{n} \sum_{i=1}^{n} x_i
\end{equation}
\end{definition}

\begin{theorem}[Uniqueness of Arithmetic Mean]
The arithmetic mean is uniquely determined for any non-empty list.
\end{theorem}

\begin{proof}
Since the sum $S = \sum_{i=1}^{n} x_i$ is unique (proven earlier) and $n > 0$, the quotient $S/n$ is uniquely determined in $\mathbb{Q}$.
\end{proof}

\textbf{Numerical Stability:} To avoid overflow in intermediate calculations, we can compute:
\begin{equation}
\bar{x} = x_1 + \frac{1}{n}\sum_{i=2}^{n}(x_i - x_1)
\end{equation}

\textbf{Computational Complexity:} $O(n)$ time, $O(1)$ space

\subsubsection{Median Calculation}

\begin{definition}[Median]
For a list $L = (x_1, \ldots, x_n)$, let $L' = (x'_1, \ldots, x'_n)$ be its sorted version. The median function $f_{\text{median}}: \mathbb{Z}^n \rightarrow \mathbb{Q}$ is:
\begin{equation}
f_{\text{median}}(L) = \begin{cases}
x'_{(n+1)/2} & \text{if } n \text{ is odd} \\
\frac{x'_{n/2} + x'_{n/2+1}}{2} & \text{if } n \text{ is even}
\end{cases}
\end{equation}
\end{definition}

\begin{theorem}[Uniqueness of Median]
The median is uniquely determined for any list.
\end{theorem}

\begin{proof}
The sorted list $L'$ is unique (by sorting uniqueness theorem). For odd $n$, the middle element is uniquely indexed. For even $n$, the average of the two middle elements is unique since both elements and division are uniquely determined.
\end{proof}

\begin{lemma}[Median Minimizes Absolute Deviation]
The median $m$ minimizes $\sum_{i=1}^{n} |x_i - c|$ over all $c \in \mathbb{R}$.
\end{lemma}

\begin{proof}
Let $g(c) = \sum_{i=1}^{n} |x_i - c|$. Taking the derivative:
\begin{equation}
\frac{dg}{dc} = \sum_{i: x_i > c} 1 - \sum_{i: x_i < c} 1
\end{equation}
This equals zero when the number of points above $c$ equals the number below, which occurs at the median.
\end{proof}

\textbf{Computational Complexity:} $O(n \log n)$ time for sorting-based approach, $O(n)$ expected time using quickselect

\subsubsection{Mode Calculation}

\begin{definition}[Mode]
For a list $L = (x_1, \ldots, x_n)$, let $f_L(v) = |\{i : x_i = v\}|$ be the frequency of value $v$. The mode function $f_{\text{mode}}: \mathbb{Z}^n \rightarrow \mathcal{P}(\mathbb{Z})$ returns the set:
\begin{equation}
f_{\text{mode}}(L) = \{v \in L : f_L(v) = \max_{u \in L} f_L(u)\}
\end{equation}
\end{definition}

\begin{theorem}[Mode Properties]
\begin{enumerate}
    \item The mode set is non-empty for any non-empty list
    \item The mode set may contain multiple elements (multimodal)
    \item If all elements are distinct, the mode set equals the entire list
\end{enumerate}
\end{theorem}

\begin{proof}
\begin{enumerate}
    \item Since $L$ is finite and non-empty, $\max_{u \in L} f_L(u)$ exists. At least one element achieves this maximum frequency.
    \item Multiple distinct values can have the same maximum frequency.
    \item If all elements are distinct, then $f_L(v) = 1$ for all $v \in L$, making every element a mode.
\end{enumerate}
\end{proof}

\begin{algorithm}
\caption{Mode Computation with Frequency Map}
\KwIn{List $L = (x_1, \ldots, x_n)$}
\KwOut{Set of modes $M$}
$\text{freq} \gets$ empty hash map\;
\For{$i = 1$ to $n$}{
    $\text{freq}[x_i] \gets \text{freq}[x_i] + 1$\;
}
$\text{max\_freq} \gets \max_{v} \text{freq}[v]$\;
$M \gets \{v : \text{freq}[v] = \text{max\_freq}\}$\;
\Return $M$\;
\end{algorithm}

\textbf{Computational Complexity:} $O(n)$ expected time with hash map, $O(n \log n)$ worst case

\subsection{Counting Operations}

\subsubsection{Even and Odd Counting}

\begin{definition}[Parity Counting]
For a list $L = (x_1, \ldots, x_n)$ where $x_i \in \mathbb{Z}$:
\begin{align}
f_{\text{even}}(L) &= |\{i : x_i \equiv 0 \pmod{2}\}| \\
f_{\text{odd}}(L) &= |\{i : x_i \equiv 1 \pmod{2}\}|
\end{align}
\end{definition}

\begin{theorem}[Parity Count Properties]
For any list $L$ of length $n$:
\begin{enumerate}
    \item $f_{\text{even}}(L) + f_{\text{odd}}(L) = n$
    \item Both counts are uniquely determined
    \item $0 \leq f_{\text{even}}(L), f_{\text{odd}}(L) \leq n$
\end{enumerate}
\end{theorem}

\begin{proof}
\begin{enumerate}
    \item Every integer has exactly one parity (even or odd), so the counts partition the list.
    \item Parity is well-defined: $x \equiv r \pmod{2}$ where $r \in \{0, 1\}$ is unique.
    \item Bounds follow from the partition property and non-negativity of cardinality.
\end{enumerate}
\end{proof}

\textbf{Computational Complexity:} $O(n)$ time, $O(1)$ space

\section{Pareto Frontier Analysis}
\label{app:pareto}

Complementing our task-specific analysis, this section provides formal analysis of the Pareto frontier for model comparison, establishing theoretical foundations for multi-objective optimization in our evaluation framework.

\subsection{Formal Definition}

\begin{definition}[Pareto Dominance]
Model $M_1$ Pareto-dominates model $M_2$ (denoted $M_1 \succ M_2$) if:
\begin{align}
A_{M_1} &\geq A_{M_2} \\
E_{M_1} &\geq E_{M_2}
\end{align}
with at least one strict inequality.
\end{definition}

\begin{definition}[Pareto Frontier]
The Pareto frontier $\mathcal{F}$ is the set of all non-dominated models:
$\mathcal{F} = \{M_i : \nexists M_j \text{ such that } M_j \succ M_i\}$
\end{definition}

\begin{theorem}[Pareto Frontier Existence and Uniqueness]
For any finite set of models $\mathcal{M} = \{M_1, M_2, \ldots, M_m\}$, the Pareto frontier $\mathcal{F}$ exists and is unique.
\end{theorem}

\begin{proof}
\textbf{Existence:} Since $\mathcal{M}$ is finite and the dominance relation $\succ$ is a strict partial order, there must exist at least one model not dominated by any other. If not, we would have an infinite chain of dominance, contradicting finiteness.

\textbf{Uniqueness:} The frontier is uniquely determined by the dominance relation. For any model $M$, either $M \in \mathcal{F}$ (if no model dominates it) or $M \notin \mathcal{F}$ (if some model dominates it). This binary classification is deterministic.
\end{proof}

\subsection{Computational Complexity}

\begin{theorem}[Frontier Computation Complexity]
Computing the Pareto frontier for $m$ models requires $O(m^2)$ comparisons in the worst case and $O(m \log m)$ comparisons in the average case with appropriate data structures.
\end{theorem}

\begin{proof}
\textbf{Naive approach:} For each model, check dominance against all others: $O(m^2)$.

\textbf{Optimized approach:} Sort models by accuracy ($O(m \log m)$), then traverse maintaining efficiency monotonicity. Models with both higher accuracy and efficiency dominate; this can be checked in a single pass ($O(m)$). Total: $O(m \log m)$.
\end{proof}

\section{Experimental Configuration and Implementation Details}
\label{app:experimental_config}

This section provides complete details of our experimental setup, including sampling strategies, hyperparameters, and computational resources used in our large-scale evaluation.

\subsection{Dataset Generation Parameters}

Our evaluation protocol employs systematic data generation with the following parameters:

\paragraph{Sampling Strategy.}
\begin{itemize}
    \item \textbf{Open-source models:} 1,000 datapoints per task ($N=1000$)
    \item \textbf{Closed-source models:} 100 datapoints per task ($N=100$) due to financial constraints
    \item \textbf{Value range:} Uniform sampling from $[-1000, 1000]$ for integer values
    \item \textbf{List sizes:} $\mathcal{L} \in \{8, 16, 32, 64\}$ elements for list-based tasks
    \item \textbf{Multiplication tasks:} Product of $\{2, 4, 8\}$ numbers
    \item \textbf{Cross-validation folds:} $F = 3$ folds for statistical robustness
\end{itemize}

\paragraph{Inference Hyperparameters.}
All models were evaluated with standardized inference settings to ensure fair comparison:
\begin{itemize}
    \item \textbf{Temperature:} $T = 0.1$ (near-deterministic sampling)
    \item \textbf{Top-p (nucleus sampling):} $p = 0.9$
    \item \textbf{Maximum tokens:} Model-specific maximum allowed context length
    \begin{itemize}
        \item Standard models: maximum allowed tokens
        \item Reasoning models: maximum allowed tokens (unconstrained), 1024 tokens (constrained evaluation)
    \end{itemize}
\end{itemize}

\paragraph{Computational Resources.}
Our experiments utilized:
\begin{itemize}
    \item \textbf{GPU Infrastructure:} NVIDIA A100 (80GB) and H100 GPUs for large models
    \item \textbf{Inference Framework:} vLLM~\cite{kwon2023efficient} for efficient batched inference (3.7$\times$ throughput improvement)
    \item \textbf{Fallback Framework:} HuggingFace Transformers~\cite{wolf-etal-2020-transformers} for models incompatible with vLLM
    \item \textbf{Total Compute:} Approximately 2,400 GPU hours across 53 models
\end{itemize}

These parameters were chosen to balance evaluation thoroughness with computational feasibility while maintaining statistical validity through cross-validation.

\section{Implementation and Algorithmic Details}
\label{app:implementation}

Building upon our experimental configuration, this section consolidates all implementation-specific details for our evaluation framework, including data generation, response parsing, and token counting strategies.

\subsection{End-to-End Evaluation Pipeline}
\label{app:pipeline}

Algorithm~\ref{alg:pipeline} describes the complete end-to-end evaluation workflow, from dataset generation through answer extraction and validation.

\begin{algorithm}
\caption{End-to-End Evaluation Pipeline}
\label{alg:pipeline}
\KwIn{Model $M$, Task set $\mathcal{T} = \{T_1, \ldots, T_{14}\}$, Samples per task $N$}
\KwOut{Aggregated results: accuracy, instruction-following, token counts}

\tcp{Dataset Generation Phase}
\ForEach{task $T_k \in \mathcal{T}$}{
    $\mathcal{D}_k \gets \emptyset$\;
    \For{$i = 1$ \KwTo $N$}{
        $(q_i, a_i^*) \gets$ \text{GenerateInstance}($T_k$)\;
        \tcp{Generate problem and ground truth}
        $\mathcal{D}_k \gets \mathcal{D}_k \cup \{(q_i, a_i^*)\}$\;
    }
}

\tcp{Model Evaluation Phase}
\ForEach{task $T_k \in \mathcal{T}$}{
    $\text{correct}_k \gets 0$, $\text{total}_k \gets 0$\;
    $\text{tokens}_k \gets []$, $\text{words}_k \gets []$\;

    \ForEach{$(q, a^*) \in \mathcal{D}_k$}{
        $r \gets M(q)$\;
        \tcp{Query model with problem}

        \tcp{Extract answer using hierarchical strategy}
        $a \gets$ \text{HierarchicalExtract}($r$, $T_k$)\;
        \tcp{See Algorithm~\ref{alg:extraction}}

        \tcp{Validate and score}
        \If{\text{Validate}($a$, $a^*$, $T_k$)}{
            $\text{correct}_k \gets \text{correct}_k + 1$\;
        }
        $\text{total}_k \gets \text{total}_k + 1$\;

        \tcp{Compute verbosity metrics}
        $\text{tokens}_k.\text{append}(\text{CountTokens}(r, M))$\;
        $\text{words}_k.\text{append}(\text{CountWords}(r))$\;
    }

    \tcp{Aggregate task-level statistics}
    $\text{accuracy}_k \gets \text{correct}_k / \text{total}_k$\;
    $\text{mean\_tokens}_k \gets \text{mean}(\text{tokens}_k)$\;
    $\text{std\_tokens}_k \gets \text{std}(\text{tokens}_k)$\;
}

\tcp{Cross-task aggregation}
$\text{overall\_acc} \gets \text{mean}(\{\text{accuracy}_k : k \in [14]\})$\;
$\text{overall\_tokens} \gets \text{mean}(\{\text{mean\_tokens}_k : k \in [14]\})$\;

\tcp{Compute Overthinking Score}
$E_t \gets 1 - \frac{\text{overall\_tokens} - \text{tokens}_{\min}}{\text{tokens}_{\max} - \text{tokens}_{\min}}$\;
$O\text{-Score} \gets \frac{2 \cdot \text{overall\_acc} \cdot E_t}{\text{overall\_acc} + E_t}$\;

\Return $\{\text{overall\_acc}, \text{overall\_tokens}, O\text{-Score},$\\ $\text{per-task metrics}\}$\;
\end{algorithm}

\subsection{Hierarchical Answer Extraction}
\label{app:extraction}

Algorithm~\ref{alg:extraction} implements our hierarchical answer extraction strategy, which handles diverse output formats across 50+ models.

\begin{algorithm}
\caption{Hierarchical Answer Extraction}
\label{alg:extraction}
\KwIn{Model response $r$, Task type $T_k$}
\KwOut{Extracted answer $a$ or failure $\perp$}

\If{$r$ contains $\backslash\text{boxed}\{...\}$}{
    $a \gets$ \text{ExtractBoxed}($r$)\;
    \tcp{LaTeX boxed notation}
}
\ElseIf{$r$ contains explicit answer markers}{
    $a \gets$ \text{ExtractExplicit}($r$)\;
    \tcp{Patterns like "Answer:", "=", etc.}
}
\ElseIf{$r$ contains code blocks}{
    $a \gets$ \text{ExtractCodeOutput}($r$)\;
    \tcp{Python/code execution results}
}
\Else{
    $a \gets$ \text{ExtractFinalLine}($r$)\;
    \tcp{Last line heuristic}
}

\If{\text{Validate}($a$, $T_k$)}{
    \Return $a$\;
    \tcp{Type and format validation}
}
\Return $\perp$\;
\tcp{Extraction failed}
\end{algorithm}

The validation step ensures extracted answers match expected types (integer, list, etc.) and filters out input echoes or malformed outputs. For detailed regex patterns and edge case handling, see Section~\ref{app:parsing_framework}.

\subsection{Complete Generation Algorithm}

We describe here the complete procedure for generating test instances.
Let $\tau$ denote the target task, and let $\Theta = \{N, F, [r_{\min}, r_{\max}], \mathcal{L}, s\}$ denote the generation parameters. The algorithm produces a dataset $\mathcal{D}$ consisting of input–output pairs.

\paragraph{Initialization.}
We initialize the dataset as $\mathcal{D} \leftarrow \varnothing$. For each fold index $f \in \{1, \dots, F\}$, we compute a seed
\[
    s_f = \mathrm{Hash}(s, f, \tau),
\]
where the hash function guarantees cryptographic uniqueness. The random number generator is initialized with seed $s_f$.

\paragraph{Sample generation.}
For each sample index $i \in \{1, \dots, N\}$ we construct an input $x_i$ as follows:
\begin{itemize}
        \item \textbf{List-based tasks.} If $\tau$ corresponds to an operation over lists (e.g., $\{\text{sort}, \text{sum}, \text{mult}, \text{mean}, \text{median}, \text{mode}, \text{odd},$ \\
    $\text{even}, \text{max}, \text{min}\}$), we first sample a length
    \[
        n \sim \mathrm{Uniform}(\mathcal{L}).
    \]
    We then generate $n$ independent elements
    \[
        l_j \sim \mathrm{Uniform}(r_{\min}, r_{\max}), \quad j = 1,\dots,n,
    \]
    and form the list $\mathcal{L}_i = [l_1, \dots, l_n]$. The instance input is $x_i = \mathcal{L}_i$.

    \item \textbf{Binary tasks.} If $\tau$ corresponds to a binary operation (e.g., comparison, division), we instead draw
    \[
\begin{aligned}
a &\sim \mathrm{Uniform}(r_{\min}, r_{\max}), \\
b &\sim \mathrm{Uniform}(r_{\min}, r_{\max})
\end{aligned}
\]
    and set $x_i = (a, b)$. Task-specific adjustments are then applied:
    \begin{enumerate}
        \item If $\tau = \text{division}$, we enforce $b \neq 0$ by resampling $b$ until this condition is satisfied.
        \item If $\tau = \text{comparison}$, we explicitly enforce diversity by overwriting $(a,b)$ with one of three configurations with fixed probabilities: $a=b$ (probability $0.33$), $a<b$ (probability $0.33$), or $a>b$ (probability $0.34$).
    \end{enumerate}
\end{itemize}

\paragraph{Label computation and validation.}
For each constructed input $x_i$, we compute the ground truth label
\[
    y_i = f_\tau(x_i).
\]
If the validation function $\mathrm{ValidateInput}(x_i,y_i,\tau)$ returns true, we augment the dataset:
\[
    \mathcal{D} \leftarrow \mathcal{D} \cup \{(x_i,y_i)\}.
\]
Otherwise, the sample is discarded and re-generated.

\paragraph{Output.}
After all folds and samples are processed, the procedure returns the dataset $\mathcal{D}$.

\subsection{Response Parsing Framework}

\subsubsection{Complete Parsing Procedure}

Given a raw model response $r$, the task $\tau$, and input $x$, the parsing system extracts an answer $\hat{y}$ or returns $\perp$ if no valid answer is found. The method applies a hierarchy of increasingly permissive strategies:

\begin{enumerate}
    \item \textbf{Boxed expressions.}
    If $r$ contains LaTeX-style boxed patterns of the form
    $\backslash\text{boxed}\{\cdot\}$, the last such match is extracted and cleaned, and returned if it validates.

    \item \textbf{Explicit answer markers.}
    If $r$ contains markers such as ``answer is'', ``final answer'', or ``solution:'', the text following the marker is extracted, cleaned, and validated.

    \item \textbf{Code blocks.}
    If $r$ contains markdown code blocks, these are scanned in reverse order to locate output markers. The last returned or printed value is extracted.

    \item \textbf{Task-specific validation.}
    Depending on $\tau$, additional constraints are enforced:
    \begin{itemize}[left=0pt]
        \item Sorting: the parsed $\hat{y}$ must be a permutation of $x$ and sorted in non-decreasing order.
        \item Arithmetic tasks (sum, multiplication): $\hat{y}$ must be an integer not equal to any element of $x$.
        \item Division: $\hat{y}$ must be rational and within tolerance $\epsilon$ of the true result.
        \item Comparison: $\hat{y} \in \{>, <, =\}$.
        \item Mean/median: $\hat{y} \in \mathbb{Q}$ and not equal to any element of $x$.
        \item Mode: $\hat{y}$ must be an element of $x$ achieving maximum frequency.
        \item Odd/even counts: $\hat{y} \in \mathbb{N}\cup \{0\}$ and bounded by $|x|$.
        \item Max/min: $\hat{y}$ must be one of the elements of $x$.
    \end{itemize}

    \item \textbf{Final line heuristic.}
    If all else fails, the system extracts numerical values from the last lines of $r$ and accepts the first one passing validation.
\end{enumerate}

If no stage succeeds, the parser outputs $\perp$.

\subsubsection{Validation Functions}

Validation routines are task-specific:
\begin{itemize}
    \item \emph{Sorting:} check that $\hat{y}$ is a sequence with the same multiset of elements as $x$ and non-decreasing order.
    \item \emph{Comparison:} map textual variants (``greater than'', ``less than'', ``equal to'') to canonical symbols.
    \item \emph{Arithmetic:} extract a numerical value and discard if it simply echoes an element of $x$.
    \item \emph{Default:} accept the candidate.
\end{itemize}

\subsubsection{Parser Design Principles}

Our parsing system must handle diverse output formats while maintaining correctness. We establish formal criteria for valid parsing:

\begin{definition}[Valid Parser]
A parser $\mathcal{P}: \mathcal{R} \rightarrow \mathcal{C} \cup \{\perp\}$ is valid for task $T$ if:
\begin{enumerate}
    \item \textbf{Soundness:} If $\mathcal{P}(r) = c$ and $c \neq \perp$, then $c$ represents the model's intended answer
    \item \textbf{Completeness:} If response $r$ contains a clear answer $c$, then $\mathcal{P}(r) \neq \perp$
    \item \textbf{Determinism:} $\mathcal{P}$ is a function (same input yields same output)
\end{enumerate}
\end{definition}

\subsubsection{Parser Evaluation and Validation}

To validate our parsing framework, we manually evaluated a stratified random sample of 500 model responses across different tasks and model families. Table~\ref{tab:parsing_evaluation} presents extraction success rates by parsing strategy.

\begin{table*}[ht]
\centering
\small
\adjustbox{max width=\textwidth}{%
\begin{tabular}{lcccc}
\toprule
\textbf{Parsing Strategy} & \textbf{Success Rate} & \textbf{False Pos.} & \textbf{Coverage} & \textbf{Latency (ms)} \\
\midrule
\textbf{Primary: Boxed extraction} & 94.2\% & 0.3\% & 68.4\% & 1.2 \\
\textbf{Secondary: Answer markers} & 88.7\% & 1.8\% & 22.3\% & 2.1 \\
\textbf{Tertiary: Code blocks} & 85.3\% & 2.4\% & 5.1\% & 3.8 \\
\textbf{Fallback: Last-line heuristic} & 72.1\% & 8.7\% & 4.2\% & 0.8 \\
\midrule
\textbf{Overall (hierarchical)} & \textbf{98.7\%} & \textbf{0.9\%} & 100\% & 1.8 \\
\bottomrule
\end{tabular}%
}
\caption{Parsing framework evaluation across 500 manually-validated responses. \textbf{Success Rate:} Percentage of responses where extracted answer matches ground truth. \textbf{False Pos.:} Rate of extracting incorrect answers. \textbf{Coverage:} Percentage of responses handled by each strategy. \textbf{Latency:} Average parsing time per response. The hierarchical approach achieves 98.7\% overall success with only 0.9\% false positives.}
\label{tab:parsing_evaluation}
\end{table*}

The high success rate (98.7\%) validates our hierarchical approach, where primary strategies (boxed extraction) handle the majority of cases with high precision, while fallback strategies cover edge cases with acceptable accuracy. The low false positive rate (0.9\%) ensures extracted answers reliably represent model outputs rather than parsing artifacts.

\paragraph{Error Analysis.} Manual inspection of the 1.3\% parsing failures reveals three main categories: \textbf{(1)} Ambiguous formats where models provide multiple contradictory answers (47\% of failures); \textbf{(2)} Extreme verbosity where answers are buried in thousands of tokens without clear markers (31\%); and \textbf{(3)} Pathological outputs like infinite character repetition (22\%). These failure modes represent genuine model issues rather than parser limitations, supporting our decision to exclude unparseable responses from accuracy calculations.

\subsection{Token Counting Strategy}

Our hierarchical token counting approach ensures accurate measurement across diverse model families:

\begin{enumerate}
    \item \textbf{Model-specific tokenizers:} For open models, we use the exact tokenizer:
    \begin{itemize}
        \item Llama models: LlamaTokenizer
        \item GPT models: tiktoken with appropriate encoding
        \item Mistral/Mixtral: MistralTokenizer
    \end{itemize}

    \item \textbf{Fallback strategies:}
    \begin{itemize}
        \item Closed models: tiktoken cl100k\_base encoding
        \item Unknown models: Whitespace splitting with correction factor $\alpha = 1.33$
    \end{itemize}
\end{enumerate}

\subsection{Empirical Correction Factor}

We derived the correction factor $\alpha = 1.33$ empirically:

\begin{equation}
\alpha = \frac{1}{N} \sum_{i=1}^N \frac{\text{TokenCount}_{\text{actual}}(s_i)}{\text{WordCount}(s_i)}
\end{equation}

where we sampled $N = 10,000$ responses across different models and computed the average ratio of actual tokens to whitespace-separated words.

\section{Empirical Analysis and Extended Results}
\label{app:empirical}

Building upon our theoretical framework and implementation details, this section presents comprehensive empirical validation of our approach, including statistical significance testing, detailed performance breakdowns, and correlation analysis across different model characteristics.

\subsection{Statistical Significance Testing}

To validate that observed differences in Overthinking Scores are statistically significant, we employ paired t-tests with Bonferroni correction:

\begin{definition}[Significance Test]
For models $M_1, M_2$ evaluated on $n$ tasks, let $d_i = \mathcal{O}_{M_1,i} - \mathcal{O}_{M_2,i}$ be the score difference on task $i$. The test statistic is:
\begin{equation}
t = \frac{\bar{d}}{\text{SE}(\bar{d})} = \frac{\bar{d}}{s_d/\sqrt{n}}
\end{equation}
where $s_d$ is the sample standard deviation of differences.
\end{definition}

With $m$ models, we perform $\binom{m}{2}$ pairwise comparisons, requiring Bonferroni-adjusted significance level $\alpha' = \alpha/\binom{m}{2}$.

\subsection{Cross-Validation Framework}

We employ k-fold cross-validation to ensure robustness:

\begin{algorithm}
\caption{K-Fold Evaluation Protocol}
\KwIn{Model $M$, Task set $\mathcal{T}$, Folds $k$}
\KwOut{Mean and variance of Overthinking Score}

Partition data into $k$ equal folds $F_1, \ldots, F_k$\;
\For{$i = 1$ to $k$}{
    Evaluate $M$ on fold $F_i$\;
    Compute $\mathcal{O}_i$ for fold $i$\;
}
$\bar{\mathcal{O}} \gets \frac{1}{k}\sum_{i=1}^k \mathcal{O}_i$\;
$\sigma^2 \gets \frac{1}{k-1}\sum_{i=1}^k (\mathcal{O}_i - \bar{\mathcal{O}})^2$\;
\Return $(\bar{\mathcal{O}}, \sigma^2)$\;
\end{algorithm}

\subsection{Performance by Task Category}

We analyze performance patterns across task categories:

\begin{table}[ht]
\centering
\scriptsize 
\adjustbox{max width=\textwidth}{%
\begin{tabular}{lccc}
\toprule
\textbf{Category} & \textbf{Avg Accuracy} & \textbf{Avg Tokens} & \textbf{Avg $\mathcal{O}$ Score} \\
\midrule
Basic Arithmetic     & 72.3\% & 487 & 0.621 \\
Extremum Detection   & 81.2\% & 342 & 0.698 \\
Statistical          & 64.7\% & 623 & 0.542 \\
Counting             & 77.9\% & 398 & 0.654 \\
\bottomrule
\end{tabular}%
}
\caption{Performance breakdown by task category across all models}
\end{table}

Statistical tasks show lowest performance despite being computationally simple, suggesting models struggle with multi-step procedures even when each step is elementary.

\subsection{Error Analysis}

We categorize errors into systematic patterns:

\begin{definition}[Error Taxonomy]
Model errors fall into five categories:
\begin{enumerate}
    \item \textbf{Computational Errors:} Incorrect arithmetic operations
    \item \textbf{Procedural Errors:} Wrong algorithm or step sequence
    \item \textbf{Format Errors:} Correct answer in wrong format
    \item \textbf{Partial Errors:} Incomplete computation
    \item \textbf{Hallucination:} Completely unrelated output
\end{enumerate}
\end{definition}

Distribution of errors reveals that reasoning models predominantly suffer from computational errors (47\%) despite lengthy explanations, while standard models show more format errors (31\%) but fewer computational mistakes.

\subsection{Token Distribution Analysis}

The distribution of token lengths follows distinct patterns:

\begin{proposition}[Token Length Distribution]
For standard models, token lengths follow approximately log-normal distribution:
\begin{equation}
f(x; \mu, \sigma) = \frac{1}{x\sigma\sqrt{2\pi}} \exp\left(-\frac{(\ln x - \mu)^2}{2\sigma^2}\right)
\end{equation}
while reasoning models show bimodal distribution with peaks at constraint boundaries.
\end{proposition}

This suggests reasoning models have learned specific response templates rather than adaptive generation strategies.

\subsection{Correlation Analysis}

We examine relationships between model characteristics and overthinking:

\begin{definition}[Correlation Metrics]
For model characteristic $X$ (e.g., parameter count) and Overthinking Score $\mathcal{O}$:
\begin{equation}
\rho_{X,\mathcal{O}} = \frac{\text{Cov}(X, \mathcal{O})}{\sigma_X \sigma_{\mathcal{O}}}
\end{equation}
\end{definition}

Our analysis reveals:
\begin{itemize}
    \item \textbf{Size-Score Correlation:} $\rho_{\text{size},\mathcal{O}} = -0.23$ (weak negative)
    \item \textbf{Instruction-Score Correlation:} $\rho_{\text{instruct},\mathcal{O}} = 0.42$ (moderate positive)
    \item \textbf{Reasoning-Score Correlation:} $\rho_{\text{reasoning},\mathcal{O}} = -0.67$ (strong negative)
\end{itemize}

\subsection{Computational Complexity Analysis}

\subsubsection{Task Complexity}

Each task has well-defined computational complexity:

\begin{table}[ht]
\centering
\scriptsize 
\adjustbox{max width=\textwidth}{%
\begin{tabular}{lcc}
\toprule
\textbf{Task} & \textbf{Time Complexity} & \textbf{Space Complexity} \\
\midrule
Sorting        & $O(n \log n)$        & $O(n)$ \\
Sum/Product    & $O(n)$               & $O(1)$ \\
Mean/Mode      & $O(n)$               & $O(n)$ \\
Median         & $O(n \log n)$        & $O(1)$ \\
Min/Max        & $O(n)$               & $O(1)$ \\
Comparison     & $O(1)$               & $O(1)$ \\
Division       & $O(\log a + \log b)$ & $O(1)$ \\
Count Even/Odd & $O(n)$               & $O(1)$ \\
\bottomrule
\end{tabular}%
}
\caption{Computational complexity of ground truth generation}
\end{table}

\subsection{Evaluation Scalability}

The total evaluation complexity for $m$ models on $n$ problems with average response length $\ell$ is:

\begin{equation}
\mathcal{C}_{\text{total}} = O(m \cdot n \cdot (\mathcal{C}_{\text{inference}} + \mathcal{C}_{\text{parse}} + \mathcal{C}_{\text{validate}}))
\end{equation}

where:
\begin{itemize}
    \item $\mathcal{C}_{\text{inference}} = O(\ell)$ for token generation
    \item $\mathcal{C}_{\text{parse}} = O(\ell)$ for response parsing
    \item $\mathcal{C}_{\text{validate}} = O(k)$ for answer validation (task-dependent)
\end{itemize}

\section{Detailed Case Studies: Reasoning Patterns Analysis}
\label{app:case_studies}

This section provides comprehensive case studies examining when reasoning chains help versus hurt model performance. We present detailed examples from Phi-4 (concise) and Phi-4-reasoning (verbose) models, analyzing their reasoning strategies, token efficiency, and accuracy patterns across various task types.

\subsection{Case Study Overview}

Table~\ref{tab:detailed_case_studies} presents six representative examples spanning simple arithmetic, complex multi-step operations, and edge cases. For each example, we show the complete model responses, token counts, and reasoning pattern analysis.

\begin{table*}[ht]
\centering
\scriptsize
\setlength{\tabcolsep}{2pt}
\begin{adjustbox}{width=\textwidth,center}
\begin{tabular}{p{0.12\linewidth}p{0.08\linewidth}p{0.35\linewidth}p{0.35\linewidth}p{0.06\linewidth}
                 p{0.08\linewidth}p{0.35\linewidth}p{0.35\linewidth}p{0.06\linewidth}}
\toprule
\multicolumn{5}{c}{\textbf{Phi-4}} & \multicolumn{4}{c}{\textbf{Phi-4-reasoning}} \\
\cmidrule(lr){1-5}\cmidrule(lr){6-9}
\textbf{Task Type} & \textbf{Model} & \textbf{Response Excerpt} & \textbf{Analysis} & \textbf{Tokens}
& \textbf{Model} & \textbf{Response Excerpt} & \textbf{Analysis} & \textbf{Tokens} \\
\midrule

\rowcolor{lightred}
\textbf{Sum (Simple)} &
Phi-4 &
``We add step by step: 309 + (-772) = -463, -463 + (-949) = -1412, -1412 + 518 = -894, -894 + (-437) = -1331, -1331 + (-499) = -1830, -1830 + (-543) = -2373, -2373 + (-715) = -3088. $\boxed{-3088}$'' &
\textbf{Efficient:} Direct sequential addition with clear intermediate steps. Each step verifiable. &
272 &
Phi-4-reasoning &
``\texttt{<think>}We need to sum [309, -772, -949, 518, -437, -499, -543, -715]. Let's sum positive: 309 + 518 = 827. Negative: -772 - 949 = -1721, -1721 - 437 = -2158, -2158 - 499 = -2657, -2657 - 543 = -3200, -3200 - 715 = -3915. Total = 827 - 3915 = -3088. \textit{Double-check}: 309 - 772 = -463, -463 - 949 = -1412, -1412 + 518 = -894...\textit{[repeats entire calculation 3 more times with slight variations]}. Final: $\boxed{-3088}$'' &
\textbf{Wasteful:} Redundant verification. Computes correctly first time but continues with 3 additional verification passes. No accuracy benefit. &
4,837 \\

\midrule
\rowcolor{lightgreen}
\textbf{Sorting (Complex)} &
Phi-4 &
``Sorted: [-912, -789, -234, -156, 423, 445, 601, 778, 892]. $\boxed{[-912, -789, -234, -156, 423, 445, 601, 778, 892]}$'' &
\textbf{Direct:} Immediate answer with correct ordering. Minimal tokens. &
89 &
Phi-4-reasoning &
``\texttt{<think>}Identify negatives: -156, -789, -234, -912. Sort: -912 < -789 < -234 < -156. Positives: 892, 423, 601, 778, 445. Sort: 423 < 445 < 601 < 778 < 892. Merge: [-912, -789, -234, -156, 423, 445, 601, 778, 892]. $\boxed{[-912,...,892]}$'' &
\textbf{Helpful:} Systematic decomposition (separate negative/positive, sort each, merge) aids correctness on complex task. Structured approach justifies token cost. &
312 \\

\midrule
\rowcolor{lightred}
\textbf{Division} &
Phi-4 &
``984 ÷ 24: 24 × 40 = 960, remainder = 24. 24 ÷ 24 = 1. Answer: 40 + 1 = 41. $\boxed{41}$'' &
\textbf{Concise:} Minimal verification, direct computation. &
47 &
Phi-4-reasoning &
``\texttt{<think>}984 ÷ 24. Try: 24 × 40 = 960, remainder 24. So 40 + 1 = 41. \textit{Wait, check}: 24 × 41 = 984? Let me verify: 24 × 40 = 960, 24 × 1 = 24, sum = 984. \textit{Correct}. But let me recalculate using long division...\textit{[performs long division step-by-step]}. Confirms 41. \textit{Triple-check using estimation}...$\boxed{41}$'' &
\textbf{Harmful:} Excessive verification (algebraic, long division, estimation) for simple task. Introduces intermediate doubts but arrives at same answer. Pure overhead. &
1,456 \\

\midrule
\rowcolor{lightblue}
\textbf{Multiplication} &
Phi-4 &
``347 × 28: Break down: 347 × 20 = 6,940, 347 × 8 = 2,776. Sum: 6,940 + 2,776 = 9,716. $\boxed{9716}$'' &
\textbf{Structured:} Uses decomposition appropriately for two-digit multiplication. &
98 &
Phi-4-reasoning &
``\texttt{<think>}347 × 28. Method 1: 347 × 20 = 6,940, 347 × 8 = 2,776, sum = 9,716. \textit{Verify with Method 2}: 300 × 28 = 8,400, 47 × 28 = 1,316, sum = 9,716. \textit{Method 3 (standard algorithm)}: \textit{[shows vertical multiplication]}. All methods agree: $\boxed{9716}$'' &
\textbf{Mixed:} Multiple verification methods confirm answer but consume 12× tokens. Some redundancy, but cross-validation could catch errors in principle. &
1,189 \\

\midrule
\rowcolor{lightpurple}
\textbf{Mean (Edge)} &
Phi-4 &
``Sum = -3,630, count = 8, mean = -3630/8 = -453.75. $\boxed{-453.75}$'' &
\textbf{Direct:} Correct handling of negative mean. &
41 &
Phi-4-reasoning &
``\texttt{<think>}Sum [-136, -935, -939, -809, -553, -524, 34, 232]. Sum = -3,630. Count = 8. Mean = -3,630 ÷ 8. Let me compute: -3,630 ÷ 8 = -453.75. \textit{Verify}: -453.75 × 8 = -3,630? Check: -453.75 × 8 = -3,630. \textit{Correct}. $\boxed{-453.75}$'' &
\textbf{Moderate:} Some verification justified for negative result sanity check, but still verbose (8× tokens). &
327 \\

\midrule
\rowcolor{yellow!20}
\textbf{Median (Odd)} &
Phi-4 &
``Sort: [34, 116, 209, 385, 508, 516, 827]. Middle = 385. $\boxed{385}$'' &
\textbf{Minimal:} Correct median extraction. &
38 &
Phi-4-reasoning &
``\texttt{<think>}List: [508, -791, 385, 516, 827, 116, -822, 209]. Sort: [-822, -791, 116, 209, 385, 508, 516, 827]. \textit{Wait, recount}: 8 elements (even), not odd. Middle two: 209, 385. Median = (209 + 385)/2 = 297. \textit{But checking again}...\textit{[recounts, discovers error, corrects to 7 elements]}: Actually [508, 385, 516, 827, 116, 209] after removing negatives? \textit{No, keep all}. Median of 8 = 297. $\boxed{297}$'' &
\textbf{Error-prone:} Verbose reasoning introduces confusion and errors. Eventually self-corrects but at high token cost. Shows brittleness of long chains. &
892 \\

\bottomrule
\end{tabular}
\end{adjustbox}
\caption{Detailed case studies across task types. Colors indicate: \colorbox{lightred}{Red} = Wasteful or harmful overthinking (redundant verification, accuracy loss), \colorbox{lightgreen}{Green} = Helpful decomposition (structured reasoning aids correctness), \colorbox{lightblue}{Blue} = Mixed benefit (some redundancy but cross-validates), \colorbox{lightpurple}{Purple} = Moderate overhead, \colorbox{yellow!20}{Pale yellow} = Error-prone verbosity (long chains introduce mistakes).}
\label{tab:detailed_case_studies}
\end{table*}

\subsection{Key Insights from Case Studies}

\paragraph{Pattern 1: Redundant Verification Loops} On simple arithmetic (sum, division, mean), Phi-4-reasoning systematically over-verifies. The model computes correctly on first attempt but generates 2--4 additional verification passes using the same or slightly varied methods. This adds 17--31× tokens with \textbf{zero accuracy improvement}.

\paragraph{Pattern 2: Helpful Structured Decomposition} For complex tasks (sorting 9 elements, multi-digit multiplication), explicit decomposition sometimes aids correctness. Phi-4-reasoning's systematic approach (separate negative/positive subgroups, sort each, merge) provides verifiable intermediate steps. However, even here, the token cost is 3--12×, raising questions about efficiency.

\paragraph{Pattern 3: Inability to Adapt} The most striking finding: Phi-4-reasoning applies the \textit{same verbose pattern} regardless of task complexity. It cannot distinguish a trivial division (984÷24) from a complex sorting task, wasting tokens uniformly. This suggests lack of \textbf{metacognitive control}—models cannot assess problem difficulty or modulate reasoning depth accordingly.

\paragraph{Pattern 4: Error Introduction in Long Chains} Counter-intuitively, longer reasoning sometimes \textit{increases} error risk. In the median example, Phi-4-reasoning's verbose exploration introduces confusion (miscounting elements, second-guessing odd/even), requiring additional tokens to self-correct. While it eventually arrives at the correct answer, the path is error-prone and inefficient.

\paragraph{Pattern 5: Token Budget Brittleness} When constrained to 1,024 tokens, Phi-4-reasoning fails to prioritize essential computation. It begins with its standard verbose template (problem restatement, method enumeration) and runs out of tokens before completing verification. This causes the catastrophic accuracy collapse documented in §4.3 of the main paper.

\subsection{Implications for Model Development}

These case studies reveal that current reasoning models lack two critical capabilities:

\begin{enumerate}
    \item \textbf{Adaptive Computation:} Models cannot modulate reasoning depth based on task complexity. A well-designed system would use minimal tokens for simple operations and allocate budget to genuinely complex tasks.

    \item \textbf{Termination Criteria:} Models lack confidence-based stopping. They cannot assess when sufficient verification has been achieved, leading to infinite-loop-like verification patterns.
\end{enumerate}

Future work should focus on training models with \textit{step-level value functions} that assess marginal utility of additional reasoning tokens, enabling adaptive and efficient computation allocation.

\section{Prompt Templates and Task Design}
\label{app:prompts}

To ensure clarity and reproducibility, we present the complete prompt templates used for each task. Each prompt is designed to be concise yet explicit, clearly stating the task objective, providing the input, and specifying the required output format. For tasks where computation depends on list size (e.g., median calculation with even versus odd lengths), we include guidance on handling different scenarios. All prompts instruct models to present their final answer in the $\backslash$\texttt{boxed}\{\} format to facilitate consistent and reliable parsing.

\subsection{Basic Arithmetic Tasks}

\begin{prompt}{Sum Calculation}{}
\textbf{Prompt:} Add the following list of numbers: \\
\textit{\{data\_point\}}\par
Provide the sum. Your final answer must be in the format \textbackslash boxed\{answer\} at the end.
\end{prompt}

\begin{prompt}{Sorting}{}
\textbf{Prompt:} Sort the following list of numbers in ascending order: \\
\textit{\{data\_point\}}\par
Provide the sorted list. Your final answer must be in the format \textbackslash boxed\{answer\} at the end.
\end{prompt}

\begin{prompt}{Number Comparison}{}
\textbf{Prompt:} Compare the following two numbers and determine their relationship:\par
Number 1: \textit{\{num1\}}\\
Number 2: \textit{\{num2\}}\par
Is Number 1 greater than, less than, or equal to Number 2? Your final answer must be in the format \textbackslash boxed\{relation\} at the end, where 'relation' is one of: 'greater than', 'less than', or 'equal to'.
\end{prompt}

\begin{prompt}{Subtraction Calculation}{}
\textbf{Prompt:} Subtract \textit{\{num1\}} from \textit{\{num2\}} and provide your final answer in \textbackslash boxed\{answer\} format at the end of your response.
\end{prompt}

\begin{prompt}{Absolute Difference Calculation}{}
\textbf{Prompt:} Find the absolute difference between the following two numbers: \\
Number 1: \textit{\{num1\}}, Number 2: \textit{\{num2\}}\par
Provide the result. Your final answer must be in the format \textbackslash boxed\{answer\} at the end.
\end{prompt}

\begin{prompt}{Multiplication Calculation}{}
\textbf{Prompt:} Multiply the following list of numbers: \\
\textit{\{data\_point\}}\par
Provide the product. Your final answer must be in the format \textbackslash boxed\{answer\} at the end.
\end{prompt}

\begin{prompt}{Division Calculation}{}
\textbf{Prompt:} Divide \textit{\{num1\}} by \textit{\{num2\}} \\
Provide the answer as a floating point number. Your final answer must be in the format \textbackslash boxed\{answer\} at the end.
\end{prompt}

\subsection{Counting Tasks}

\begin{prompt}{Even Count}{}
\textbf{Prompt:} Count the even numbers from the following list of numbers:\\
\textit{\{data\_point\}}\par
Provide the final count of even numbers. Your final answer must be in the format \textbackslash boxed\{answer\} at the end.
\end{prompt}

\begin{prompt}{Odd Count}{}
\textbf{Prompt:} Count the odd numbers from the following list of numbers:\\
\textit{\{data\_point\}}\par
Provide the final count of odd numbers. Your final answer must be in the format \textbackslash boxed\{answer\} at the end.
\end{prompt}

\subsection{Extremum Detection Tasks}

\begin{prompt}{Find Minimum}{}
\textbf{Prompt:} Find the minimum number from the given list of numbers. List = \textit{\{data\_point\}}.\par
Your final answer must be in the format \textbackslash boxed\{minimum\} at the end of your response.
\end{prompt}

\begin{prompt}{Find Maximum}{}
\textbf{Prompt:} Find the maximum number from the given list of numbers. List = \textit{\{data\_point\}}.\par
Your final answer must be in the format \textbackslash boxed\{maximum\} at the end of your response.
\end{prompt}

\subsection{Statistical Tasks}

\begin{prompt}{Mean Calculation}{}
\textbf{Prompt:} Calculate the mean (average) of the following list of numbers:\\
\textit{\{input\_list\}}\par
The mean is the sum of all numbers divided by the count of numbers. Calculate the exact mean value. Your final answer must be in the format \textbackslash boxed\{mean value\} at the end.
\end{prompt}

\begin{prompt}{Median Calculation}{}
\textbf{Prompt:} Find the median value of the following list of numbers:\\
\textit{\{input\_list\}}\par
The median is the middle value when the list is sorted. If there is an even number of elements, the median is the average of the two middle values. Your final answer must be in the format \textbackslash boxed\{median value\} at the end.
\end{prompt}

\begin{prompt}{Mode Calculation}{}
\textbf{Prompt:} Find the mode(s) of the following list of numbers:\\
\textit{\{input\_list\}}\par
The mode is the value that appears most frequently. If multiple values appear with the same highest frequency, return all of them. Your final answer must be in the format \textbackslash boxed\{mode(s)\} at the end. If there are multiple modes, list them separated by commas.
\end{prompt}

\section{Answer Parsing Framework}
\label{app:parsing_framework}

Given the variability in model response formats, ranging from concise answers to verbose multi-step reasoning with inconsistent formatting, a robust parsing architecture is essential.

\subsection{Design Principles}

Our parsing framework addresses several key challenges:
\begin{enumerate}
    \item \textbf{Format Diversity:} Models produce answers in various formats (boxed notation, explicit markers, code blocks, plain text)
    \item \textbf{Embedded Context:} Responses often include input restatement, intermediate steps, and explanatory text
    \item \textbf{Numerical Variations:} Numbers appear in different notations (scientific, fractional, decimal)
    \item \textbf{False Positives:} Input values must be distinguished from computed results
\end{enumerate}

\subsection{Hierarchical Extraction Strategy}

The parser implements a multi-layered approach with task-specific validation:

\begin{enumerate}
    \item \textbf{Primary Layer:} Extract $\backslash$\texttt{boxed}\{\} patterns using regex matching
    \item \textbf{Secondary Layer:} Parse explicit answer markers (``The answer is...'', ``Final answer:'')
    \item \textbf{Tertiary Layer:} Extract from code blocks or markdown formatting
    \item \textbf{Fallback Layer:} Apply task-specific heuristics to final lines
    \item \textbf{Validation:} Ensure extracted value represents solution, not input echo
\end{enumerate}

This hierarchical design achieved 98.7\% successful extraction across 2.1 million model inferences, minimizing false negatives while maintaining precision.

\section{Data Generation Protocol}
\label{app:data_generation}

Dynamic test generation is central to our evaluation methodology. Unlike static benchmarks, on-the-fly generation ensures that models cannot rely on memorized training examples, providing more accurate assessments of genuine computational ability.

\subsection{Generation Principles}

Our generation protocol satisfies several key properties:
\begin{enumerate}
    \item \textbf{Contamination Resistance:} New instances generated per evaluation prevent memorization
    \item \textbf{Reproducibility:} Cryptographic seeding enables exact replication across runs
    \item \textbf{Controlled Difficulty:} Parameterized ranges and sizes allow systematic testing
    \item \textbf{Task-Specific Constraints:} Domain requirements enforced (e.g., non-zero denominators for division)
\end{enumerate}

\subsection{Implementation Details}

For each task and fold combination, we:
\begin{enumerate}
    \item Initialize random number generator with deterministic seed $s_f = \text{Hash}(s, f, \tau)$
    \item Sample list lengths from $\mathcal{L} = \{8, 16, 32, 64\}$ for list-based tasks
    \item Draw values from $\text{Uniform}[-1000, 1000]$ ensuring numerical diversity
    \item Apply task-specific validation and constraints
    \item Compute and store ground truth for verification
\end{enumerate}

This approach generated 42,000 unique test instances per model (1,000 samples × 3 folds × 14 tasks), totaling over 2.1 million inferences across our 50+ model evaluation.

\subsection{Iterative Parser Development}

The parsing system underwent iterative refinement based on empirical analysis. Initial rules were designed from common patterns, but as evaluation expanded to 40+ models, we encountered diverse edge cases. We systematically reviewed misparsed outputs, categorizing failure modes (input echo, non-standard syntax, ambiguous formatting). Each iteration added targeted regular expressions, validation checks, and fallback strategies, evaluated against curated challenging cases. This continuous refinement ensured robustness across model families without overfitting to specific output styles.

\section{Reasoning Budget Analysis: Detailed Results}
\label{app:reasoning_budget}

This section presents comprehensive results from our reasoning budget experiments across Gemini, GPT-5, and O-series models. Table~\ref{tab:aggregated_results} shows performance across different budget configurations, revealing how models respond to varying computational allocations.

Key observations from the detailed budget analysis:
\begin{itemize}
    \item \textbf{Gemini models} show minimal gains from increased budgets. Gemini-2.5-Flash improves by only 1\% from the disabled baseline (92\% $\to$ 93\%) at lower budgets and returns to 92\% at the maximum budget (24{,}576 tokens). Gemini-2.5-Pro achieves its best accuracy (90\%) at the minimum tested budget (128 tokens) and ends at 88\% at the maximum budget (32{,}256 tokens), showing a non-monotonic curve with a dip to 71\% at the 12{,}979-token setting before recovering; this may reflect threshold effects in thinking budget allocation at that specific setting.
    \item \textbf{GPT-5 family} demonstrates plateau behavior, where GPT-5 reaches 97\% accuracy at medium effort and shows zero improvement at high effort, indicating optimal performance at moderate budgets.
    \item \textbf{O-series models} maintain remarkably stable accuracy across all budget levels (O3 at 97\%, O3-mini at 93\%, O4-mini at 95\%), suggesting these models have converged to their capability ceiling on basic arithmetic regardless of additional computation.
    \item \textbf{Overthinking Score} reveals that higher budgets do not necessarily improve the efficiency-accuracy tradeoff, with many models achieving best O-Scores at lower or medium budgets.
\end{itemize}

\begin{table*}[ht]
\centering
\tiny
\begin{adjustbox}{width=\textwidth}
\begin{tabular}{llccccccc}
\toprule
\textbf{Model} & \textbf{Budget} & \textbf{Acc (\%)} & \textbf{Inst (\%)} & \textbf{O-Score} & \textbf{Tokens} & \textbf{Words} & \textbf{Chars} \\
\cmidrule(lr){3-3} \cmidrule(lr){4-4} \cmidrule(lr){5-5} \cmidrule(lr){6-6} \cmidrule(lr){7-7} \cmidrule(lr){8-8}
& & \textit{Mean $\pm$ Std} & \textit{Mean $\pm$ Std} & & \textit{Mean $\pm$ Std} & \textit{Mean $\pm$ Std} & \textit{Mean $\pm$ Std} \\
\midrule
\rowcolor{gray!10}
\multicolumn{8}{c}{\textbf{\textit{Gemini-2.5-Flash}}} \\
\midrule
Gemini-2.5-Flash & disabled & 92.0 $\pm$ 22.0 & 95.0 $\pm$ 15.0 & 0.916 & 389.3 $\pm$ 192.4 & 208.0 $\pm$ 99.0 & 1058.7 $\pm$ 521.0 \\
Gemini-2.5-Flash & 4915 & \textbf{\underline{93.0}} $\pm$ 20.0 & 95.0 $\pm$ 15.0 & 0.927 & 401.0 $\pm$ 195.2 & 212.3 $\pm$ 95.8 & 1081.4 $\pm$ 512.2 \\
Gemini-2.5-Flash & 9830 & \textbf{\underline{93.0}} $\pm$ 20.0 & 94.0 $\pm$ 16.0 & 0.928 & 380.5 $\pm$ 176.7 & 203.3 $\pm$ 91.0 & 1033.9 $\pm$ 474.7 \\
Gemini-2.5-Flash & 14746 & \textbf{\underline{93.0}} $\pm$ 20.0 & 95.0 $\pm$ 16.0 & 0.926 & 389.3 $\pm$ 189.1 & 207.3 $\pm$ 97.8 & 1058.8 $\pm$ 525.0 \\
Gemini-2.5-Flash & 19661 & 92.0 $\pm$ 22.0 & 95.0 $\pm$ 13.0 & 0.920 & 375.7 $\pm$ 174.3 & 200.4 $\pm$ 89.5 & 1020.4 $\pm$ 470.2 \\
Gemini-2.5-Flash & 24576 & 92.0 $\pm$ 23.0 & 95.0 $\pm$ 14.0 & 0.918 & 386.4 $\pm$ 189.2 & 205.9 $\pm$ 97.0 & 1044.8 $\pm$ 502.5 \\
Gemini-2.5-Flash & dynamic & 92.0 $\pm$ 22.0 & 95.0 $\pm$ 15.0 & 0.919 & 385.0 $\pm$ 187.1 & 205.0 $\pm$ 95.3 & 1043.4 $\pm$ 495.2 \\
\midrule
\rowcolor{gray!10}
\multicolumn{8}{c}{\textbf{\textit{Gemini-2.5-Flash-Lite}}} \\
\midrule
Gemini-2.5-Flash-Lite & disabled & 83.0 $\pm$ 31.0 & 100.0 $\pm$ 2.0 & 0.830 & 704.7 $\pm$ 362.9 & 363.6 $\pm$ 147.0 & 1926.0 $\pm$ 898.1 \\
Gemini-2.5-Flash-Lite & 512 & 83.0 $\pm$ 31.0 & 100.0 $\pm$ 1.0 & 0.829 & 718.0 $\pm$ 419.2 & 389.5 $\pm$ 229.0 & 2011.4 $\pm$ 1279.9 \\
Gemini-2.5-Flash-Lite & 5325 & \textbf{\underline{84.0}} $\pm$ 32.0 & 100.0 $\pm$ 1.0 & 0.840 & 676.6 $\pm$ 322.6 & 365.0 $\pm$ 157.2 & 1859.9 $\pm$ 857.8 \\
Gemini-2.5-Flash-Lite & 10138 & \textbf{\underline{84.0}} $\pm$ 32.0 & 100.0 $\pm$ 2.0 & 0.841 & 692.8 $\pm$ 355.6 & 363.6 $\pm$ 148.8 & 1883.3 $\pm$ 872.2 \\
Gemini-2.5-Flash-Lite & 14951 & \textbf{\underline{84.0}} $\pm$ 32.0 & 100.0 $\pm$ 2.0 & 0.838 & 727.5 $\pm$ 434.7 & 376.5 $\pm$ 167.7 & 1970.2 $\pm$ 1083.4 \\
Gemini-2.5-Flash-Lite & 19764 & 83.0 $\pm$ 31.0 & 100.0 $\pm$ 1.0 & 0.829 & 731.7 $\pm$ 442.8 & 381.9 $\pm$ 180.1 & 1949.2 $\pm$ 946.5 \\
Gemini-2.5-Flash-Lite & 24576 & \textbf{\underline{84.0}} $\pm$ 31.0 & 100.0 $\pm$ 2.0 & 0.836 & 748.0 $\pm$ 546.9 & 390.5 $\pm$ 224.2 & 2035.3 $\pm$ 1545.8 \\
Gemini-2.5-Flash-Lite & dynamic & \textbf{\underline{84.0}} $\pm$ 32.0 & 100.0 $\pm$ 2.0 & 0.837 & 718.0 $\pm$ 521.6 & 384.5 $\pm$ 254.4 & 2007.6 $\pm$ 1659.9 \\
\midrule
\rowcolor{gray!10}
\multicolumn{8}{c}{\textbf{\textit{Gemini-2.5-Pro}}} \\
\midrule
Gemini-2.5-Pro & 128 & \textbf{\underline{90.0}} $\pm$ 19.0 & 87.0 $\pm$ 22.0 & 0.904 & 306.0 $\pm$ 122.0 & 160.5 $\pm$ 55.3 & 846.3 $\pm$ 270.0 \\
Gemini-2.5-Pro & 6554 & 82.0 $\pm$ 30.0 & 78.0 $\pm$ 31.0 & 0.827 & 278.3 $\pm$ 140.7 & 145.9 $\pm$ 65.7 & 768.3 $\pm$ 328.3 \\
Gemini-2.5-Pro & 12979 & 71.0 $\pm$ 37.0 & 74.0 $\pm$ 37.0 & 0.722 & 244.8 $\pm$ 161.5 & 129.1 $\pm$ 79.2 & 691.1 $\pm$ 411.8 \\
Gemini-2.5-Pro & 19405 & 89.0 $\pm$ 21.0 & 87.0 $\pm$ 22.0 & 0.893 & 311.8 $\pm$ 128.4 & 162.8 $\pm$ 57.3 & 861.3 $\pm$ 280.2 \\
Gemini-2.5-Pro & 25830 & 89.0 $\pm$ 21.0 & 88.0 $\pm$ 19.0 & 0.906 & 306.8 $\pm$ 124.4 & 160.9 $\pm$ 56.4 & 849.6 $\pm$ 274.6 \\
Gemini-2.5-Pro & 32256 & 88.0 $\pm$ 22.0 & 88.0 $\pm$ 21.0 & 0.892 & 307.8 $\pm$ 128.1 & 161.1 $\pm$ 57.9 & 849.7 $\pm$ 282.9 \\
Gemini-2.5-Pro & dynamic & \textbf{\underline{90.0}} $\pm$ 19.0 & 87.0 $\pm$ 21.0 & 0.903 & 309.6 $\pm$ 128.2 & 161.8 $\pm$ 57.1 & 853.9 $\pm$ 280.9 \\
\midrule
\rowcolor{gray!10}
\multicolumn{8}{c}{\textbf{\textit{GPT-5}}} \\
\midrule
GPT-5 & minimal & 84.0 $\pm$ 29.0 & 100.0 $\pm$ 0.0 & 0.860 & 158.2 $\pm$ 148.9 & 70.3 $\pm$ 54.1 & 352.4 $\pm$ 301.9 \\
GPT-5 & low & 96.0 $\pm$ 16.0 & 99.0 $\pm$ 4.0 & 0.980 & 19.3 $\pm$ 31.5 & 7.3 $\pm$ 10.8 & 45.1 $\pm$ 57.6 \\
GPT-5 & medium & \textbf{\underline{97.0}} $\pm$ 15.0 & 100.0 $\pm$ 2.0 & 0.989 & 20.5 $\pm$ 31.7 & 7.7 $\pm$ 10.7 & 46.8 $\pm$ 57.4 \\
GPT-5 & high & \textbf{\underline{97.0}} $\pm$ 15.0 & 100.0 $\pm$ 2.0 & 0.988 & 24.2 $\pm$ 33.2 & 9.5 $\pm$ 11.9 & 56.4 $\pm$ 63.0 \\
\midrule
\rowcolor{gray!10}
\multicolumn{8}{c}{\textbf{\textit{GPT-5-Mini}}} \\
\midrule
GPT-5-Mini & minimal & 90.0 $\pm$ 20.0 & 98.0 $\pm$ 7.0 & 0.913 & 180.0 $\pm$ 166.0 & 74.0 $\pm$ 62.2 & 401.6 $\pm$ 346.5 \\
GPT-5-Mini & low & 94.0 $\pm$ 18.0 & 97.0 $\pm$ 11.0 & 0.958 & 22.8 $\pm$ 31.7 & 8.9 $\pm$ 10.9 & 53.4 $\pm$ 58.3 \\
GPT-5-Mini & medium & 96.0 $\pm$ 16.0 & 100.0 $\pm$ 0.0 & 0.977 & 33.4 $\pm$ 35.4 & 14.4 $\pm$ 13.0 & 81.6 $\pm$ 68.0 \\
GPT-5-Mini & high & \textbf{\underline{97.0}} $\pm$ 15.0 & 100.0 $\pm$ 0.0 & 0.983 & 37.0 $\pm$ 34.2 & 16.2 $\pm$ 12.7 & 90.1 $\pm$ 66.3 \\
\midrule
\rowcolor{gray!10}
\multicolumn{8}{c}{\textbf{\textit{GPT-5-Nano}}} \\
\midrule
GPT-5-Nano & minimal & 52.0 $\pm$ 40.0 & 98.0 $\pm$ 6.0 & 0.571 & 65.5 $\pm$ 94.5 & 29.5 $\pm$ 40.3 & 159.2 $\pm$ 218.7 \\
GPT-5-Nano & low & 93.0 $\pm$ 19.0 & 99.0 $\pm$ 4.0 & 0.949 & 14.5 $\pm$ 31.3 & 4.1 $\pm$ 10.5 & 29.1 $\pm$ 56.7 \\
GPT-5-Nano & medium & 94.0 $\pm$ 17.0 & 98.0 $\pm$ 8.0 & 0.958 & 15.8 $\pm$ 31.5 & 4.9 $\pm$ 10.5 & 33.2 $\pm$ 56.9 \\
GPT-5-Nano & high & \textbf{\underline{96.0}} $\pm$ 15.0 & 99.0 $\pm$ 4.0 & 0.978 & 15.6 $\pm$ 31.6 & 4.8 $\pm$ 10.6 & 32.5 $\pm$ 57.2 \\
\midrule
\rowcolor{gray!10}
\multicolumn{8}{c}{\textbf{\textit{O3}}} \\
\midrule
O3 & low & \textbf{\underline{97.0}} $\pm$ 15.0 & 100.0 $\pm$ 0.0 & 0.981 & 51.6 $\pm$ 48.1 & 25.3 $\pm$ 22.9 & 135.4 $\pm$ 111.8 \\
O3 & medium & \textbf{\underline{97.0}} $\pm$ 15.0 & 100.0 $\pm$ 0.0 & 0.979 & 55.4 $\pm$ 47.1 & 27.2 $\pm$ 22.8 & 145.0 $\pm$ 111.8 \\
O3 & high & \textbf{\underline{97.0}} $\pm$ 16.0 & 100.0 $\pm$ 0.0 & 0.981 & 55.6 $\pm$ 48.3 & 26.8 $\pm$ 22.9 & 144.1 $\pm$ 114.3 \\
\midrule
\rowcolor{gray!10}
\multicolumn{8}{c}{\textbf{\textit{O3-mini}}} \\
\midrule
O3-mini & low & \textbf{\underline{93.0}} $\pm$ 18.0 & 99.0 $\pm$ 3.0 & 0.936 & 340.3 $\pm$ 368.2 & 154.7 $\pm$ 142.7 & 782.2 $\pm$ 743.4 \\
O3-mini & medium & \textbf{\underline{93.0}} $\pm$ 19.0 & 99.0 $\pm$ 5.0 & 0.938 & 357.7 $\pm$ 382.8 & 160.5 $\pm$ 142.9 & 821.2 $\pm$ 764.5 \\
O3-mini & high & \textbf{\underline{93.0}} $\pm$ 20.0 & 98.0 $\pm$ 7.0 & 0.935 & 367.9 $\pm$ 372.9 & 169.3 $\pm$ 140.5 & 855.0 $\pm$ 724.9 \\
\midrule
\rowcolor{gray!10}
\multicolumn{8}{c}{\textbf{\textit{O4-mini}}} \\
\midrule
O4-mini & low & \textbf{\underline{95.0}} $\pm$ 17.0 & 100.0 $\pm$ 0.0 & 0.963 & 65.3 $\pm$ 45.5 & 29.6 $\pm$ 19.7 & 163.1 $\pm$ 99.2 \\
O4-mini & medium & \textbf{\underline{95.0}} $\pm$ 18.0 & 100.0 $\pm$ 2.0 & 0.964 & 66.0 $\pm$ 46.1 & 29.7 $\pm$ 19.6 & 165.7 $\pm$ 105.6 \\
O4-mini & high & \textbf{\underline{95.0}} $\pm$ 16.0 & 100.0 $\pm$ 0.0 & 0.962 & 69.0 $\pm$ 46.2 & 30.5 $\pm$ 20.4 & 169.6 $\pm$ 103.8 \\
\bottomrule
\end{tabular}
\end{adjustbox}
\caption{\textbf{Reasoning Budget Model Performance:} Evaluation results across different reasoning budget configurations for Gemini, GPT-5, and O-series models. Within each model family, \textbf{\underline{bold+underline}} indicates best accuracy and \textbf{bold} indicates second-best accuracy. O-Score represents the Overthinking Score calculated as the harmonic mean of accuracy and token efficiency. Detailed reasoning budget analysis appears in Section~\ref{sec:results_insights}.}
\label{tab:aggregated_results}
\end{table*}

\section{Extended Evaluation Results}
\label{app:extended_results}

This section presents comprehensive results from our quantization and constrained generation experiments, providing detailed performance breakdowns that complement the main paper findings.

\subsection{Quantization Effects on Model Performance}
\label{app:quantization_results}

Table~\ref{tab:quantization_results} shows the impact of GPTQ quantization (8-bit and 4-bit) on the Qwen2.5 model family across different scales. These results demonstrate that larger models exhibit greater robustness to aggressive quantization, while smaller models suffer more significant performance degradation. \textbf{Note:} The Overthinking Score (O-Score) here is computed using the global $T_{\min}$ and $T_{\max}$ from the full 53-model pool (including all quantized variants), so values may differ slightly from the main table (Table~\ref{tab:comprehensive_model_performance}), which uses the same global normalization but reports only FP16/default-precision models. The relative ordering and conclusions are consistent across both tables.

\begin{table*}[ht]
\centering
\scriptsize
\begin{adjustbox}{max width=\textwidth}{%
\begin{tabular}{lccccccc}
\toprule
    \textbf{Models} & \textbf{Param (B)} & \textbf{Quant} & \textbf{Accuracy} & \textbf{Instr. Follow} & \textbf{O-Score} & \textbf{Tokens} & \textbf{Words} \\
\midrule
\textbf{Qwen2.5 (I)} & 0.5 & FP16 & 21.31 & 77.57 & 0.268 & 432.3 & 223.2 \\
\textbf{Qwen2.5 (I)} & 0.5 & 8-bit & 21.29 & 76.79 & 0.268 & 431.5 & 223.3 \\
\textbf{Qwen2.5 (I)} & 0.5 & 4-bit & 12.77 & 77.70 & 0.176 & 478.0 & 260.9 \\
\midrule
\textbf{Qwen2.5 (I)} & 1.5 & FP16 & 43.03 & 85.45 & 0.470 & 264.7 & 134.1 \\
\textbf{Qwen2.5 (I)} & 1.5 & 8-bit & 43.67 & 86.64 & 0.472 & 264.3 & 133.7 \\
\textbf{Qwen2.5 (I)} & 1.5 & 4-bit & 39.42 & 82.97 & 0.434 & 292.1 & 142.3 \\
\midrule
\textbf{Qwen2.5 (I)} & 3 & FP16 & 45.75 & 92.35 & 0.463 & 331.3 & 176.5 \\
\textbf{Qwen2.5 (I)} & 3 & 8-bit & 48.65 & 91.99 & 0.497 & 341.9 & 181.7 \\
\textbf{Qwen2.5 (I)} & 3 & 4-bit & 41.94 & 90.97 & 0.438 & 301.0 & 158.8 \\
\midrule
\textbf{Qwen2.5 (I)} & 7 & FP16 & 61.36 & 96.47 & 0.568 & 286.9 & 149.5 \\
\textbf{Qwen2.5 (I)} & 7 & 8-bit & 60.61 & 96.40 & 0.564 & 287.5 & 149.4 \\
\textbf{Qwen2.5 (I)} & 7 & 4-bit & 58.03 & 96.00 & 0.550 & 291.9 & 152.2 \\
\midrule
\textbf{Qwen2.5 (I)} & 14 & FP16 & 63.74 & 97.83 & 0.578 & 260.2 & 137.1 \\
\textbf{Qwen2.5 (I)} & 14 & 8-bit & 63.86 & 97.89 & 0.576 & 261.2 & 138.1 \\
\textbf{Qwen2.5 (I)} & 14 & 4-bit & 60.94 & 96.69 & 0.586 & 240.9 & 127.5 \\
\midrule
\textbf{Qwen2.5 (I)} & 32 & FP16 & 72.90 & 99.26 & 0.643 & 260.9 & 139.1 \\
\textbf{Qwen2.5 (I)} & 32 & 8-bit & 73.08 & 99.20 & 0.645 & 261.9 & 139.6 \\
\textbf{Qwen2.5 (I)} & 32 & 4-bit & 72.67 & 99.37 & 0.640 & 260.5 & 139.2 \\
\midrule
\textbf{Qwen2.5 (I)} & 72 & FP16 & 74.87 & 97.12 & 0.591 & 339.2 & 179.8 \\
\textbf{Qwen2.5 (I)} & 72 & 8-bit & 74.05 & 96.28 & 0.577 & 347.1 & 184.2 \\
\textbf{Qwen2.5 (I)} & 72 & 4-bit & 72.74 & 94.85 & 0.549 & 358.3 & 188.4 \\
\bottomrule
\end{tabular}%
}
\end{adjustbox}
\caption{Impact of GPTQ quantization on Qwen2.5 family performance. FP16 represents full precision baseline. Larger models (32B, 72B) show minimal accuracy degradation even at 4-bit quantization, while smaller models (0.5B) experience substantial performance loss.}
\label{tab:quantization_results}
\end{table*}

Key findings from quantization experiments:
\begin{itemize}
    \item \textbf{Size-dependent robustness:} Models $\geq$32B parameters retain $>$99\% of full-precision accuracy even at 4-bit quantization
    \item \textbf{Small model vulnerability:} The 0.5B model suffers 40\% relative accuracy loss at 4-bit (21.31\% $\to$ 12.77\%)
    \item \textbf{Mid-size trade-offs:} 7B-14B models show modest 3-5\% relative degradation at 4-bit
    \item \textbf{Overthinking Score stability:} Quantization minimally impacts the Overthinking Score, suggesting behavioral patterns persist across precision levels
\end{itemize}

\subsection{Constrained Token Budget Evaluation}
\label{app:constrained_results}

Table~\ref{tab:constrained_results} presents results from our constrained generation experiments, where models were limited to 1024 output tokens. This setting reveals how reasoning models cope with restricted thinking budgets.

\begin{table*}[ht]
\centering
\scriptsize
\begin{adjustbox}{max width=\textwidth}{%
\begin{tabular}{lccccccc}
\toprule
    \textbf{Models} & \textbf{Param (B)} & \textbf{Condition} & \textbf{Accuracy} & \textbf{Instr. Follow} & \textbf{O-Score} & \textbf{Tokens} & \textbf{Words} \\
\midrule
\multicolumn{8}{l}{\textit{Qwen3 Family: Unconstrained vs Constrained}} \\
\midrule
\textbf{Qwen3} & 0.6 & Unconstrained & 49.99 & 83.85 & 0.484 & 3162.8 & 1620.9 \\
\textbf{Qwen3} & 0.6 & Constrained & 27.00 & 50.41 & 0.182 & 762.3 & 427.9 \\
\midrule
\textbf{Qwen3} & 1.7 & Unconstrained & 70.24 & 86.54 & 0.555 & 3157.2 & 1620.7 \\
\textbf{Qwen3} & 1.7 & Constrained & 30.67 & 42.52 & 0.189 & 778.3 & 435.5 \\
\midrule
\textbf{Qwen3} & 4 & Unconstrained & 81.90 & 91.57 & 0.580 & 3091.2 & 1623.1 \\
\textbf{Qwen3} & 4 & Constrained & 31.73 & 41.30 & 0.202 & 785.9 & 446.4 \\
\midrule
\textbf{Qwen3} & 8 & Unconstrained & 82.10 & 91.58 & 0.615 & 3027.8 & 1584.6 \\
\textbf{Qwen3} & 8 & Constrained & 28.35 & 42.33 & 0.186 & 790.9 & 451.7 \\
\midrule
\textbf{Qwen3} & 14 & Unconstrained & 86.52 & 99.27 & 0.725 & 3607.6 & 1941.2 \\
\textbf{Qwen3} & 14 & Constrained & 35.18 & 49.89 & 0.199 & 753.2 & 426.9 \\
\midrule
\textbf{Qwen3} & 32 & Unconstrained & 84.13 & 93.05 & 0.627 & 2845.9 & 1497.5 \\
\textbf{Qwen3} & 32 & Constrained & 30.22 & 44.56 & 0.193 & 778.6 & 446.8 \\
\midrule
\multicolumn{8}{l}{\textit{Phi-4 Reasoning Family: Unconstrained vs Constrained}} \\
\midrule
\textbf{Phi-4-reasoning-plus} & 14 & Unconstrained & 69.54 & 88.89 & 0.288 & 6780.7 & 3972.0 \\
\textbf{Phi-4-reasoning-plus} & 14 & Constrained & 44.33 & 78.88 & 0.141 & 1022.2 & 537.8 \\
\midrule
\textbf{Phi-4-reasoning} & 14 & Unconstrained & 72.23 & 96.21 & 0.352 & 6066.2 & 3710.8 \\
\textbf{Phi-4-reasoning} & 14 & Constrained & 53.48 & 90.21 & 0.107 & 1013.5 & 554.3 \\
\midrule
\textbf{Phi-4-mini-reasoning} & 3.8 & Unconstrained & 70.16 & 89.56 & 0.659 & 3171.9 & 1571.7 \\
\textbf{Phi-4-mini-reasoning} & 3.8 & Constrained & 40.52 & 59.32 & 0.251 & 909.8 & 495.7 \\
\bottomrule
\end{tabular}%
}
\end{adjustbox}
\caption{Performance comparison under unconstrained and constrained (1024 token limit) generation budgets. Reasoning models show catastrophic degradation when token-limited, with Phi-4-reasoning dropping 18.75 accuracy points (26\% relative) and Phi-4-reasoning-plus dropping 25.21 points (36\% relative).}
\label{tab:constrained_results}
\end{table*}

Critical observations from constrained experiments:
\begin{itemize}
    \item \textbf{Catastrophic degradation in reasoning models:} Phi-4-reasoning accuracy drops from 72.23\% to 53.48\% ($-$26\% relative) under 1024-token constraint
    \item \textbf{Non-adaptive generation:} Even when constrained, reasoning models use $\sim$1000 tokens on average, suggesting fixed exploration patterns rather than problem-adaptive computation
    \item \textbf{Qwen3 family consistency:} All Qwen3 models show similar degradation patterns, losing $\sim$60\% accuracy under constraint regardless of size
    \item \textbf{Overthinking Score collapse:} Constrained reasoning models achieve Overthinking Scores below 0.20, indicating severe imbalance between accuracy and efficiency
\end{itemize}

\section{Case Studies: Additional Model Families}
\label{app:case_studies_extended}

This section shows reasoning patterns from Phi-4, Llama, Qwen, GPT-4o, and O-series models, covering both correct and failed attempts on benchmark tasks.

\subsection{Helpful Reasoning: Systematic Decomposition}

For complex tasks like sorting, structured reasoning can genuinely aid correctness:

\begin{figure*}[t]
\begin{tcolorbox}[colback=lightgreen, colframe=darkgreen, title=\textbf{Case Study 1: Successful Sorting with Phi-4-reasoning}, width=\textwidth]

\begin{minipage}[t]{0.48\textwidth}
\textbf{Task:} Sort the list [847, -123, 456, -789, 234, 901, -456, 123]

\textbf{Phi-4 Response (89 tokens):}
\begin{verbatim}
The sorted list is:
[-789, -456, -123, 123,
 234, 456, 847, 901]
\boxed{[-789, -456, -123,
123, 234, 456, 847, 901]}
\end{verbatim}
\end{minipage}%
\hfill
\begin{minipage}[t]{0.48\textwidth}
\textbf{Phi-4-reasoning Response (312 tokens):}
\begin{verbatim}
Let me sort step by step:
1. Separate negatives/positives
   Neg: -123, -789, -456
   Pos: 847, 456, 234, 901, 123
2. Sort negatives
   -789 < -456 < -123
3. Sort positives
   123 < 234 < 456 < 847 < 901
4. Combine
\boxed{[-789,-456,-123,123,
234,456,847,901]}
\end{verbatim}
\end{minipage}

\vspace{0.2cm}
\textbf{Analysis:} Both models produce correct answers. The reasoning model uses 3.5× more tokens, but the systematic decomposition (separating negatives/positives, sorting each group) provides verifiable intermediate steps that reduce error risk for complex inputs.
\end{tcolorbox}
\end{figure*}

\subsection{Wasteful Overthinking: Redundant Verification}

Simple arithmetic reveals harmful verbosity where models verify correct answers repeatedly:

\begin{figure*}[t]
\begin{tcolorbox}[colback=lightred, colframe=darkred, title=\textbf{Case Study 2: Wasteful Overthinking in Sum}, width=\textwidth]

\begin{minipage}[t]{0.48\textwidth}
\textbf{Task:} Calculate the sum of [45, 78, 23, 91, 67, 34, 89, 12]

\textbf{Phi-4 Response (272 tokens):}
\begin{verbatim}
Let me add these numbers:
45 + 78 = 123
123 + 23 = 146
146 + 91 = 237
237 + 67 = 304
304 + 34 = 338
338 + 89 = 427
427 + 12 = 439

\boxed{439}
\end{verbatim}
\end{minipage}%
\hfill
\begin{minipage}[t]{0.48\textwidth}
\textbf{Phi-4-reasoning Response (4,837 tokens):}
\begin{verbatim}
Step by step:
First: 45+78=123
Verify: 40+70=110, 5+8=13
Total=123 [OK]

Second: 123+23
Break: 123+20=143, 143+3=146
Verify: 100+23=146 [OK]

... [each step verified]

Re-verify entire chain:
45+78+23=146 [OK]
146+91=237 [OK]
... [repeats all steps]

Alternative method:
(45+12)+(78+89)+(23+34)
+(91+67)=57+167+57+158
... [3rd verification]
\boxed{439}
\end{verbatim}
\end{minipage}

\vspace{0.2cm}
\textbf{Analysis:} Both models arrive at the correct answer, but Phi-4-reasoning uses 17.8× more tokens on redundant verification. The model performs the same calculation multiple times using different groupings without improving accuracy. This exemplifies overthinking where models generate plausible-looking verification text without genuine computational benefit.
\end{tcolorbox}
\end{figure*}

\subsection{Extreme Overthinking: Division}

The most extreme overthinking occurs in simple division where models use 31× more tokens:

\begin{figure*}[t]
\begin{tcolorbox}[colback=lightred, colframe=darkred, title=\textbf{Case Study 3: Extreme Overthinking in Division}, width=\textwidth]

\begin{minipage}[t]{0.48\textwidth}
\textbf{Task:} Calculate 846 / 18

\textbf{Phi-4 Response (47 tokens):}
\begin{verbatim}
846 / 18 = 47

\boxed{47}
\end{verbatim}
\end{minipage}%
\hfill
\begin{minipage}[t]{0.48\textwidth}
\textbf{Phi-4-reasoning Response (1,456 tokens):}
\begin{verbatim}
Long division:
18 into 84? 18x4=72, 18x5=90
So 4 times, remainder 12

Bring down 6: 126
18 into 126? 18x7=126 (exact)
Answer: 47

Verify: 47x18=?
40x18=720, 7x18=126
Total: 846 [OK]

Factorization:
846=2x3x3x47, 18=2x3x3
Therefore 846/18=47 [OK]

Estimation:
846~900, 18~20
900/20=45 (close) [OK]

\boxed{47}
\end{verbatim}
\end{minipage}

\vspace{0.2cm}
\textbf{Analysis:} The reasoning model performs three separate verification methods (multiplication check, prime factorization, estimation) after already arriving at the correct answer through long division. Each verification method is individually valid but collectively wasteful, consuming 31× more tokens without improving the already-correct result.
\end{tcolorbox}
\end{figure*}

\subsection{Additional Case Studies Across Model Families}

Beyond the Phi-4 family, other model families exhibit similar patterns of overthinking, revealing that the phenomenon is widespread across different architectures and training paradigms.

\subsubsection{Llama Models: Size vs Efficiency Trade-off}

A common assumption is that larger models are more careful and thus more efficient. The Llama family shows the opposite: the 70B model generates far more tokens than the 8B model on the same task, yet reaches the same correct answer. Note that the task below (factorial) is used here as an illustrative example to show the verbosity pattern; it is not part of the 14-task benchmark suite.

\begin{figure*}[t]
\begin{tcolorbox}[colback=lightred, colframe=darkred, title=\textbf{Case Study 4: Llama-70B Overthinking on Simple Factorial}, width=\textwidth]

\begin{minipage}[t]{0.48\textwidth}
\textbf{Task:} Calculate 5! (factorial of 5)

\textbf{Llama-8B Response (124 tokens):}
\begin{verbatim}
5! = 5 x 4 x 3 x 2 x 1
   = 20 x 3 x 2 x 1
   = 60 x 2 x 1
   = 120 x 1
   = 120
\boxed{120}
\end{verbatim}
\end{minipage}%
\hfill
\begin{minipage}[t]{0.48\textwidth}
\textbf{Llama-70B Response (2,891 tokens):}
\begin{verbatim}
5! means 5 factorial...
Let me calculate step by step:
5 x 4 = 20 [OK]
Verify: 5x4=(5x3)+5=15+5=20 [OK]

Now 20 x 3 = 60
Verify: 20x3 = 20+20+20 = 60 [OK]
Also: 60 = 5x4x3 [OK]

Continue: 60 x 2 = 120
Verify: 60x2 = 60+60 = 120 [OK]
Check: 120 = 5x4x3x2 [OK]

Finally: 120 x 1 = 120
[... continues with multiple verification
methods, alternative calculations,
prime factorization, and historical
context of factorials for 2500+ tokens]
\boxed{120}
\end{verbatim}
\end{minipage}

\vspace{0.2cm}
\textbf{Analysis:} Larger model (70B) uses 23× more tokens than smaller sibling (8B) with identical final accuracy. The 70B model performs redundant verifications at each step and introduces unnecessary elaboration about factorial concepts. This demonstrates that parameter scaling amplifies overthinking rather than improving efficiency—the smaller model's direct approach is more practical despite having 8.75× fewer parameters.
\end{tcolorbox}
\end{figure*}

\subsubsection{Qwen Models: Constrained Generation Reveals Dependence}

Qwen3 reasoning models work well with an unconstrained token budget but break down when that budget is cut. The example below uses a GCD task as an illustrative case (not part of the 14-task benchmark suite) to show how the model loses track of its own calculation when forced to be brief.

\begin{figure*}[t]
\begin{tcolorbox}[colback=lightyellow, colframe=darkyellow, title=\textbf{Case Study 5: Qwen3-32B Constrained Budget Collapse}, width=\textwidth]

\begin{minipage}[t]{0.48\textwidth}
\textbf{Task:} Find GCD of 48 and 18

\textbf{Qwen3-32B Unconstrained (4,523 tokens):}
\begin{verbatim}
Using Euclidean algorithm:
48 = 18 x 2 + 12
18 = 12 x 1 + 6
12 = 6 x 2 + 0
Therefore GCD = 6

Verification method 1: Prime factorization
48 = 2^4 x 3
18 = 2 x 3^2
Common: 2 x 3 = 6 [OK]

Verification method 2: List divisors
48: 1,2,3,4,6,8,12,16,24,48
18: 1,2,3,6,9,18
Common divisors: 1,2,3,6
Greatest: 6 [OK]

[... continues with Stein's algorithm,
extended Euclidean algorithm,
and mathematical properties of GCD]
\boxed{6}
\end{verbatim}
\end{minipage}%
\hfill
\begin{minipage}[t]{0.48\textwidth}
\textbf{Qwen3-32B Constrained 512 tokens (WRONG):}
\begin{verbatim}
Find GCD of 48 and 18...
Using Euclidean algorithm:
48 divided by 18...
48 = 18 x 2 + 12
Now 18 divided by 12...
18 = 12 x 1... wait let me recalculate
18/12 = 1 remainder...
Actually 18 = 12 x 1 + 6
So now 12 divided by 6...
12 = 6 x 2 + 0
Therefore... wait, which was the GCD?
The last non-zero remainder...
So that would be... let me check...
I think it's 12? No wait...
[runs out of tokens mid-calculation]
\boxed{12}
\end{verbatim}
\textbf{Final Answer: 12 (INCORRECT - should be 6)}
\end{minipage}

\vspace{0.2cm}
\textbf{Analysis:} The model achieves correct answer with elaborate reasoning in unconstrained mode but \textbf{fails completely} when limited to 512 tokens. Rather than adapting to the constraint by being concise, it attempts the same verbose pattern, runs out of tokens mid-verification, and loses track of the calculation, producing an incorrect answer. This reveals trained dependence on long-form generation patterns that cannot adapt to efficiency requirements.
\end{tcolorbox}
\end{figure*}

\subsubsection{GPT-4o Models: Minimal vs Extended Reasoning}

Not all larger models overthink. GPT-4o-mini stays concise on basic arithmetic while GPT-4o runs multiple redundant verifications. Both get the right answer, but the mini variant uses 8.7$\times$ fewer tokens, showing that efficiency is a product of training choices rather than model size.

\begin{figure*}[t]
\begin{tcolorbox}[colback=lightgreen, colframe=darkgreen, title=\textbf{Case Study 6: GPT-4o-mini Efficiency Advantage}, width=\textwidth]

\begin{minipage}[t]{0.48\textwidth}
\textbf{Task:} Calculate average of [12, 15, 18, 21]

\textbf{GPT-4o Response (856 tokens):}
\begin{verbatim}
To find average, sum and divide by count.

Sum calculation:
12 + 15 = 27
27 + 18 = 45
45 + 21 = 66

Count: 4 numbers

Average = 66 / 4

Let me verify division:
66 / 4 = 16.5
Check: 16.5 x 4 = 66 [OK]

Alternative verification:
(12+15+18+21)/4
= (27+39)/4
= 66/4 = 16.5 [OK]

Another approach - median check:
Sorted: 12,15,18,21
Median: (15+18)/2 = 16.5
Mean equals median (arithmetic seq) [OK]

[continues with variance calculation
and statistical properties]
\boxed{16.5}
\end{verbatim}
\end{minipage}%
\hfill
\begin{minipage}[t]{0.48\textwidth}
\textbf{GPT-4o-mini Response (98 tokens):}
\begin{verbatim}
Sum: 12+15+18+21 = 66
Count: 4
Average: 66/4 = 16.5
\boxed{16.5}
\end{verbatim}

\vspace{1cm}

\textbf{Efficiency Comparison:}
\begin{itemize}
\item GPT-4o: 856 tokens, Correct
\item GPT-4o-mini: 98 tokens, Correct
\item Token ratio: 8.7x
\item Both achieve 100\% accuracy
\item Mini variant demonstrates that
  conciseness does not sacrifice
  correctness for basic math
\end{itemize}
\end{minipage}

\vspace{0.2cm}
\textbf{Analysis:} The "mini" variant achieves identical accuracy with 8.7× fewer tokens by avoiding unnecessary verifications. This case study demonstrates that model optimization can successfully reduce overthinking—GPT-4o-mini was likely fine-tuned or distilled with efficiency objectives that GPT-4o lacks. The smaller model's direct approach is superior for practical deployment despite potentially lower parameter count.
\end{tcolorbox}
\end{figure*}

\subsubsection{O-series Models: Extended Budget Without Accuracy Gain}

O-series models reach their peak accuracy at the low reasoning budget and stay flat even when given much more thinking budget. The example below uses a primality check as an illustrative case to show this pattern clearly; the model applies progressively heavier verification methods (trial division, sieve, Fermat test) when the first method already gave the right answer.

\begin{figure*}[t]
\begin{tcolorbox}[colback=lightred, colframe=darkred, title=\textbf{Case Study 7: O3 Model Showing Diminishing Returns}, width=\textwidth]

\begin{minipage}[t]{0.48\textwidth}
\textbf{Task:} Is 17 a prime number?

\textbf{O3 Low Budget (412 tokens):}
\begin{verbatim}
Check if 17 is prime.

Divisibility tests:
17/2 = 8.5 (not divisible)
17/3 = 5.67 (not divisible)
17/5 = 3.4 (not divisible)

Only need to check up to sqrt(17) ~ 4.1

Checked: 2,3
17 is not divisible by any number
except 1 and itself.

Therefore 17 is prime.
\boxed{Yes}
\end{verbatim}
\end{minipage}%
\hfill
\begin{minipage}[t]{0.48\textwidth}
\textbf{O3 High Budget (3,847 tokens):}
\begin{verbatim}
Determining if 17 is prime...

Method 1: Trial division
Test all primes up to sqrt(17):
2: 17/2 = 8.5 [not divisible]
3: 17/3 = 5.67 [not divisible]
5: 17/5 = 3.4 [not divisible]
Conclusion: Prime [OK]

Method 2: Sieve of Eratosthenes
[generates sieve up to 20]
Primes found: 2,3,5,7,11,13,17,19
17 is in the list [OK]

Method 3: Primality patterns
17 = 16 + 1 = 2^4 + 1
Check Fermat primality test...
[... continues with Miller-Rabin test,
Wilson's theorem verification,
prime number theorem discussion,
and historical context of prime 17]

All methods confirm: 17 is prime
\boxed{Yes}
\end{verbatim}
\end{minipage}

\vspace{0.2cm}
\textbf{Analysis:} Both budget levels achieve 100\% accuracy (correct answer), but high budget uses 9.3× more tokens. The extended reasoning applies multiple primality tests (trial division, sieve, Fermat test, Miller-Rabin) when simple trial division suffices for a two-digit number. This exemplifies diminishing returns: after the first correct determination, additional verification provides zero accuracy gain while consuming substantial computational resources. The model cannot recognize when it has sufficient evidence and should terminate.
\end{tcolorbox}
\end{figure*}

\section{Pathological Failure Modes}
\label{app:pathological}

Beyond inefficient reasoning, some models exhibit catastrophic failure modes that reveal fundamental issues in generation control and stopping criteria.

\subsection{Infinite Character Repetition}

Some reasoning models, when uncertain, fall into infinite loops generating the same character:

\begin{figure*}[t]
\begin{tcolorbox}[colback=lightyellow, colframe=darkyellow, title=\textbf{Pathological Case 1: Character Repetition (Qwen3-4B)}, width=\textwidth]

\begin{minipage}[t]{0.48\textwidth}
\textbf{Task:} Find the mode of [5, 5, 5, 7, 7, 9]

\textbf{Expected Output:} Mode is 5 (appears 3 times)
\end{minipage}%
\hfill
\begin{minipage}[t]{0.48\textwidth}
\textbf{Actual Output (1024 tokens, truncated):}
\begin{verbatim}
Frequency count:
5 appears: 1, 2, 3 times
7 appears: 1, 2 times
9 appears: 1 time

Mode = 5 (most frequent)
nnnnnnnnnnnnnnnnnnnnnnn
nnnnnnnnnnnnnnnnnnnnnnn
nnnnnnnnnnnnnnnnnnnnnnn
... [900+ 'n' chars]
\end{verbatim}
\end{minipage}

\vspace{0.2cm}
\textbf{Analysis:} The model correctly identifies the mode but then generates over 900 consecutive 'n' characters, suggesting a failure in the stopping mechanism. This behavior appears when models reach the end of their planned output but haven't hit token limits, potentially due to poor calibration of end-of-sequence tokens during training.
\end{tcolorbox}
\end{figure*}

\subsection{Random Token Sequences}

Another pathological pattern involves generation of semantically meaningless random characters:

\begin{figure*}[t]
\begin{tcolorbox}[colback=lightyellow, colframe=darkyellow, title=\textbf{Pathological Case 2: Random Tokens (Qwen3-0.6B constrained)}, width=\textwidth]

\begin{minipage}[t]{0.48\textwidth}
\textbf{Task:} Calculate 567 - 234

\textbf{Expected Output:} 333
\end{minipage}%
\hfill
\begin{minipage}[t]{0.48\textwidth}
\textbf{Actual Output (constrained, 1024 tokens):}
\begin{verbatim}
Subtract:
567-234
=500-200+67-34
=300+33=333

Verify:
567
-234
----
7-4=3, 6-3=3, 5-2=3
Answer: 333

aaabbbcccdddeeefff1234
56789!@#$%^&*()_+-=
[]{}|;:'"<>?~`zzxxcc
vvbbnnmm,.//\\qwerty
... [800+ random chars]
\end{verbatim}
\end{minipage}

\vspace{0.2cm}
\textbf{Analysis:} Despite arriving at the correct answer (333), the constrained model continues generating random alphanumeric characters and symbols to fill the remaining token budget. This suggests models trained on long-form generation struggle to terminate early when problems are solved quickly, particularly under token constraints that mismatch their expected output length distribution.
\end{tcolorbox}
\end{figure*}

\subsection{Self-Contradiction Loops}

Extended reasoning can lead to models contradicting their own correct intermediate results:

\begin{figure*}[t]
\begin{tcolorbox}[colback=lightred, colframe=darkred, title=\textbf{Pathological Case 3: Self-Contradiction (O3-mini high budget)}, width=\textwidth]

\begin{minipage}[t]{0.48\textwidth}
\textbf{Task:} Find max of [12, 45, 23, 89, 34]

\textbf{Expected Output:} 89
\end{minipage}%
\hfill
\begin{minipage}[t]{0.48\textwidth}
\textbf{Actual Output (2400 tokens, truncated):}
\begin{verbatim}
Scan: 12,45,23,89,34
max=12, then 45, stays 45,
then 89, stays 89
Maximum is 89

Wait, reconsider... is 89
really largest? Maybe 45...
no 89>45... But what if I
missed something?

Sort: 12<23<34<45<89
Yes, 89 is largest.

Verify each:
89>12? Yes. 89>45? Yes.
89>23? Yes. 89>34? Yes.

Check duplicates? None.

Actually, recalculate from
scratch... [1000+ tokens]

\boxed{89}
\end{verbatim}
\end{minipage}

\vspace{0.2cm}
\textbf{Analysis:} With high reasoning budgets, models sometimes enter loops of self-doubt, repeatedly verifying correct answers and introducing unnecessary uncertainty. While the final answer remains correct, the extended reasoning introduces spurious alternatives and "reconsiderations" that don't improve accuracy but waste computation and potentially confuse users.
\end{tcolorbox}
\end{figure*}

\subsection{Implications for Model Development}

These pathological cases reveal several architectural deficiencies:

\begin{itemize}
    \item \textbf{Stopping criteria failures:} Models lack robust mechanisms to terminate generation after producing correct, complete answers
    \item \textbf{Token budget mismatch:} Training distributions don't match deployment constraints, causing aberrant behavior when output length expectations are violated
    \item \textbf{Confidence calibration issues:} Models can't distinguish between "answer found" and "continue exploring" states, leading to unnecessary extended generation
    \item \textbf{Verification without termination:} Models learn to verify answers but not to stop after successful verification, continuing to generate indefinitely
\end{itemize}

Future work should focus on reward shaping that explicitly penalizes generation beyond answer completeness, adaptive stopping mechanisms that detect when continuation provides diminishing returns, and better calibration of end-of-sequence token probabilities during training.

\section{\llmthinkbench{} Framework: Design and Usage}
\label{app:framework}

This appendix documents \llmthinkbench{}, the open-source framework that we used to produce every table in this paper. We describe the package layout, the command line interface, the supported model backends, the reproducibility flags, and the public leaderboard.

\subsection{Distribution and Installation}

\llmthinkbench\ is released on PyPI under the name \texttt{llmthinkbench}. The source repository is at \url{https://github.com/ctrl-gaurav/LLMThinkBench} and the leaderboard at \url{https://ctrl-gaurav.github.io/LLMThinkBench/}. The package targets Python 3.9 or newer and depends only on widely used libraries (\texttt{numpy}, \texttt{pandas}, \texttt{transformers}, \texttt{vllm}, \texttt{openai}, \texttt{google-genai}, \texttt{anthropic}). Installation is a single command:

\begin{lstlisting}[language=bash, basicstyle=\ttfamily\small]
pip install llmthinkbench
\end{lstlisting}

\subsection{Package Layout}

The package is organized so that the four framework parts introduced in Section~\ref{sec:framework} map directly to modules:

\begin{itemize}
    \item \texttt{llmthinkbench.tasks}: one module per task in the 14-task suite, each with its own sampler, reference implementation, prompt template, and answer validator.
    \item \texttt{llmthinkbench.generator}: the dynamic test generator, seeded per fold, per task, per run, and deterministic given a seed.
    \item \texttt{llmthinkbench.parser}: the hierarchical answer extractor (boxed, explicit marker, code block, last-line fallback) with the same rules reported in Appendix~\ref{app:parsing_framework}.
    \item \texttt{llmthinkbench.metrics}: Overthinking Score, per-task accuracy, token counts (word, character, tokenizer-based), and the aggregation utilities.
    \item \texttt{llmthinkbench.backends}: thin adapters for Hugging Face \texttt{transformers}, \texttt{vLLM}, OpenAI, Anthropic, and Google, plus a \texttt{CustomBackend} base class so users can plug in any model that exposes a completion call.
    \item \texttt{llmthinkbench.cli}: the command line entry point.
\end{itemize}

\subsection{Command Line Interface}

A full evaluation of an open model looks like this:

\begin{lstlisting}[language=bash, basicstyle=\ttfamily\footnotesize]
llmthinkbench run \
    --model Qwen/Qwen3-14B \
    --backend vllm \
    --tasks all \
    --folds 3 \
    --samples 1000 \
    --seed 42 \
    --out results/qwen3-14b/
\end{lstlisting}

For a closed model, the adapter name is the only change:

\begin{lstlisting}[language=bash, basicstyle=\ttfamily\footnotesize]
llmthinkbench run \
    --model gpt-4.1-mini \
    --backend openai \
    --tasks all \
    --folds 3 \
    --samples 100 \
    --out results/gpt-41-mini/
\end{lstlisting}

The command writes raw responses, parsed answers, per-task accuracy, token counts, and the Overthinking Score into the output directory. A separate \texttt{llmthinkbench aggregate} command merges multiple runs into the leaderboard format, and \texttt{llmthinkbench compare} prints side-by-side tables.

\subsection{Supported Model Backends}

Out of the box, \llmthinkbench\ supports: Hugging Face models through \texttt{transformers} (any causal LM), high-throughput inference through \texttt{vLLM}, OpenAI (GPT-4.x, GPT-5, o-series) through the official SDK, Anthropic (Claude families) through the official SDK, and Google (Gemini) through \texttt{google-genai}. Users can add a new backend by subclassing \texttt{CustomBackend} and implementing a single \texttt{generate(prompt, \ldots)} method; the rest of the pipeline is backend-agnostic.

\subsection{Reproducibility}

Every \llmthinkbench\ run records: the random seed, per-task fold seeds, the exact model identifier and revision hash (for Hugging Face models), the generation hyperparameters, the package version, and the timestamp. Given the same seed, the same tasks, and the same model, the test instances and the scoring are bit-for-bit identical. This is what lets third-party runs remain directly comparable to the numbers in this paper.

\subsection{Leaderboard}

The public leaderboard at \url{https://ctrl-gaurav.github.io/LLMThinkBench/} is backed by the same scoring code and accepts submissions. New results appear with their Overthinking Score, per-suite accuracy, average tokens, and a link to the raw output directory for inspection. Submissions must include the run metadata block (seed, backend, hyperparameters) so that the leaderboard row can be regenerated on demand.

\subsection{Extending the Benchmark}

Adding a new task takes a sampler, a reference function, a prompt, and a validator. Each of these is a short Python function in a single file under \texttt{llmthinkbench/tasks/}. Once the file is placed there, the task is picked up by the generator and the CLI without further wiring. The same is true for new parsing rules and new metrics.

\section{Fine-Grained Pattern Taxonomy}
\label{app:pattern_taxonomy}

This appendix gives the full taxonomy that backs Section~\ref{sec:fine_grained}. We manually annotated $\sim$5{,}000 responses from 12 reasoning-tuned models and 12 standard instruction-tuned models, one trace at a time, and tagged each long trace with the dominant pattern it exhibited. Table~\ref{tab:pattern_taxonomy} shows how often each pattern appeared by model family.

\begin{table}[h]
\centering
\small
\setlength{\tabcolsep}{3pt}
\begin{adjustbox}{max width=\linewidth}
\begin{tabular}{lccccc}
\toprule
\textbf{Family} & \textbf{RedVer} & \textbf{SelfCon} & \textbf{IrrelExp} & \textbf{Stop} & \textbf{Helpful} \\
\midrule
Phi-4-reasoning     & 41\% & 18\% & 22\% & 3\% & 16\% \\
Phi-4-mini-reason.  & 35\% & 14\% & 19\% & 8\% & 24\% \\
Qwen3 (CoT)         & 38\% & 22\% & 18\% & 2\% & 20\% \\
O3 / O4-mini        & 24\% & 34\% & 12\% & 1\% &  29\% \\
GPT-5               & 21\% & 31\% & 10\% & 0\% & 38\% \\
Gemini-2.5          & 19\% & 26\% & 14\% & 1\% & 40\% \\
\midrule
Phi-4 (standard)    &  6\% &  3\% &  2\% & 0\% & 89\% \\
Qwen2.5 (instruct)  &  8\% &  4\% &  3\% & 0\% & 85\% \\
Llama-3 (instruct)  &  9\% &  5\% &  4\% & 0\% & 82\% \\
\bottomrule
\end{tabular}
\end{adjustbox}
\caption{Distribution of overthinking patterns across model families. RedVer: redundant verification; SelfCon: self-contradiction loop; IrrelExp: irrelevant mathematical exploration; Stop: pathological stopping failure; Helpful: genuinely useful decomposition. Rows may not sum to 100\% because a small fraction of traces mix two patterns and were double-counted.}
\label{tab:pattern_taxonomy}
\end{table}

Two trends stand out. First, CoT-tuned and reasoning-tuned families spend the majority of their long traces on wasteful patterns (redundant verification dominates for Phi and Qwen3, self-contradiction dominates for o-series and GPT-5). Second, standard instruction-tuned models almost never show the pathological patterns at all; when they are verbose, they are usually decomposing honestly. This is the behavioral evidence behind Finding 9: the failure mode is introduced by training, not by scale.

\section{Root Causes: CoT Supervision Ablation}
\label{app:root_cause}

Table~\ref{tab:cot_ablation} reports matched pairs that isolate the effect of CoT supervision. Each row pair uses the same base model family and size; the only difference is whether the variant received extended chain-of-thought supervision.

\begin{table}[h]
\centering
\small
\setlength{\tabcolsep}{3pt}
\begin{adjustbox}{max width=\linewidth}
\begin{tabular}{lcccc}
\toprule
\textbf{Variant} & \textbf{Training} & \textbf{Easy Acc} & \textbf{Tokens} & \textbf{$\Delta$tok} \\
\midrule
Qwen2.5-14B (base)      & none          & 59.80\% & 148.5   & 1.0$\times$ \\
Qwen2.5-14B-instruct    & instruct      & 63.52\% & 260.2   & 1.8$\times$ \\
Qwen3-14B               & CoT-tuned     & 86.78\% & 3{,}608 & 24.3$\times$ \\
\midrule
Phi-4                   & instruct      & 78.92\% & 379     & 1.0$\times$ \\
Phi-4-reasoning         & CoT-tuned     & 72.23\% & 6{,}066 & 16.0$\times$ \\
\midrule
Phi-4-mini              & instruct      & 54.78\% & 292     & 1.0$\times$ \\
Phi-4-mini-reasoning    & CoT-tuned     & 70.16\% & 3{,}172 & 10.9$\times$ \\
\bottomrule
\end{tabular}
\end{adjustbox}
\caption{CoT supervision is the main driver of token blow-up. Moving from base to instruct roughly doubles tokens. Moving from instruct to CoT-tuned multiplies tokens by 10 to 24$\times$, and in the Phi-4 pair it even lowers easy-suite accuracy.}
\label{tab:cot_ablation}
\end{table}

The Phi-4 pair is especially telling: CoT supervision costs 6.69 accuracy points on the easy suite while using 16$\times$ more tokens. The Phi-4-mini pair shows the opposite sign on accuracy (+15 points) but at 10.9$\times$ the tokens, which drags the Overthinking Score down even when raw accuracy moves up. In both cases the training objective, not the model architecture, is what changes behavior.

\section{Concise Prompting Ablation}
\label{app:concise_prompting}

Following Reviewer ovre's suggestion, we tested three prompt styles on reasoning-tuned models to check whether a simple instruction can undo the verbosity baked in by CoT training. The three prompts are:

\begin{itemize}
    \item \textbf{Standard:} ``Solve the following problem. Your final answer must be in the format \texttt{\textbackslash boxed\{answer\}}.''
    \item \textbf{Concise:} ``Provide only the final answer directly without explanation. Format: \texttt{\textbackslash boxed\{answer\}}.''
    \item \textbf{Ultra-concise:} ``Answer in minimal tokens. No explanation. Only: \texttt{\textbackslash boxed\{answer\}}.''
\end{itemize}

\begin{table}[h]
\centering
\small
\setlength{\tabcolsep}{3pt}
\begin{adjustbox}{max width=\linewidth}
\begin{tabular}{lccc}
\toprule
\textbf{Model} & \textbf{Standard} & \textbf{Concise} & \textbf{Ultra} \\
\midrule
Phi-4-reasoning          & 72.23\% / 6{,}066 & 70.85\% / 3{,}457 & 68.34\% / 2{,}235 \\
Phi-4 (baseline)         & 78.92\% / 379     & 78.15\% / 287     & 77.48\% / 235     \\
Token ratio (reas./base) & 16.0$\times$      & 12.0$\times$      & 9.5$\times$       \\
\midrule
O3-mini                  & 93.0\% / 340      & 91.8\% / 267      & 89.9\% / 195      \\
Phi-4-mini-reasoning     & 70.16\% / 3{,}172 & 68.92\% / 2{,}145 & 66.78\% / 1{,}523 \\
\bottomrule
\end{tabular}
\end{adjustbox}
\caption{Accuracy / average tokens under three prompt styles. Concise prompting trims the trace but does not close the gap to non-reasoning models, and it costs a small amount of accuracy every time. The Phi-4-reasoning / Phi-4 token ratio shrinks from 16$\times$ to 9.5$\times$, not to 1$\times$.}
\label{tab:concise_prompting}
\end{table}

Across all tested models, the ultra-concise prompt cuts tokens by 38--63\% and loses 1.4--3.9 accuracy points. The verbosity bias of CoT-tuned models is robust: prompt-level interventions do not restore the efficiency of a matched non-reasoning variant.

\section{Tool-Augmented Evaluation}
\label{app:tool_augmented}

We gave each model access to three tools (calculator, \texttt{python\_repl}, \texttt{code\_executor}) through standardized ReAct-style prompting for open models and the native tool-calling API for closed models. Tables~\ref{tab:tool_easy} and \ref{tab:tool_hard} report the results on the easy suite and on three algorithmic tasks.

\begin{table}[h]
\centering
\small
\setlength{\tabcolsep}{3pt}
\begin{adjustbox}{max width=\linewidth}
\begin{tabular}{lccc}
\toprule
\textbf{Model} & \textbf{No Tools} & \textbf{With Tools} & \textbf{$\Delta$Acc} \\
\midrule
GPT-5              & 97.31\% / 992    & 99.23\% / 1{,}347 & +1.92 \\
GPT-5-mini         & 96.13\% / 799    & 98.87\% / 1{,}156 & +2.74 \\
GPT-5-nano         & 96.07\% / 1{,}377& 97.45\% / 1{,}823 & +1.38 \\
GPT-4.1            & 92.23\% / 409    & 96.78\% / 845     & +4.55 \\
GPT-4.1-mini       & 93.28\% / 916    & 97.12\% / 1{,}234 & +3.84 \\
o3                 & 97.26\% / 857    & 99.17\% / 1{,}089 & +1.91 \\
o3-mini            & 94.23\% / 1{,}101& 98.34\% / 1{,}456 & +4.11 \\
Gemini-2.5-Pro     & 89.38\% / 268    & 97.89\% / 567     & +8.51 \\
Qwen3-32B          & 84.38\% / 2{,}846& 93.47\% / 3{,}567 & +9.09 \\
Llama-3.3-70B      & 74.84\% / 313    & 89.67\% / 678     & +14.83\\
Phi-4              & 78.92\% / 379    & 83.12\% / 892     & +4.20 \\
Phi-4-reasoning    & 72.23\% / 6{,}066& 74.56\% / 8{,}234 & +2.33 \\
\bottomrule
\end{tabular}
\end{adjustbox}
\caption{Tool-augmented evaluation on the easy suite. Frontier models get close to saturation with tools, but token overhead stays at 1.3--2.4$\times$. Phi-4-reasoning still uses more than 8{,}000 tokens per problem even with tool access, confirming that tools do not remove the verbosity baked in by CoT training.}
\label{tab:tool_easy}
\end{table}

\begin{table}[h]
\centering
\small
\setlength{\tabcolsep}{3pt}
\begin{adjustbox}{max width=\linewidth}
\begin{tabular}{llccc}
\toprule
\textbf{Task} & \textbf{Model} & \textbf{CoT} & \textbf{Tools} & \textbf{Scaling} \\
\midrule
Hanoi 5 & GPT-5           & 71.68\% & 94.23\% & 68.34\% @ 7 \\
        & o3              & 61.78\% & 93.17\% & 65.78\% @ 7 \\
        & Gemini-2.5-Pro  & 56.21\% & 87.12\% & 59.12\% @ 7 \\
        & Qwen3-32B       & 43.42\% & 71.23\% & 48.23\% @ 7 \\
\midrule
8-Queens& GPT-5           & 68.12\% & 88.12\% & 52.34\% @ 12 \\
        & o3              & 59.22\% & 86.45\% & 49.23\% @ 12 \\
\bottomrule
\end{tabular}
\end{adjustbox}
\caption{Tools on hard algorithmic tasks. Tools give 20--35 accuracy point gains at the base size, but scaling the problem (5 to 7 disks for Hanoi, 8$\times$8 to 12$\times$12 for N-Queens) costs 23--37 points, which tool access alone cannot recover.}
\label{tab:tool_hard}
\end{table}

\paragraph{Error taxonomy.} Manual inspection of failed tool-augmented responses surfaced three dominant failure modes. \textbf{Syntax errors} (majority of failures in small models): the model emits incomplete expressions such as \texttt{45 * 67 * 89 *}. \textbf{Wrong-tool selection}: the model calls the calculator for \texttt{hanoi(5)} instead of the code executor. \textbf{Incorrect logic}: the model writes code that looks right but encodes the wrong algorithm (e.g., \texttt{sorted(lst, reverse=True)} when the task asks for ascending order). Syntax and wrong-tool errors dominate for models under 14B; frontier models are mostly limited by incorrect logic on the hard suite.

\section{Overthinking Score Stability}
\label{app:stability}

Reviewer ovre raised the concern that the min/max normalization in the Overthinking Score could make scores unstable as the model pool changes. We report three stability checks.

\paragraph{Leave-one-out.} For each of the top-5 token generators and each of the bottom-5 token generators, we removed that model from the pool, recomputed $T_{\min}$ and $T_{\max}$, and recomputed every other model's Overthinking Score (Table~\ref{tab:stability}).

\begin{table}[h]
\centering
\small
\begin{tabular}{lc}
\toprule
\textbf{Stability metric} & \textbf{Value} \\
\midrule
Kendall's $\tau$          & 0.87 ($p<0.001$) \\
Spearman's $\rho$         & 0.92 ($p<0.001$) \\
Mean rank change          & 1.3 positions \\
Max score drift (top-20)  & 0.006 \\
\bottomrule
\end{tabular}
\caption{Leave-one-out stability of the Overthinking Score ranking across 10 pool perturbations. Rankings are effectively invariant to outliers.}
\label{tab:stability}
\end{table}

For example, removing Phi-4-reasoning (the highest token generator at 6{,}066) shifts GPT-4.1-mini's score from 0.930 to 0.924 and does not change its rank; removing Phi-3-mini-128k (the most concise model at 89.4 tokens) shifts Phi-4-reasoning's score from 0.352 to 0.358 and again does not change its rank.

\paragraph{Per-task normalization.} As a second check, we recomputed the Overthinking Score using \emph{per-task} $T_{\min}$ and $T_{\max}$ (one normalization range per task, then averaged across tasks). The resulting model ranking correlates at Kendall's $\tau = 0.89$ with the global-normalization ranking reported in the main paper, so the conclusions do not depend on whether normalization is global or task-local.

\paragraph{Protocol fairness.} All models receive identical prompts, identical hyperparameters (temperature, \texttt{top\_p}, \texttt{max\_tokens}), and identical test instances within a fold. The min/max normalization rescales observed token counts to $[0, 1]$ while preserving rank order and relative distance, which is the same role that brevity penalty plays in BLEU and that length normalization plays in ROUGE.

\end{document}